\documentclass{article}

\PassOptionsToPackage{numbers, compress}{natbib}

 \usepackage[main, final]{neurips_2026}
\usepackage[utf8]{inputenc} % allow utf-8 input
\usepackage[T1]{fontenc}    % use 8-bit T1 fonts
\usepackage{hyperref}       % hyperlinks
\usepackage{url}            % simple URL typesetting
\usepackage{booktabs}       % professional-quality tables
\usepackage{amsfonts}       % blackboard math symbols
\usepackage{nicefrac}       % compact symbols for 1/2, etc.
\usepackage{microtype}      % microtypography
\usepackage{xcolor}         % colors
\usepackage{colortbl}       % \cellcolor / \rowcolor for table highlights
\usepackage{graphicx}
\usepackage{amsmath}
\usepackage{comment}
\usepackage{natbib}
\usepackage{amsmath}

\usepackage{subcaption}
\usepackage{array}
\usepackage{multirow}

\usepackage{hyperref}

\usepackage{amssymb}
\usepackage{mathtools}
\usepackage{amsthm}
\usepackage{natbib}
\usepackage{algorithm}
\usepackage{algpseudocode}
\usepackage{wrapfig}
\newcommand{\meanstd}[2]{#1\,{\scriptsize$\pm$\,#2}}
\newcommand{\retPearson}{-0.997}
\newcommand{\retSlope}{14}

\newcommand{\retFwSlope}{8}
\newcommand{\retFwFirstGain}{2.0}
\newcommand{\retFwRestGain}{0.8}
\title{The Lexical Imprint of Pre-training Data}

\author{%
  Jeonghoon Kim\thanks{Equal contribution\qquad$^{\dagger}$Korea Advanced Institute of Science and Technology (KAIST)}\\
  NAVER Cloud \& KAIST\footnotemark \\ % 2026-11-20: the KAIST note shares the line of the equal-contribution note (text in the \thanks above); \footnotemark still sets the dagger and steps the counter, so the internship note keeps its double dagger
  \And
  Woojin Chung\footnotemark[1]\phantom{$^{*}$}\thanks{Work done during an internship at NAVER Cloud} \\ % \phantom keeps the two marks from overlapping (the style sets each mark in a zero-width box) 
  NAVER Cloud \\ 
  \And
  Woomin Song  \\
  KAIST \\
  \AND
  Cheonbok Park \\
  NAVER Cloud \& KAIST \\
  \And
  Nako Sung \\
  NAVER Cloud \\
\And
  Jinwoo Shin \\
  KAIST \\
\\
}

\begin{document}

\maketitle

%%%%%% For correspondence %%%%%
\begingroup
\renewcommand{\thefootnote}{}
\footnotetext{Correspondence to Jinwoo Shin (\texttt{jinwoo@kaist.ac.kr}) and Jeonghoon Kim (\texttt{jeonghoon.samuel@gmail.com}). }
% \footnotemark\footnotetext{Correspondence to Jinwoo Shin (\texttt{jinwoo@kaist.ac.kr}) and Jeonghoon Kim (\texttt{jeonghoon.samuel@gmail.com}). }
\endgroup
%%%%%%%%%%%%%%%%%%%%%%%%%%%%%%%

%TL;DR: This paper uses Unigram Cross-Entropy (UCE) to show that zero-shot benchmark behavior carries a measurable lexical imprint of the pre-training corpus.
\vskip -0.15in  % 2026-10-26: offsets the internship footnote line so the main text still ends on page 10 (2026-10-09 had restored the default spacing)
\begin{abstract}
  How much of zero-shot benchmark accuracy is captured by word-level alignment between a pre-training corpus and the benchmark target? We study this question through word-level unigram cross-entropy (UCE), a tokenizer-agnostic, model-free measure of lexical alignment between a reference corpus and a target benchmark. Likelihood training shapes the model's predictive distribution from the pre-training corpus, and zero-shot scores under likelihood-based evaluation depend in part on the lexical alignment between that corpus and the benchmark target. Cleanly tracing how the lexical imprint of pre-training data scales with corpus--benchmark alignment requires controlled from-scratch pre-training that varies corpus identity while holding the training recipe fixed. Across this controlled sweep, spanning $10$ zero-shot benchmarks, $4$ pre-training corpora, and $3$ model scales, we find that corpus--benchmark lexical alignment, UCE, consistently tracks zero-shot accuracy on every benchmark, with corpora more aligned with the benchmark (lower UCE) achieving higher accuracy. The same signal extends from measurement to control at finer scopes. At the document level, strengthening the imprint by selecting training documents to maximize lexical alignment with the benchmark produces the accuracy gains predicted by the mechanism. At the prompt level, matching the imprint per prompt by routing each prompt to the drafter whose domain corpus is most lexically aligned with it produces measurable throughput gains over a generalist-drafter baseline. Taken together, pre-training corpora leave a measurable lexical imprint on base-model zero-shot behavior, and UCE provides a model-free probe of lexical alignment between reference and target data.
\end{abstract}

\section{Introduction}

\begin{figure}[t!]
\vskip -0.1in
    \centering
    \includegraphics[width=\textwidth]{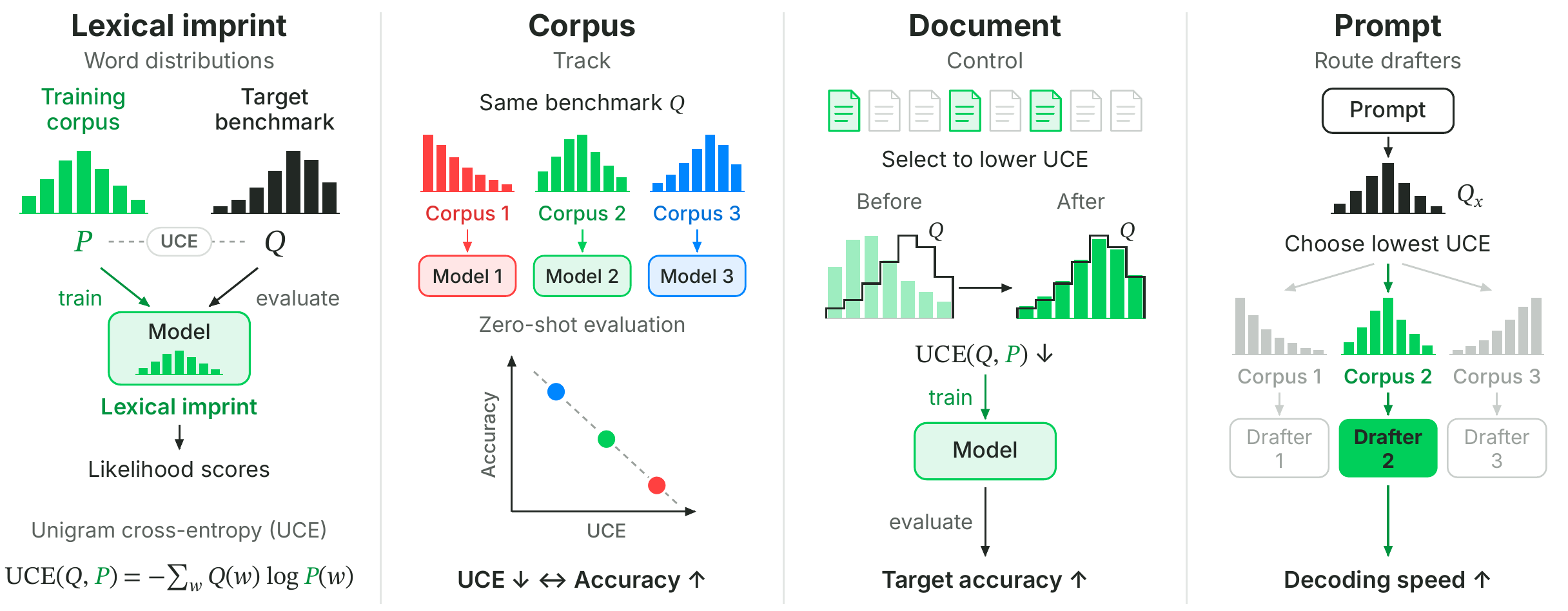}
    \caption{Training on a corpus with word distribution $P$ leaves a lexical imprint on the model, partly visible in likelihood scores on a target benchmark with word distribution $Q$. Word-level unigram cross-entropy (UCE), a tokenizer-agnostic measure of $P$--$Q$ alignment, is a proxy for this imprint. Across training corpora, lower UCE goes with higher zero-shot accuracy on the same benchmark. Selecting training documents that lower UCE strengthens the imprint and raises target accuracy. Routing each prompt to the drafter whose corpus has the lowest UCE against it raises decoding speed.
    }
    \label{fig:concept}
\vskip -0.1in
\end{figure}

Modern language models are typically assessed on zero-shot benchmarks, which are constructed around representative problem formulations and treated as quantitative measurements of model capability \citep{goldblum24,huh24,MMLU,mmlu-pro,humanity_exam,AGI}. A quantitative account of how distinct the benchmark distribution is from the pre-training corpus is a natural input to interpreting such accuracy. This relationship has been studied at the level of individual benchmark items and pieces of the lexicon, where term-level frequency of benchmark vocabulary tracks few-shot accuracy \citep{razeghi_2022}, test-instance lexical popularity correlates with task performance \citep{ruzzetti26_lexical}, and corpus-side perplexity signatures predict performance across panels of pretrained models \citep{wu25_benchsig}. These post-hoc analyses operate on pretrained checkpoints, and the corpus identity that drives the lexical signal is entangled with training recipe, scale, and training history. Cleanly tracing how the lexical imprint of pre-training data scales with corpus--benchmark alignment therefore calls for controlled from-scratch pre-training across corpora, a setting in which this alignment remains under-characterized at scale.

%While the qualitative claim that alignment between training data and downstream evaluation matters is not new, the quantitative version remains an open question. Models with low training loss on high-quality pre-training data tend to achieve stronger downstream performance \citep{gadre25,peri-ln,spikenomore}, and this correlation is consistent with viewing language modeling as a form of compression \citep{deletang24,huang24,depeiges25}, under which likelihood-based predictions on benchmark data are expected to be more reliable when training and benchmark distributions are close. 

Empirical evidence suggests that models achieving lower training loss on high-quality pre-training data tend to attain stronger downstream performance \citep{gadre25,peri-ln,spikenomore}. This relationship is consistent with the compression view of language modeling \citep{deletang24,huang24,depeiges25}, under which likelihood-based predictions on benchmark data are expected to be more reliable when training and benchmark distributions are close. Concretely, how much of zero-shot benchmark performance can the corpus--benchmark distributional alignment alone explain? We approach this question information-theoretically, by summarizing both the pre-training corpus and a benchmark target through their word-level unigram distributions and treating their cross-entropy (UCE) as a measure of distributional alignment. We adopt the word as the fundamental unit since words are discrete, semantically meaningful elements of language\footnote{We follow the standard lexicographic notion of a word as the smallest meaningful unit of speech that can stand by itself (cf.\ the Oxford English Dictionary entry for ``word'').}.
%Our experiments span $11$ benchmarks, $4$ pre-training corpora at token budgets from $8.5\mathrm{B}$ to $70\mathrm{B}$, model scales from $400\mathrm{M}$ to $3.36\mathrm{B}$ parameters, and a $30$B-parameter Stage-2 pre-training case study at production scale.

Across corpora, benchmark vocabulary is fully covered during pre-training, so zero-shot performance differences are driven by how probability mass is allocated over benchmark-relevant words. The lower unigram cross-entropy reflects a closer alignment of corpus-benchmark word-distribution and predicts higher zero-shot accuracy in all benchmarks we test. This admits a simple interpretation. Likelihood-based pre-training learns a conditional distribution from the training corpus, while zero-shot evaluation performs probabilistic inference using the learned distribution by scoring the conditional likelihoods of benchmark candidates. Corpus-level word-distribution alignment may therefore provide a low-order proxy connecting training data to benchmark behavior.

%on which zero-shot evaluation operates. Corpus-level word-distribution alignment may provide a low-order proxy connecting training data to benchmark behavior.

%The corpus word distribution may provide a low-order proxy linking train data to benchmark behavior.

%Likelihood-based pre-training shapes the model’s conditional distribution from the training corpus, on which zero-shot evaluation operates. And corpus word distribution may serve as a tractable low-order proxy connecting pre-training data and benchmark behavior.

%Likelihood training and likelihood-based zero-shot evaluation both operate through word probabilities. The corpus word distribution may provide a natural point of contact between pre-training data and benchmark behavior.

\paragraph{Contributions.}
We propose unigram cross-entropy (UCE) as a single corpus-side lens for tracing the lexical imprint of pre-training data on zero-shot behavior, applied at three points of the language-model lifecycle (Figure~\ref{fig:concept}).
\emph{(i)~At the corpus level.} Across the controlled from-scratch sweep, UCE matches the zero-shot accuracy ordering on every benchmark, with cross-corpus differences captured by how probability mass is allocated to benchmark-relevant words.
\emph{(ii)~At the document level.} Treating UCE as a per-document objective gives the unigram cross-entropy filter (UCE-filter), which controls document-level lexical alignment under a fixed token budget. Under the same from-scratch grid, alignment-driven selection reproduces the corpus-level UCE--accuracy relationship at the document scope.
\emph{(iii)~At the prompt level.} In speculative decoding with a pool of corpus-specialized drafters, evaluating UCE between each prompt and each drafter's training corpus produces a third instance of the same functional acting at a finer scope, with measurable throughput gains over a generalist-drafter baseline.
Together, through this exploration, lexical alignment between a pre-training corpus and a benchmark target, measured by word-level unigram cross-entropy, provides a measurable trace of the lexical imprint that pre-training data leaves on zero-shot behavior.
% An extended discussion of related work is provided in Appendix~\ref{apdx:related}.
%Therefore, we refer to this measurable alignment as a lexical imprint of pre-training data on zero-shot behavior.

\section{Motivation: From Likelihood Training to Lexical Imprints}
\label{sec:motivation}

The downstream performance of modern pre-trained language models is typically assessed by selecting the least surprising candidate under the learned distribution, that is, choosing the option with the highest conditional likelihood given the input \citep{lm-eval-harness}\footnote{Zero-shot multiple-choice evaluation in lm-evaluation-harness is fundamentally likelihood-based. Given a question, the model concatenates each candidate answer and selects the one with the lowest language modeling loss.}. By maximizing next-token likelihood over the training corpus, language-model pre-training estimates empirical conditional patterns, producing a parametric approximation to the data-generating distribution through empirical cross-entropy minimization \citep{deletang24, huang24,depeiges25,lm-eval-harness, chung_vfi_2025, yoran25}. Accordingly, neural language models can be viewed as parametric conditional estimators analogous to classical n-gram models, as both aim to approximate the underlying conditional distribution despite differing parameterizations.

%Since language models are trained to maximize next-token likelihood over the training corpus, the language modeling objective yields a parametric estimator that approximates the data-generating distribution by minimizing KL divergence to the empirical data distribution \citep{deletang24, huang24,depeiges25,lm-eval-harness, chung_vfi_2025, yoran25}. Accordingly, neural language models can be viewed as parametric statistical estimators analogous to classical n-gram models, as both aim to approximate the underlying conditional distribution despite differing parameterizations.

This view motivates the working hypothesis that the model's predicted conditional distribution is, at least approximately, aligned with that statistically estimated from its training data. This hypothesis is supported by empirical and theoretical findings showing that neural language models often capture low-order n-gram statistics, particularly in early training, and that such n-gram statistics provide a meaningful and relatively tight lower bound on model expressivity \citep{Nguyen24,Svete24,Ge25}. Moreover, higher-order n-gram models trained on sufficient data can improve the performance of language models when combined with them \citep{Liu24}. From this perspective, likelihood-based pre-training learns a conditional distribution from the training corpus, while zero-shot evaluation performs probabilistic inference under that distribution by selecting the most likely candidate.
%Since zero-shot evaluation selects the most likely candidate under the learned distribution, both pre-training and downstream predictions rely on the conditional distribution induced by the training corpus. 
Although this structure is infeasible to estimate directly at corpus scale, the corpus-level word-frequency distribution may provide a tractable low-order proxy for the shared dependence between pre-training data and downstream behavior.
%shaped by the training corpus. Both pre-training and zero-shot evaluation therefore pass through the conditional structure shaped by the same corpus, and the corpus's word-level statistics become a natural common cause connecting pre-training data and downstream behavior. 
We refer to this trace as the \emph{lexical imprint} of pre-training data on zero-shot behavior, made measurable through corpus--benchmark word-distribution alignment. This leads to a central research question.

\begin{quote}
\centering
\emph{How much of zero-shot benchmark performance is captured by the alignment between corpus and benchmark word distributions?}
\end{quote}

A quantitative answer would distinguish accuracy gains attributable to corpus alignment from gains reflecting generalization beyond it. Direct estimation of conditional alignment at corpus scale is intractable, so Section~\ref{sec:uce} operationalizes this question through a tractable word-distribution proxy that bridges the pre-training corpus and the benchmark target.

\section{Measuring Lexical Alignment with Unigram Cross-Entropy}
\label{sec:uce}
Our goal is to empirically evaluate how alignment between the conditional distributions of the training and benchmark data affects language model performance. We first introduce word-level unigram cross-entropy (UCE) as a stable and efficient proxy for measuring this alignment (Section \ref{main:metric}), and then motivate the use of word-level unigram cross-entropy as the primary metric (Section \ref{main:metric_reason}).

\subsection{Definition of Unigram Cross-Entropy}\label{main:metric}

Conditional distributions and entropy rates are difficult to estimate directly at corpus scale because of finite-sample bias, data sparsity, smoothing sensitivity, and the limited reliability of long-context statistics, so we proxy them with the word-level unigram distribution (Figure~\ref{fig:concept}). The proxy is coarse but computationally efficient, robust to sparsity, and implicitly reflects character-level n-gram structure since individual words contain rich non-i.i.d. character patterns~\citep{Rajaraman24}.

Let $P$ and $Q$ denote the empirical word-frequency distributions of the pre-training corpus and the benchmark dataset, which we call the \emph{reference distribution} and the \emph{target distribution}, respectively. Word-level unigram cross-entropy (UCE) compares the two datasets solely through these marginals,
\begin{equation*}
UCE\!\left(Q, P\right)
= - \sum_{w \in \mathcal{V}_b} Q(w)\,\log P(w)
= H(Q) + D_{\mathrm{KL}}\!\left(Q \,\|\, P\right),
\end{equation*}
where $\mathcal{V}_b$ is the benchmark vocabulary. Up to the constant $H(Q)$, UCE equals the KL divergence between $Q$ and $P$, so it provides a clean measure of their overlap. Words are obtained by Unicode NFKC normalization, punctuation stripping, and whitespace splitting (no lowercasing), with full implementation details in Appendix~\ref{apdx:uce_implementation}.

%and we apply Laplace (add-one) smoothing to $P_{\text{pre}}$ for words absent from the pre-training corpus, with full implementation details in Appendix~\ref{apdx:uce_implementation}.

\subsection{Why Words and Unigrams?}\label{main:metric_reason}

We adopt word-level unigram cross-entropy for two reasons. First, subword tokenizers are many-to-one mappings, so token-level cross-entropy conflates genuine distributional mismatch with segmentation artefacts induced by token boundaries that do not align with the true data-generating process \citep{phan24,lesci25}. Word-level granularity removes this confound by operating at a unit common to the corpus and the benchmark. Second, although higher-order n-gram cross-entropy for fixed n decomposes into a target-dependent constant and a conditional KL divergence between the two datasets, reliable estimation at corpus scale is hindered by sparsity in the context space, making the estimate sensitive to finite-sample bias and smoothing choices. Word-level unigram cross-entropy instead decomposes exactly into the marginal entropy of the target distribution and the KL divergence between the corresponding word distributions. It therefore avoids the estimation difficulties associated with higher-order conditional statistics.

\section{Corpus-Level Lexical Alignment Tracks Zero-Shot Behavior} \label{sec:corpus_rank}

\begin{figure*}[t!]
\vskip -0.1in
    \centering
    \includegraphics[width=\textwidth]{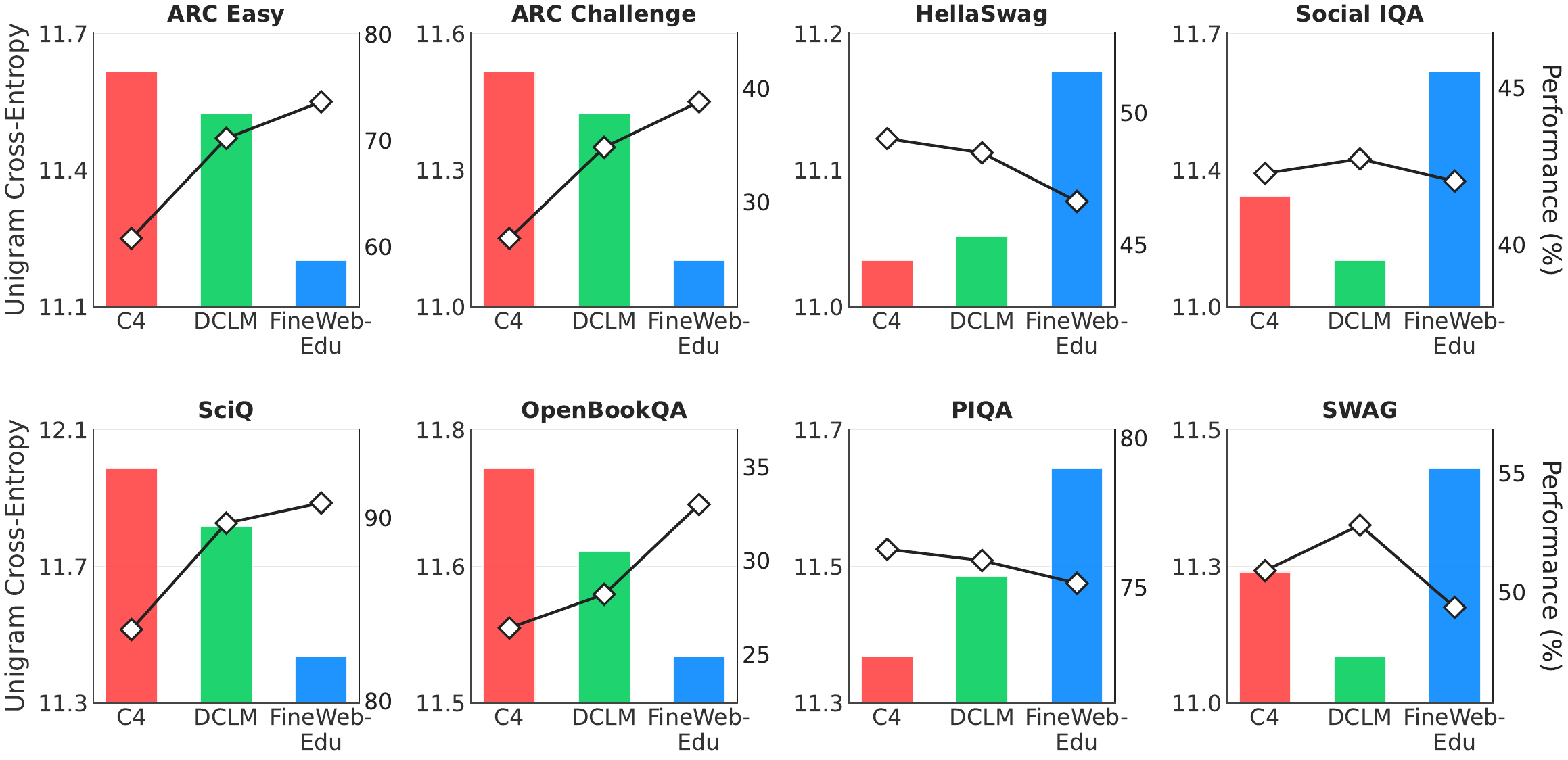}
    \caption{Word-level unigram cross-entropy (left axis, bars, bits $\downarrow$) and zero-shot accuracy (right axis, diamonds) across $8$ benchmarks for a $3.36\mathrm{B}$ model trained on $60\mathrm{B}$ tokens of \textcolor[HTML]{FF4040}{\textbf{C4}}, \textcolor[HTML]{00CF5A}{\textbf{DCLM}}, or \textcolor[HTML]{0086FF}{\textbf{FineWeb-Edu}}. Lower cross-entropy tracks higher accuracy in every panel. Full per-task results across all scales are in Appendix~\ref{apdx:word_overlap_correlation}.}
    \label{fig:figure1}
\vskip -0.1in
\end{figure*}

This section examines whether the word-level unigram cross-entropy (UCE) of Section~\ref{sec:uce} tracks zero-shot benchmark accuracy at the corpus level, and whether the trend is stable across pre-training corpora, model scales, and random subsamples.

\subsection{From-Scratch Pre-training and Evaluation Protocol} \label{main:setting}

For the training setup, we pre-train decoder-only Transformer models~\citep{llama2} from scratch at three scales ($400\mathrm{M}$, $1.33\mathrm{B}$, $3.36\mathrm{B}$) on four pre-training corpora spanning a range of data quality, namely the high-quality FineWeb-Edu~\citep{fineweb24} and DCLM~\citep{datacomp}, the earlier high-quality C4~\citep{c4}, and the comparatively lower-quality OpenWebText~\citep{openwebtext}. From C4, DCLM, and FineWeb-Edu we draw three random sub-samples of $8.5\mathrm{B}$, $26\mathrm{B}$, and $60\mathrm{B}$ tokens. The matched (model, subset) pairings $(400\mathrm{M}, 8.5\mathrm{B})$, $(1.33\mathrm{B}, 26\mathrm{B})$, and $(3.36\mathrm{B}, 60\mathrm{B})$ each sit at or near the Chinchilla compute-optimal ratio of $20$ tokens per parameter \citep{hoffmann22_chinchilla}. OpenWebText is only $\sim\!9\mathrm{B}$ tokens in total and is therefore evaluated at the $8.5\mathrm{B}$ scale only. All from-scratch runs in this section and in Section~\ref{sec:ce_filter_results} use AdamW with warmup and cosine learning-rate decay, a sequence length of $2048$, and a global batch size of $256$, and count tokens with the GPT-2 tokenizer~\citep{gpt2}. The two studies differ in peak learning rate, warmup length, $\beta_2$, and weight decay (Appendix~\ref{Apdx:setting}, Tables~\ref{tab:train-config}--\ref{tab:cefilter-arch}).

For the evaluation setup, we measure zero-shot performance on a pool of $12$ benchmarks (ARC~Easy~\citep{ARC}, ARC~Challenge~\citep{ARC}, HellaSwag~\citep{Hellaswag}, MMLU~\citep{MMLU}, SciQ~\citep{sciq}, OpenBookQA~\citep{openbookqa}, PIQA~\citep{piqa}, LAMBADA~\citep{lambada}, SocialIQA~\citep{siqa}, SWAG~\citep{swag}, CommonsenseQA~\citep{commonsenseqa}, and BoolQ~\citep{boolq}). This section uses the $10$ benchmarks other than CommonsenseQA and BoolQ, and the UCE-filter study of Section~\ref{sec:ce_filter_results} uses the $11$ other than MMLU. Per-task metrics differ between the two studies and are listed in Appendix~\ref{apdx:setting:corpus}. Each corpus is used as released, and Section~\ref{sec:contamination} audits all four with a $13$-gram index over the $12$ benchmarks and tests whether removing the overlapping benchmark items or training documents changes the UCE-filter results.

\subsection{Lexical Alignment Matches Corpus-Level Benchmark Rankings} \label{main:main_result}
% [2026-11-20] Figure 3 placed here (was after Section 4.3) so that it floats to the page where Section 4.3 begins
\begin{figure}[t!]
\vskip -0.2in
    \centering
    \includegraphics[width=\columnwidth]{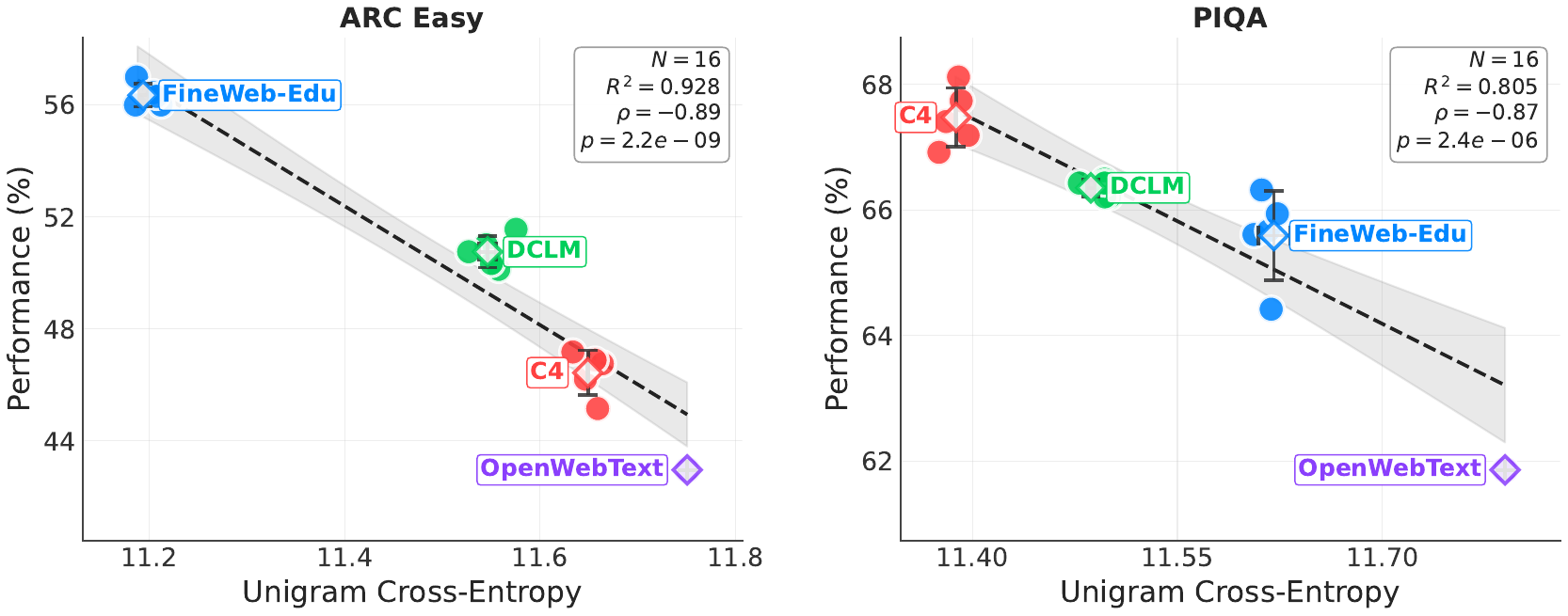}
    \caption{Cross-entropy (bits $\downarrow$) versus zero-shot accuracy for $N{=}16$ random $8.5\mathrm{B}$-token subsets at the $400\mathrm{M}$ scale ($5$ each from \textcolor[HTML]{FF4040}{\textbf{C4}}, \textcolor[HTML]{00CF5A}{\textbf{DCLM}}, and \textcolor[HTML]{0086FF}{\textbf{FineWeb-Edu}}, and $1$ from \textcolor[HTML]{8A3FFC}{\textbf{OpenWebText}}). Solid dots mark per-subset points, open diamonds with whiskers mark per-corpus mean $\pm 1\sigma$.}
    \label{fig:figure2}
\vskip -0.1in
% \vskip -0.05in
\end{figure}

To assess whether corpus--benchmark alignment in word-frequency distributions predicts downstream performance and ranks pre-training corpora accordingly, we measure the UCE between every training subset of Section~\ref{main:setting} and each of the $10$ benchmarks, and compare the corpus ranking by UCE with the ranking by downstream score at each (model, subset) pairing.

For a $3.36\mathrm{B}$ model trained on $60\mathrm{B}$ tokens, Figure~\ref{fig:figure1} shows that UCE predicts the accuracy order of the three corpora on all eight benchmarks. The newer DCLM and FineWeb-Edu, commonly regarded as higher-quality data~\citep{fineweb24,datacomp}, lead with the lowest UCE on six of them, FineWeb-Edu on the knowledge-heavy ARC Easy/Challenge, SciQ, and OpenBookQA, and DCLM on SocialIQA and SWAG. Their UCE is higher than C4's wherever they trail C4, on HellaSwag and PIQA for both and on SocialIQA and SWAG for FineWeb-Edu. UCE thus tracks the task-dependent overlap that quality reputation misses, and this overlap plays a central role in shaping downstream performance. Because curation and genre also vary across these corpora, the corpus-level UCE--accuracy relationship is associational, and Section~\ref{sec:data_selection} tests it within each source corpus at a fixed token budget.

Decomposing the corpus--benchmark overlap shows that all four corpora attain near-saturated coverage of the benchmark vocabulary at our scales and differ in how they allocate probability mass over that vocabulary. On every benchmark in Figure~\ref{fig:figure1}, the higher a corpus scores, the smaller the share of benchmark word mass that falls on its long tail of rare words. Appendix~\ref{apdx:coverage_decomp} discusses this decomposition, and Appendix~\ref{apdx:word_overlap_correlation} shows that the inverse relationship between performance and UCE holds across the $10$ benchmarks, dataset sizes, and model scales.

\subsection{The Imprint Is Stable Across Random Subsets} \label{main:consistent}

To test whether the inverse relationship depends on the specific subset chosen, we sample five different $8.5\mathrm{B}$-token subsets each from C4, DCLM, and FineWeb-Edu and one from OpenWebText, then pre-train a $400\mathrm{M}$ model on each. The inverse relationship holds across the $16$ runs in Figure~\ref{fig:figure2}. Within each corpus the subsets have nearly the same UCE and differ little in accuracy.

\begin{comment}
\begin{algorithm}[t]
\caption{Row selection in Unigram Cross-Entropy filter}
\label{alg:ce_filter_select}
\begin{algorithmic}[1]
%\State \textbf{Input.}
%\Statex \quad Benchmark distribution $Q$ over $\mathcal{V}_b$
%\Statex \quad Training rows $\mathcal{T}=\{r_i\}_{i=1}^N$
%\Statex \quad Removal budget $K$
\State \textbf{Input:} benchmark distribution $Q$ over $\mathcal{V}_b$, training rows $\mathcal{T}=\{r_i\}_{i=1}^N$, removal budget $K$
\State Compute full-corpus counts $c(w)$, total token count $M$, the smoothed training distribution $P$, and $H(Q,P)$
\State Compute benchmark-aware training counts $c(w)$ over $\mathcal{V}_b$ (flooring zero counts at one), baseline cross-entropy $H(Q,P)$
\For{each row $r_i\in\mathcal{T}$}
    \State Extract row-local counts $c_i(w)$
    \State Analytically compute $\Delta H_i = H(Q,P^{\setminus r_i}) - H(Q,P)$
\EndFor
\[
\mathcal{R}
\gets
\left\{
r_i \in \mathcal{T}
:
i \in \operatorname{arg\,topK}_{j}\left(-\Delta H_j\right)
\right\}.
\]
\State \textbf{Output.}\ \ $\mathcal{T} \setminus \mathcal{R}$
\end{algorithmic}
\end{algorithm}
\end{comment}

\section{Controlling Document-Level Lexical Alignment via Data Selection}
\label{sec:data_selection}

Section~\ref{sec:corpus_rank} established that lower unigram cross-entropy (UCE) tracks higher zero-shot accuracy at the corpus level. This section moves one step beyond that descriptive observation and asks what happens when the pre-training corpus is made more lexically aligned with the benchmark target at a fixed token budget. To do so, we propose the UCE-filter, a data-selection tool that controls corpus--benchmark lexical alignment through row-local cross-entropy filtering\footnote{A \emph{row} here denotes a single training example, namely one document drawn from the source corpus. We adopt this terminology from the row-oriented format of Common Crawl-derived pre-training corpora such as DCLM~\citep{datacomp} and FineWeb-Edu~\citep{fineweb24}, which are distributed with one document per row. The UCE-filter scores and selects at this granularity.}. Our aim is to use it as a controlled probe that turns the descriptive corpus-level signal of Section~\ref{main:main_result} into a prescriptive selection criterion under a known target, so that the analytical relationship between reference corpus, benchmark target, and language-model behavior can be measured rather than only described. We do not advance the UCE-filter as a stand-alone data-selection method.

\subsection{Document-Level Unigram Cross-Entropy Filtering}
\label{sec:ce_filter}

% \vskip -0.1in 
\begin{algorithm}[t]
% \vskip -0.1in 
\caption{One-shot row selection in the UCE filter}
\label{alg:ce_filter_select}
\begin{algorithmic}[1]
%\State \textbf{Input:} benchmark distribution $Q$ over $\mathcal{V}_b$, training rows $\mathcal{T}=\{r_i\}_{i=1}^N$, removal budget $K$
%\Statex \quad Compute benchmark-aware training counts $c(w)$ over $\mathcal{V}_b$ (flooring zero counts at one)
\State \textbf{Input:} reference distribution $P$, target distribution $Q$, training rows
$\mathcal{T}=\{r_i\}_{i=1}^N$
\Statex removal budget $K$, pre-training corpus counts $c(w)$
%\State Flooring zero counts for word $w \in P$ where $w \notin \mathcal{V}_b$, Compute cross-entropy $H(Q,P)$
\State  Set
$B \gets \sum_{w\in\mathcal{V}_b}\mathbf{1}[c(w)=0]$, 
\ 
Define
$P(w)
\gets
\frac{\max\{c(w),1\}}{C+B}$
for all $w\in\mathcal{V}_b$
\For{each row $r_i\in\mathcal{T}$}
    \State Compute $C_i \gets \sum_{w \in \mathcal{V}_b} c_i(w)$ and row-local counts $c_i(w)$ 
    \State Define $P^{\setminus r_i}(w)
    \gets
    \frac{\max\{c(w)-c_i(w),1\}}
    {C-C_i+B^{\setminus r_i}}$ where  
    $B^{\setminus r_i}
    \gets
    \sum_{w\in\mathcal{V}_b}
    \mathbf{1}[c(w)-c_i(w) \leq 0]$
    \State Analytically compute 
    \[
    \Delta H_i \gets H(Q,P^{\setminus r_i})-H(Q,P)
    \]
\EndFor
\State Select 
\[ 
\mathcal{R} \gets \left\{ r_i \in \mathcal{T} : i \in \operatorname{arg\,topK}_{j\in\{1,\ldots,N\}} \left(-\Delta H_j\right) \right\}, \quad |\mathcal{R}|=K. 
\]

\State \textbf{Output:} $\mathcal{T}\setminus\mathcal{R}$
\end{algorithmic}
% \vskip -0.04in  % 2026-10-09: default spacing restored, page 10 has room
\end{algorithm}
% \vskip -0.2in 

We define the UCE-filter, a row-local rule that retains the subset of training rows lowering the corpus--benchmark cross-entropy $H(Q, P)$. The row-local update $\Delta H_r$ is analytically tractable and rankable in a single linear pass, which is what a controlled probe requires at multi-billion-row scale. For efficiency, we use one-shot ranking which ignores interactions among removed rows. Importance resampling based on hashed n-gram features \citep{xie_dsir_2023} and the cross-entropy difference filter of \citet{moore_lewis_2010} are the closest existing rules. For each training row $r$, the UCE-filter measures corpus--benchmark cross-entropy by
\[
\Delta H_r = H(Q, P^{\setminus r_i}) - H(Q, P),
\]
where $Q$ is the benchmark unigram distribution and $P$ is the reference unigram distribution. The UCE-filter discards the rows with the smallest $\Delta H_r$, namely those whose presence most increases $H(Q, P)$, and retains the rest. 
%The UCE-filter retains the rows with the largest $\Delta H_r$, namely those whose presence most reduces $H(Q, P)$, and discards the rest. 
Because removing a row only changes the counts of words appearing in that row, $\Delta H_r$ admits an analytic row-local update from row-local counts and benchmark-aware corpus counts, without re-estimating the full training distribution. Algorithm~\ref{alg:ce_filter_select} summarizes the procedure. The sign convention, the row-local efficiency argument, and the parallel implementation we use at multi-billion-row scale are deferred to Appendix~\ref{sec:ce_filter_algo}.

\begin{figure}[t]
\vskip -0.1in
\centering
\includegraphics[width=\textwidth]{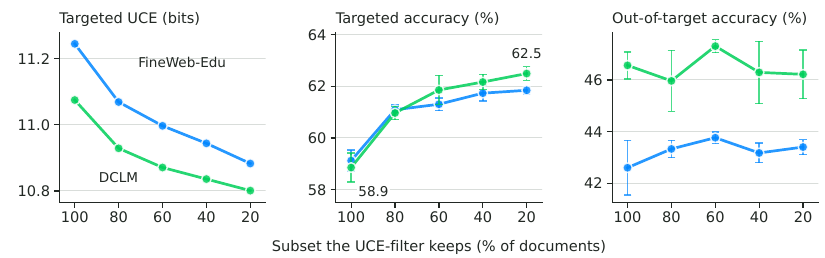}
\caption{Targeted UCE of the subsets that the UCE-filter keeps from \textcolor[HTML]{00CF5A}{\textbf{DCLM}} and \textcolor[HTML]{0086FF}{\textbf{FineWeb-Edu}} (left), and zero-shot accuracy of $1.34\mathrm{B}$ models trained from scratch on $70\mathrm{B}$ tokens of each subset (middle and right, drawn with equal vertical spans). Targeted averages the $7$ tasks used for selection (ARC-Easy, ARC-Challenge, HellaSwag, SciQ, OpenBookQA, PIQA, SWAG), and Out-of-target the $4$ other tasks (LAMBADA, CommonsenseQA, BoolQ, SocialIQA).}
\label{fig:retention_sweep}
\vskip -0.1in
\end{figure}

% ===== [added 2026-11-17, draft] retention sweep: accuracy against the retention rate of the UCE-filter =====
% Data and scripts: paper_work/retention_sweep/ (collect_results.py, make_retention_figure.py, update_paper.py) on the working branch.
% Plotted: DCLM and (from 2026-11-18) FineWeb-Edu, every training seed finished so far (\retStatus). 2026-11-20 (user):
% retention 100-20% only (the 10% and 5% runs stay out of the paper), and at 100% and 20% the Table 1+2 runs (Random,
% UCE-filter) of each seed are plotted (update_paper.py --ref paper). The r and slope in the results paragraph are for DCLM.
\subsection{Lower Target UCE Tracks Higher Target Accuracy}
\label{sec:ce_filter_retention}

To test whether zero-shot accuracy on a target follows the amount by which selection lowers UCE against it, we vary the share of rows that the UCE-filter retains from DCLM and from FineWeb-Edu. The target $Q$ is aggregated over $7$ commonsense tasks, denoted CSR7 (ARC-Easy, ARC-Challenge, HellaSwag, SciQ, OpenBookQA, PIQA, SWAG). The filter ranks the rows of each corpus once against this $Q$, and each arm keeps the top $20$, $40$, $60$, or $80\%$ of the ranking, with the unfiltered corpus as the $100\%$ arm. Under the protocol of Section~\ref{main:setting}, a $1.34\mathrm{B}$ model (counting the tied embedding) is trained from scratch on $70\mathrm{B}$ tokens sampled uniformly from each arm, so the arms differ only in the retained share. Each arm is repeated under three independent training seeds, and the $100\%$ and $20\%$ arms are the Random and UCE-filter runs that Section~\ref{sec:ce_filter_results} compares with other selection methods. Four out-of-target tasks, denoted OOT-4 (LAMBADA, CommonsenseQA, BoolQ, SocialIQA), test whether a gain is specific to the target. Their word distribution shares $60\%$ of its probability mass with that of CSR7 (total variation distance $0.40$), close to what each CSR7 task shares with the other six (Table~\ref{tab:benchmark_cross_uce}).

Figure~\ref{fig:retention_sweep} shows that on DCLM, Targeted accuracy rises at every step toward a smaller retained share, as the Targeted UCE of the retained subset falls. The two changes also match in size, since both are largest when the first $20\%$ of rows are discarded and smaller at every later step. Across the retention rates, Targeted accuracy is close to linear in Targeted UCE (Pearson $r=\retPearson$, about $\retSlope$~pp per bit of UCE reduction). The out-of-target average has no such trend and moves in both directions around the value of the unfiltered corpus. Within one source corpus and at a fixed token budget, accuracy on the target thus follows how far selection lowers UCE against it, so the lexical imprint that Section~\ref{sec:corpus_rank} observes across corpora also appears when alignment is changed within a corpus. On FineWeb-Edu, most of the gain in Targeted accuracy comes in the first step ($\retFwFirstGain$~pp from $100\%$ to $80\%$ retention and $\retFwRestGain$~pp from $80\%$ to $20\%$), and a linear fit of Targeted accuracy on Targeted UCE has a slope of about $\retFwSlope$~pp per bit of UCE reduction, shallower than on DCLM. The out-of-target average on FineWeb-Edu changes much less than Targeted accuracy and has no consistent direction between $80\%$ and $20\%$ retention. At equal Targeted UCE the two corpora can differ in Targeted accuracy, because the pooled $Q$ takes most of its word mass from HellaSwag and SWAG while Targeted accuracy weights the $7$ tasks equally. Measured against each task alone, as in Section~\ref{main:main_result}, the corpus with the lower UCE has the higher accuracy on every task but SciQ, where the two tie (Appendix~\ref{apdx:q_robustness}).

\subsection{Comparison with Other Data Selection Methods}
\label{sec:ce_filter_results}

% We test whether the UCE-filter of Section~\ref{sec:ce_filter} yields measurable downstream gains.

To compare the UCE-filter with other data selection methods at a fixed token budget, we evaluate it under the protocol of Section~\ref{main:setting} on a $2{\times}2$ grid of source (DCLM, FineWeb-Edu) and parameter scale ($1.34\mathrm{B}$, $3.33\mathrm{B}$, counting the tied embedding). The models follow the DataComp-LM architecture patterns~\citep{datacomp} and train on $70\mathrm{B}$ tokens from each selected subset, at or above the Chinchilla compute-optimal ratio for both scales \citep{hoffmann22_chinchilla}. The UCE-filter retains the $20\%$ of rows most aligned with the CSR7 target $Q$ of Section~\ref{sec:ce_filter_retention}. A random control alone is insufficient to attribute an accuracy gain specifically to alignment, since any rule that selects toward the benchmark $Q$ could raise accuracy on the targeted suite. We therefore compare against two same-source controls matched to the $70\mathrm{B}$-token budget, a row-count-matched uniform random sample and a faithful reproduction of the DSIR importance-resampling rule of \citet{xie_dsir_2023}, which targets the same $Q$ through hashed n-gram features. Each (Source, Scale, Method) cell is repeated under three independent training seeds (Appendix~\ref{apdx:seed_sweep}), and the Overall average covers the $11$ CSR7 and OOT-4 tasks. Appendices~\ref{apdx:setting:cefilter} and~\ref{sec:dsir_baseline} give the full setup and the DSIR implementation.

Table~\ref{tab:ce_filter_uce} shows that the UCE-filter lowers Targeted UCE well below both controls, and DSIR lowers it by less. These Targeted gaps far exceed selection noise, since the UCE-filter is deterministic and the Targeted UCE of Random and DSIR varies by less than $0.001$ bits across selection seeds (Appendix~\ref{apdx:seed_sweep}). The UCE-filter and DSIR subsets have out-of-target UCE values close to those of the random control, suggesting that selection mainly lowers UCE for the specified target. Within each source, the subset with the lowest Targeted UCE also has the highest Targeted accuracy (Table~\ref{tab:ce_filter_csr7_main}). Like the alignment, the UCE-filter's accuracy gain concentrates on the targeted suite, with the Targeted average above the random control in every (Source, Scale) cell and the out-of-target average close to it. Given the same $Q$, DSIR gains less on the targeted suite and falls below the random control on the out-of-target average in every cell. Selecting documents for lower UCE against a target thus raises zero-shot accuracy mainly on that target. Because $Q$ is estimated from the target tasks, the selection gains should be interpreted as an upper bound on the alignment effect under a known target rather than as zero-shot generalization to unseen targets. Appendix~\ref{sec:dsir_baseline} and Table~\ref{tab:ce_filter_csr7_full} attribute much of DSIR's Targeted shortfall against the UCE-filter, and of its LAMBADA loss, to a length bias in its document scores. BETR~\citep{mizrahi25_betr}, an embedding-similarity selector given the same CSR7 targets, raises Targeted accuracy at the cost of the out-of-target average while its Targeted UCE stays at the DSIR level, and it trails the UCE-filter on the Targeted average except at $1.34\mathrm{B}$ on DCLM (Appendix~\ref{sec:ce_family_baselines}). Appendix~\ref{apdx:q_robustness} shows that the UCE gap between the UCE-filter and random subsets is unchanged when measured against a $Q$ built from the train splits alone, without the test splits it otherwise pools. The UCE-filter's Targeted gain over the random control survives contamination removal at both ends. At a matched training seed, dropping the evaluation items flagged by the $13$-gram scan moves it by at most $0.05$~pp in every cell, and removing flagged DCLM documents before re-selection and retraining at $1.34\mathrm{B}$ keeps $3.40$~pp of the same-seed $4.13$~pp gain (Appendix~\ref{apdx:contamination}). On the Targeted average, selection-seed and training-seed accuracy variation are of the same order, and the method ordering of Table~\ref{tab:ce_filter_csr7_main} holds at every selection seed (Appendix~\ref{apdx:select_seed_sweep}). Under $5$-shot prompting the UCE-filter has the highest Targeted and out-of-target means in every cell (Appendix~\ref{apdx:ce_filter_per_task_acc_5shot}).

% ===== [2026-11-17] Tables 1 and 2 merged into one table (user decision); see the comment at the end of the tabular =====
\begin{table}[t]
\vskip -0.1in
\small
\centering
\caption{UCE (bits $\downarrow$) of the subsets selected from each source corpus, and zero-shot accuracy (\%) of $1.34\mathrm{B}$ and $3.33\mathrm{B}$ models trained from scratch on $70\mathrm{B}$ tokens of each subset, as mean $\pm$ standard deviation over three training seeds. Targeted refers to the $7$ tasks used for selection (ARC-Easy, ARC-Challenge, HellaSwag, SciQ, OpenBookQA, PIQA, SWAG), and Out-of-target to the $4$ other tasks (LAMBADA, CommonsenseQA, BoolQ, SocialIQA). UCE is measured against the word distribution pooled over the tasks of a block, and accuracy is averaged over them. A random subset has the UCE of the full source corpus. The Out-of-target UCE is measured on a fixed-seed sample of each subset, and its values are comparable only within the block.}
\label{tab:ce_filter_uce}\label{tab:ce_filter_csr7_main}
\setlength{\tabcolsep}{4pt}
\resizebox{\textwidth}{!}{%
\begin{tabular}{@{}llcc>{\columncolor[HTML]{E0F8EB}}ccc>{\columncolor[HTML]{E0F8EB}}c@{}}
\toprule
& &
\multicolumn{3}{c}{\textbf{Targeted ($7$-task)}} &
\multicolumn{3}{c}{\textbf{Out-of-target ($4$-task)}} \\
\cmidrule(lr){3-5} \cmidrule(lr){6-8}
\textbf{Source} & \textbf{Measure} &
\textbf{Random} & \textbf{DSIR}~\citep{xie_dsir_2023} & \textbf{UCE-Filter} &
\textbf{Random} & \textbf{DSIR} & \textbf{UCE-Filter} \\
\midrule

\multirow{3}{*}{DCLM}
& UCE (bits $\downarrow$) & 11.17 & 11.08 & 10.92 & 11.39 & 11.43 & 11.41 \\
& Acc.\ $1.34\mathrm{B}$ (\%) & \meanstd{58.85}{0.56} & \meanstd{59.85}{0.53} & \meanstd{62.49}{0.26}
          & \meanstd{46.57}{0.53} & \meanstd{43.79}{0.23} & \meanstd{46.22}{0.94} \\
& Acc.\ $3.33\mathrm{B}$ (\%) & \meanstd{62.95}{0.03} & \meanstd{63.01}{0.33} & \meanstd{66.26}{0.20}
          & \meanstd{48.75}{0.53} & \meanstd{46.23}{0.38} & \meanstd{49.56}{0.09} \\
\midrule

\multirow{3}{*}{FineWeb-Edu}
& UCE (bits $\downarrow$) & 11.32 & 11.20 & 10.93 & 11.62 & 11.65 & 11.58 \\
& Acc.\ $1.34\mathrm{B}$ (\%) & \meanstd{59.12}{0.41} & \meanstd{60.15}{0.08} & \meanstd{61.85}{0.15}
          & \meanstd{42.60}{1.06} & \meanstd{41.14}{1.37} & \meanstd{43.40}{0.29} \\
& Acc.\ $3.33\mathrm{B}$ (\%) & \meanstd{62.80}{0.09} & \meanstd{62.77}{0.18} & \meanstd{64.64}{0.21}
          & \meanstd{44.85}{0.49} & \meanstd{43.39}{0.33} & \meanstd{45.61}{0.27} \\
\bottomrule
\end{tabular}}
\vskip -0.1in
\end{table}

\begin{table}[t]
% \vskip -0.1in
\small
\centering
\caption{Zero-shot MMLU and MMLU-Pro accuracy of a $30\mathrm{B}$-parameter model after Stage~1 and after Stage~2, where Stage~2 uses either the production mixture or the same mixture with the web component replaced by its UCE-filtered subset. Tokens is the number of Stage~2 training tokens.}
\label{tab:ce_filter_30b}
\setlength{\tabcolsep}{8pt}
\begin{tabular}{@{}llccc@{}}
\toprule
\textbf{Stage} & \textbf{Strategy} & \textbf{Tokens} & \textbf{MMLU} & \textbf{MMLU-Pro} \\
\midrule
Stage 1 & Endpoint ($=$ Stage 2 init) & init.\               & 71.17 & 44.46          \\
\midrule
\multirow{2}{*}{Stage 2}
        & Production Mix (baseline)   & ${\sim}2\mathrm{T}$  & 75.61 & 48.64          \\
        & \cellcolor[HTML]{E0F8EB}\textbf{UCE-Filter Web Subset} & \cellcolor[HTML]{E0F8EB}$\mathbf{200}\mathrm{B}$ & \cellcolor[HTML]{E0F8EB}74.28 & \cellcolor[HTML]{E0F8EB}48.81 \\
\bottomrule
\end{tabular}
\vskip -0.1in
\end{table}

\subsection{30B Stage-2 Production-Scale Case Study}
\label{sec:ce_filter_30b}

To test whether the alignment signal isolates a subset of a heavily curated pre-training mixture that accounts for most of its gain, we apply the UCE-filter in the pre-training of an in-house $30\mathrm{B}$-parameter language model that largely follows the multi-stage pre-training recipe and architecture of Nemotron-3 Nano~\citep{nvidia_nemotron_nano_v3_2025}. The model's Stage~$2$ continues from the Stage~$1$ endpoint on a $2\mathrm{T}$-token production mixture, and the filtered variant replaces the web component of that mixture with its UCE-filtered subset, keeps the non-web components and the production hyperparameters, and trains on $200\mathrm{B}$ tokens in total. Detailed settings are in Appendix~\ref{apdx:30b_setup}.

In Table~\ref{tab:ce_filter_30b}, the UCE-filtered web variant keeps the bulk of the Stage~$2$ gain at roughly $10\times$ fewer tokens, with the production mixture's MMLU-Pro matched and most of the Stage~$1\!\to\!2$ MMLU gap closed. The MMLU shortfall is consistent with the web-only nature of the filter, since the shorter run also sees less of the code, math, and supervised data, which the filter leaves untouched. %The pattern is one of isolation rather than improvement, and we do not claim general improvement over a heavily curated full mix.

\section{Inference-Time Routing via Prompt-Level Lexical Alignment}
\label{sec:routing}

% \paragraph{Setting.} Speculative decoding drafts candidate tokens with a
% fast drafter model and verifies them in parallel against a large target
% model \citep{leviathan23, chen23_specdec}. Given a pool of $K$
% corpus-specialized drafters $\{M_d^{(k)}\}_{k=1}^{K}$ with training-word
% distributions $\{P_d^{(k)}\}$, we route each prompt $x$ (with empirical
% word distribution $P_x$) to the drafter $M_d^{(k^\star)}$ where
% $k^\star = \arg\min_k H(P_x, P_d^{(k)})$, reusing the unigram
% cross-entropy (UCE). The $P_d^{(k)}$ are precomputed once, so per-prompt
% routing reduces to a sparse dot product with negligible latency. We
% train three specialist drafters on math, code, and science subsets of
% OpenThoughts-114k~\citep{guha2025openthoughtsdatarecipesreasoning}, and
% a generalist baseline drafter on an equal-sized mixture of the three.
% All drafters use the SAM architecture~\citep{hu-etal-2025-sam} and
% matched corpus sizes. Evaluation spans math (AIME~\citep{aime}), code
% (LiveCodeBench~\citep{livecodebench}), general knowledge
% (MMLU~\citep{MMLU}), and medical question answering (MedQA~\citep{medqa}).

To examine whether lexical alignment continues to act at the prompt scope, we apply the same functional per prompt in speculative decoding, which drafts candidate tokens with a fast drafter model and verifies them in parallel against a large target model \citep{leviathan23, chen23_specdec}. Given a verifier $M_v$ (DeepSeek-R1-Distill-Llama-8B~\citep{deepseek_r1_2025}) and a pool of $K$ corpus-specialized drafters $\{M_d^{(k)}\}_{k=1}^{K}$ with corpus word distributions $\{P_d^{(k)}\}$, we route each prompt $x$ (with empirical word distribution $Q_x$) to the drafter $M_d^{(k^\star)}$ where $k^\star = \arg\min_k H(Q_x, P_d^{(k)})$, reusing the unigram cross-entropy (UCE). The $P_d^{(k)}$ are precomputed once at the corpus level, so per-prompt routing reduces to a sparse dot product with negligible latency. Specifically, we construct three specialist drafters using the OpenThoughts-114k~\citep{guha2025openthoughtsdatarecipesreasoning} dataset, each corresponding to a sub-domain (math, code, and science). As a baseline, we also construct a generalist drafter, using equal-sized subsets from each specialist drafter's training corpus. We use SAM~\citep{hu-etal-2025-sam} as the drafter architecture, and all drafters are constructed using the same corpus size for fair comparison. We evaluate the speculative decoding performance on four benchmarks spanning math (AIME~\citep{aime}), code (LiveCodeBench~\citep{livecodebench}), general knowledge (MMLU~\citep{MMLU}), and medical question answering (MedQA~\citep{medqa}).

In Table~\ref{tab:ce_routing_specdec}, UCE-based routing adds $5.3$~pp of average speedup over the generalist drafter and raises both $\tau$ and speedup on every benchmark, including MMLU and MedQA, which no specialist targets. The same alignment functional that controls corpus identity at the document scope therefore also acts at the prompt scope, making the lexical imprint visible.

\begin{table}[t]
\vskip -0.1in
\small
\centering
\caption{Mean acceptance length $\tau$ and per-token decoding speedup of speculative decoding with DeepSeek-R1-Distill-Llama-8B as the verifier, using a generalist drafter or UCE-based routing among specialist drafters. Speedup is relative to the verifier alone (Base Model).}
\label{tab:ce_routing_specdec}
\setlength{\tabcolsep}{8pt}
\resizebox{\textwidth}{!}{
\begin{tabular}{@{}lcccccccccc@{}}
\toprule
 & \multicolumn{2}{c}{AIME} & \multicolumn{2}{c}{LiveCodeBench} & \multicolumn{2}{c}{MMLU} & \multicolumn{2}{c}{MedQA} & \multicolumn{2}{c}{Average} \\ 
\cmidrule(lr){2-3} \cmidrule(lr){4-5} \cmidrule(lr){6-7} \cmidrule(lr){8-9} \cmidrule(l){10-11}
 & $\tau$ & Speedup & $\tau$ & Speedup & $\tau$ & Speedup & $\tau$ & Speedup & $\tau$ & Speedup \\
\midrule
Base Model & 1.000 & - & 1.000 & - & 1.000 & - & 1.000 & - & 1.000 & - \\
Generalist Drafter & 1.702 & 55.6\% & 1.587 & 53.0\% & 1.436 & 45.7\% & 1.409 & 45.7\% & 1.534 & 50.0\% \\
\cellcolor[HTML]{E0F8EB}\textbf{UCE-Based Routing} & \cellcolor[HTML]{E0F8EB}\textbf{1.707} & \cellcolor[HTML]{E0F8EB}\textbf{61.6\%} & \cellcolor[HTML]{E0F8EB}\textbf{1.615} & \cellcolor[HTML]{E0F8EB}\textbf{54.7\%} & \cellcolor[HTML]{E0F8EB}\textbf{1.451} & \cellcolor[HTML]{E0F8EB}\textbf{55.3\%} & \cellcolor[HTML]{E0F8EB}\textbf{1.426} & \cellcolor[HTML]{E0F8EB}\textbf{49.6\%} & \cellcolor[HTML]{E0F8EB}\textbf{1.550} & \cellcolor[HTML]{E0F8EB}\textbf{55.3\%}

\\ \bottomrule
\end{tabular}
}
\vskip -0.1in
\end{table}

\section{Discussion and Limitations}
\label{sec:discussion}

\subsection{One Alignment Functional Across Corpora, Documents, and Prompts}
\label{sec:dynamics}

Maximizing next-token likelihood minimizes KL divergence to the empirical training distribution. To first order, this aligns the model's word predictions on $\mathcal{V}_b$ with the corpus word distribution, while higher-order conditional structure is learned later and only partially, especially on the long tail~\citep{Nguyen24,Ge25,kandpal23_longtail}. The descriptive result of Section~\ref{main:main_result} is therefore a structural consequence of this ordering and reflects the optimization geometry shared by likelihood training and likelihood scoring rather than properties specific to the scales or corpora we examine. These results suggest that likelihood-based zero-shot performance partly reflects corpus–benchmark lexical alignment, alongside the capabilities each benchmark is intended to assess. 
%To the extent that the model's conditional structure adds little beyond the corpus word distribution, the zero-shot score approaches a measure of word overlap, and benchmark accuracy partly reflects corpus--benchmark lexical alignment alongside the capability it nominally targets.

The three uses of UCE in this paper are three instantiations of the same $D_{\mathrm{KL}}(Q\,\|\,P)$ functional at different scopes. Corpus-level ranking (Section~\ref{sec:corpus_rank}) evaluates $D_{\mathrm{KL}}$ once per (corpus, benchmark) pair to compare web-scale training corpora. Row-local data selection (Section~\ref{sec:data_selection}) evaluates the marginal contribution of each row to $D_{\mathrm{KL}}$ and removes the rows whose removal most lowers it. Per-prompt drafter routing (Section~\ref{sec:routing}) evaluates $D_{\mathrm{KL}}$ between each prompt's word distribution and each drafter's corpus word distribution, and assigns the prompt to the drafter with the lowest $D_{\mathrm{KL}}$.

\subsection{Contamination as an Alternative Explanation}
\label{sec:contamination}

Benchmark text inside a pre-training corpus could raise lexical alignment and accuracy together. We audit samples of all four corpora with a GPT-3-style word-level $13$-gram index~\citep{brown20_gpt3} (Table~\ref{tab:contam_items}). Nine of the $12$ benchmarks have under $0.3\%$ of items matched in every sample, and the three exceptions, BoolQ, SciQ, and MMLU, peak on FineWeb-Edu. On SWAG, SocialIQA, and LAMBADA, where at most $0.01\%$ of items match on C4, DCLM, or FineWeb-Edu, a corpus with lower UCE still scores better, so contamination is unlikely to explain the relationship there. Among the three exceptions, BoolQ supplies its passage at evaluation time, so a match on that passage cannot carry the yes or no label, and the audit does not rule out a contribution on SciQ and MMLU, where FineWeb-Edu has both the most matched items and the highest score. %In the selection study, the UCE-filter's gain over the random control survives contamination removal at both the evaluation and the training end (Section~\ref{sec:ce_filter_results}, Appendix~\ref{apdx:contamination}). 

\subsection{What Unigram Cross-Entropy Captures and Where It Fails}
UCE measures only the alignment of word distributions. It is by construction blind to conditional structure such as syntax, multi-token composition, and factual co-occurrence, so UCE is best understood as a lower-bound proxy for distributional alignment rather than a sufficient summary of corpus quality or a fundamental property of the models. UCE does not order grammar (BLiMP~\citep{BLiMP}), arithmetic (MathQA~\citep{mathqa}), or non-English (Global-PIQA~\citep{global-piqa}) benchmarks (Appendix~\ref{apdx:uce_boundary}). White\-space tokenization is a second boundary, since UCE applies most directly to whitespace-segmented languages and leaves non-segmented scripts (e.g., Chinese, Japanese) and highly agglutinative morphologies for future work.

Apart from the routing study, the $30\mathrm{B}$ case study, and $5$-shot checks, every claim concerns base-model zero-shot behavior under a controlled from-scratch protocol, and the selection gains are an upper bound under a known target (Section~\ref{sec:ce_filter_results}). Two directions follow directly from our design, a paraphrase protocol that tests how far the observed relationship depends on exact lexical realization, and a lexical decomposition that restricts UCE to a benchmark-characteristic vocabulary. Whether the lexical imprint persists through SFT and RLHF, which differ from pre-training in data volume, training objective, and what is measured, requires a controlled study built for those objectives. %A general account of what makes a model generalize would need a design that varies the other properties of pre-training data along with this lexical one.

\begin{table}[t]
\vskip -0.1in
\small
\centering
\caption{Share of benchmark items (\%, all splits) with at least one word-level $13$-gram found in a sample of up to one million documents per corpus.}
\label{tab:contam_items}
\setlength{\tabcolsep}{2pt}
\resizebox{\textwidth}{!}{%
\begin{tabular}{@{}lcccccccccccc@{}}
\toprule
 & \textbf{ARC-Challenge} & \textbf{ARC-Easy} & \textbf{BoolQ} & \textbf{CommonsenseQA} & \textbf{HellaSwag} & \textbf{LAMBADA} & \textbf{MMLU} & \textbf{OpenBookQA} & \textbf{PIQA} & \textbf{SciQ} & \textbf{SocialIQA} & \textbf{SWAG} \\
\midrule
DCLM & 0.08 & 0.02 & 3.07 & 0.00 & 0.11 & 0.00 & 0.64 & 0.00 & 0.12 & 0.64 & 0.00 & 0.00 \\
FineWeb-Edu & 0.04 & 0.06 & 11.25 & 0.00 & 0.29 & 0.00 & 1.39 & 0.02 & 0.12 & 4.91 & 0.00 & 0.00 \\
C4 & 0.00 & 0.02 & 3.14 & 0.00 & 0.12 & 0.00 & 0.37 & 0.02 & 0.07 & 0.25 & 0.01 & 0.00 \\
OpenWebText & 0.00 & 0.02 & 5.35 & 0.00 & 0.02 & 0.06 & 0.58 & 0.00 & 0.02 & 0.15 & 0.00 & 0.00 \\
\bottomrule
% [provenance 2026-07-26] rebuttal/results/t1_8_contamination.md, item-side 13-gram index (moved from Appendix F on 2026-10-21).
\end{tabular}}
\vskip -0.1in
\end{table}

% \subsection{Implications for Benchmark Design and Data Curation}
% For benchmark design, the descriptive trend (Section~\ref{main:main_result}) means that many widely used zero-shot benchmarks are only weakly out-of-distribution relative to web-scale pre-training corpora, so a benchmark whose results are largely explained by UCE reports corpus alignment more than the capability it nominally targets. We take this not as a critique of these benchmarks but as a reason to pair scores with alignment diagnostics when comparing models trained on different corpora. For data curation, UCE is a cheap, benchmark-aware signal that can be computed once per corpus and reused across selection or routing decisions, and we expect similar use in deduplication, mixture weighting, and quality-aware sampling beyond the UCE-filter (Section~\ref{sec:data_selection}) and inference-time routing (Section~\ref{sec:routing}) instantiations of this paper.

\section{Conclusion}

Pre-trained language-model behavior is shaped by many intertwined factors, including higher-order conditional structure, scale, and post-training reshaping. This paper takes up one of them, the lexical imprint that the pre-training corpus leaves on zero-shot behavior, as captured by word-level unigram cross-entropy. Its descriptive force comes from a single mechanism. Likelihood training allocates probability mass according to the pre-training corpus, and likelihood-based zero-shot evaluation partly reflects the lexical alignment between the pre-training corpus and the benchmark target. 

This descriptive signal becomes the same statistic at finer scopes. Applied as document-level data selection, it controls the lexical imprint, and this control shows it more clearly than comparing corpora can, since accuracy rises mainly on the target whose alignment is raised. Applied as prompt-level drafter routing, it raises speculative-decoding throughput. Across these uses, the analytic relationship between pre-training corpus, benchmark target, and language-model behavior is measured rather than only described. The same logic marks the limits of the account. 
UCE ignores syntax, multi-token composition and the effects of post-training. It makes one part of the imprint measurable, namely how the lexical statistics of pre-training data align with those of the target data. 

We close with:
\begin{quote}
\centering
\emph{Pre-training pulls a model’s predictive distribution toward its corpus’s empirical conditional patterns, leaving a measurable lexical imprint on zero-shot behavior.}
\end{quote}

\clearpage

\section*{Acknowledgments}
% …\end{ack}
This work was partially supported by the Institute for Information \& Communications Technology Planning \& Evaluation (IITP) grant funded by the Korean government (MSIT) (RS-2019-II190075, Artificial Intelligence Graduate School Program(KAIST) and No. RS-2022-II220713, Meta-learning Applicable to Real-world Problems).

% NeurIPS 2026 default uses numeric citations (\PassOptionsToPackage{numbers, compress}{natbib} above).
% unsrtnat is a numeric, sort-by-citation natbib-compatible style; this matches
% what the natbib options were set to.
\bibliographystyle{unsrtnat}
\bibliography{cite.bib}

%%%%%%%%%%%%%%%%%%%%%%%%%%%%%%%%%%%%%%%%%%%%%%%%%%%%%%%%%%%%

\newpage
\appendix
\onecolumn

% 2026-10-08: appendix title at the top of the first appendix page (author request)
% \begin{center}
% {\LARGE\bfseries APPENDIX\par}
{\LARGE\bfseries Appendix\par}
% \end{center}
\vspace{0.5em}

\section{Unigram Cross-Entropy Measures Marginal Mismatch}
\label{apdx:unigram_decomposition}

Let \(A\) and \(B\) denote two text corpora modeled as realizations of
stationary and ergodic sequence distributions \(P_A\) and \(P_B\) on
\(\mathcal{V}^{\mathbb{N}}\), where \(\mathcal{V}\) is the vocabulary.
Define the stationary single-token marginals
\begin{equation}
p_A(x)
:=
P_A(X_k=x),
\qquad
p_B(x)
:=
P_B(X_k=x),
\qquad x\in\mathcal{V}.
\label{eq:app_unigram_marginals}
\end{equation}
By stationarity, these marginals do not depend on \(k\). When
\(\mathcal{V}\) is finite, \(p_A\) and \(p_B\) are categorical
distributions over the vocabulary.

The unigram cross-entropy of the target marginal \(p_B\) relative to
the reference marginal \(p_A\) is
\begin{equation}
\begin{aligned}
H(p_B,p_A)
&:=
\mathbb{E}_{X\sim p_B}
\left[-\log p_A(X)\right] \\
&=
-\sum_{x\in\mathcal{V}}
p_B(x)\log p_A(x).
\end{aligned}
\label{eq:app_unigram_ce}
\end{equation}

We assume that \(p_B\) is absolutely continuous with respect to
\(p_A\), i.e.,
\begin{equation}
p_B(x)>0
\quad\Longrightarrow\quad
p_A(x)>0.
\end{equation}
Otherwise, both the cross-entropy and the corresponding KL divergence
are infinite. In finite-sample computation, this support condition can
be enforced through smoothing.

Adding and subtracting
\(-\sum_{x\in\mathcal{V}}p_B(x)\log p_B(x)\) yields
\begin{equation}
H(p_B,p_A)
=
H(p_B)
+
D_{\mathrm{KL}}(p_B\Vert p_A),
\label{eq:app_unigram_decomposition}
\end{equation}
where
\begin{equation}
H(p_B)
:=
-\sum_{x\in\mathcal{V}}
p_B(x)\log p_B(x)
\end{equation}
is the marginal entropy of the target distribution, and
\begin{equation}
D_{\mathrm{KL}}(p_B\Vert p_A)
:=
\sum_{x\in\mathcal{V}}
p_B(x)
\log
\frac{p_B(x)}{p_A(x)}
\end{equation}
is the KL divergence from the target marginal to the reference
marginal. For a fixed target corpus \(B\), the marginal entropy \(H(p_B)\) is
constant across reference corpora \(A\). Ranking reference corpora by
unigram cross-entropy is therefore equivalent to ranking them by
\(D_{\mathrm{KL}}(p_B\Vert p_A)\). Across the eleven evaluation benchmarks, however, \(H(p_B)\) itself spans \(1.86\) bits, so absolute UCE values are not comparable between benchmarks and every comparison in this paper is within a benchmark across corpora.
%The converse direction does not enjoy this cancellation. When the reference corpus is fixed and the target varies, the entropy term varies across comparisons, and across the eleven evaluation benchmarks \(H(p_B)\) spans \(1.86\) bits, from \(9.16\) on SocialIQA to \(11.02\) on BoolQ, several times the alignment differences interpreted in this paper. An absolute UCE difference between two benchmarks is therefore not interpretable on its own, and every comparison in this paper is within a benchmark across corpora.
% [provenance 2026-08-11] per-task H(Q) recomputable as own_loo - kl from
% rebuttal/results/addexp/pertask_loo.json and kl_loo_matrix.json. 

\subsection{UCE computation details}
\label{apdx:uce_implementation}

All cross-entropy and KL divergence values in the paper, including the UCE defined in Section~\ref{main:metric}, are reported in bits, with logarithms taken base $2$.

\paragraph{Tokenization.}
Words are obtained by Unicode NFKC normalization of the raw text, followed by punctuation stripping and whitespace splitting. We do not lowercase, since case-bearing words such as proper nouns carry distributional information relevant to many benchmarks. The benchmark vocabulary $\mathcal{V}_b$ is the set of distinct words obtained from the benchmark dataset under the same procedure, and word-frequency distributions $P_{\text{pre}}$ and $P_{\text{eval}}$ are computed as relative frequencies of $w \in \mathcal{V}_b$ in the pre-training corpus and the benchmark dataset respectively.

\paragraph{Zero count smoothing.}
For words $w \in \mathcal{V}_b$ that appear in the benchmark but not in the pre-training corpus, the empirical probability $P_{\text{pre}}(w)$ would be zero and the cross-entropy in Section~\ref{main:metric} would diverge. We therefore apply zero-count smoothing to $P_{\text{pre}}$ over $\mathcal{V}_b$, assigning a unit count only to $w$ absent from the pre-training corpus:
\begin{equation}
\tilde{c}(w) = \max\{c(w),1\},
\qquad
P_{\text{pre}}(w) = \frac{\tilde{c}(w)}{T+B},
\end{equation}
where $c(w)$ is the raw count of $w$ in the pre-training corpus, $T$ is the total word-token count of the pre-training corpus, and $B=\sum_{w\in\mathcal{V}_b}\mathbf{1}[c(w)=0]$ is the number of unseen benchmark-vocabulary types. Unlike Laplace smoothing, this leaves all positive counts unchanged. Because nearly all benchmark-vocabulary types are observed in the pre-training corpus, flooring only zero counts perturbs the empirical training distribution less than Laplace smoothing. The very small probability mass of $w \in B$ shown in Appendix~\ref{apdx:word_overlap_correlation} further suggests that zero-count smoothing is more appropriate than Laplace smoothing for measuring empirical distributional alignment between the pre-training corpus and benchmark corpora. The same smoothing is applied to all corpora and scales reported in Sections~\ref{main:main_result}--\ref{sec:routing}.

\section{Related Work}
\label{apdx:related}

\paragraph{Lexical traces of pre-training data.}
A line of work has linked lexical signals to language-model behavior at increasingly coarse granularities. \citet{razeghi_2022} link item-level term frequency to few-shot numerical reasoning, \citet{kandpal23_longtail} extend this to entity-level long-tail frequency in QA, \citet{wang25_distmem} formalize distributional memorization between LLM output probabilities and per-task $n$-gram counts, and \citet{ruzzetti26_lexical} relate test-instance lexical popularity to performance. Closest in spirit, \citet{wu25_benchsig} mine salient tokens from in-the-wild web corpora whose token-level perplexity, taken over a panel of $32$ pretrained models, predicts performance on $89$ benchmarks. Across all of these, the lexical signal is measured on pretrained checkpoints, and the corpus identity that drives it is entangled with training recipe, scale, tokenizer, and training history. We instead vary corpus identity under a fixed from-scratch recipe (Section~\ref{main:setting}) and intervene on document-level alignment (Section~\ref{sec:data_selection}).

\paragraph{Compression and Markov approximations.}
Section~\ref{sec:motivation} discusses work that frames language modeling as compression of the training distribution \citep{deletang24,depeiges25} and likelihood training as estimation of the empirical data distribution \citep{chung_vfi_2025}, together with work relating neural language models to $n$-gram statistics \citep{Nguyen24,Svete24,Ge25}.

\paragraph{Data selection.}
Aligning training data with a downstream target has a long history. \citet{klakow00} trains a unigram language model on the full non-domain corpus and scores each candidate segment by how removing it would change the in-domain data log-likelihood under that model.
\citet{moore_lewis_2010} introduce the cross-entropy difference filter, while \citet{Axelrod17} and \citet{Feng22} develop cynical data selection, a greedy method that chooses sentences by how much adding each one improves the language-model likelihood of the target-domain data. More recently, \citet{xie_dsir_2023} scale importance resampling using hashed $n$-gram features, and \citet{thrush25_perpcorr} use the correlation of per-domain LLM perplexity with downstream accuracy across $90$ LMs as a selection signal. \citet{datadecide} use small-scale proxy models to predict which pre-training data yields the best models at larger scale across downstream benchmarks. The UCE-filter shares this goal but operationalizes alignment as a single information-theoretic functional requiring only word counts, with the same statistic that drives our descriptive analysis (Section~\ref{main:main_result}) reused as the row-local selection signal (Section~\ref{sec:ce_filter}). Appendix~\ref{sec:ce_family_baselines} compares the UCE-filter with the embedding-based selector BETR~\citep{mizrahi25_betr} under our from-scratch grid.

\paragraph{Benchmark contamination.}
Another line of work treats the corpus--evaluation interface as a contamination problem at the level of exact or near-exact overlap \citep{sainz23}, supported by large-scale corpus auditing \citep{elazar24_wimbd}. UCE complements contamination detection by summarizing the interface at the coarser level of probability mass on benchmark words rather than exact-match overlap. Appendix~\ref{apdx:contamination} reports our own $13$-gram audit of the four corpora, with contaminated items and documents removed at the evaluation and training ends.

\paragraph{Speculative decoding.}
Section~\ref{sec:routing} routes between corpus-specialized drafters using unigram cross-entropy. The base setup follows \citet{leviathan23} and \citet{chen23_specdec}. Among learned routing policies, \citet{banditspec_2025} use bandit-based drafter selection and \citet{specrouter_2025} adapt across multiple drafter levels, while \citet{hong25_domain_drafter} report best practices for training domain-specialized drafters. A separate model-free direction retrieves candidates from non-parametric text statistics \citep{rest_2024, stand_2025}. UCE-routing keeps the parametric drafter pool fixed and selects among them via a precomputed corpus-level unigram distribution, with no per-prompt training, online learning, or retrieval at inference.

\section{Setup} \label{Apdx:setting}

The corpus-ranking study of Section~\ref{main:main_result} (\S\ref{apdx:setting:corpus}) and the UCE-filter data-selection study of Section~\ref{sec:ce_filter_results} (\S\ref{apdx:setting:cefilter}) use different model sizes, learning-rate schedules, and token budgets, so we report their configurations separately.

\subsection{Corpus-ranking experiments (Section~\ref{main:main_result})}
\label{apdx:setting:corpus}

Table~\ref{tab:train-config} gives the training recipe and Table~\ref{tab:model-config} the three LLaMA-architecture~\citep{llama2} model sizes.

\begin{table}[!ht]
  \centering
  \caption{Training configurations for the corpus-ranking experiments (Section~\ref{main:main_result}). All runs use this recipe across the three model scales of Table~\ref{tab:model-config}.}
  \label{tab:train-config}

  %---- Row 1: optimizer ----
  \begin{tabular}{lcccc}
    \toprule
    \textbf{Optimizer} & $\beta_1$ & $\beta_2$ & $\epsilon$ & \textbf{Weight Decay} \\
    \midrule
    AdamW & 0.9 & 0.999 & $10^{-8}$ & 0.1 \\
    \bottomrule
  \end{tabular}

  \vspace{0.4em}

  %---- Row 2: schedule ----
  \begin{tabular}{lccc}
    \toprule
    \textbf{Peak LR} & \textbf{Schedule} & \textbf{Warmup} & \textbf{Weight Init.} \\
    \midrule
    $3\!\times\!10^{-4}$ & Cosine & 700 steps & 0.02 \\
    \bottomrule
  \end{tabular}

  \vspace{0.4em}

  %---- Row 3: batching / hardware ----
  \begin{tabular}{lcccc}
    \toprule
    \textbf{Global Batch} & \textbf{Per-device Batch} & \textbf{Devices} & \textbf{Sequence Length} & \textbf{Seed} \\
    \midrule
    256 & 32 & 8 GPUs & 2048 & 42 \\
    \bottomrule
  \end{tabular}
\end{table}

\begin{table}[!ht]
  \centering
  \caption{Model configurations for the corpus-ranking experiments (Section~\ref{main:main_result}). Sizes count non-embedding parameters.}
  \label{tab:model-config}
  \begin{tabular}{lccccc}
    \toprule
    \textbf{Size} & \(n_{\text{layers}}\) & \(n_{\text{heads}}\) & \(d_{\text{model}}\) & \(d_{\text{ffn}}\) & \(d_{\text{head}}\) \\
    \midrule
    402M & 24 & 8  & 1024 & 4096 & 128 \\
    1.33B & 30 & 30 & 1920 & 5120 & 64 \\
    3.36B & 32 & 20 & 2560 & 10240 & 128 \\
    \bottomrule
  \end{tabular}
\end{table}

\paragraph{Evaluation metric.}
For LAMBADA, the corpus-ranking tables in Appendix~\ref{apdx:word_overlap_correlation} report perplexity (lower better), while the UCE-filter experiment of Section~\ref{sec:ce_filter_results} uses last-word accuracy (higher better, see Table~\ref{tab:ce_filter_csr7_full}). Last-word accuracy is too noisy to separate corpora reliably on the $400\mathrm{M}$--$3.36\mathrm{B}$ corpus-ranking grid but informative on the $1.34\mathrm{B}$--$3.33\mathrm{B}$ UCE-filter grid trained on $70\mathrm{B}$ tokens. All other benchmarks use accuracy throughout. For the selection experiments of Section~\ref{sec:ce_filter_results} each task is scored with its primary lm-evaluation-harness metric, length-normalized accuracy for ARC-Easy/Challenge, HellaSwag, OpenBookQA, and PIQA, and plain accuracy for SciQ, SWAG, LAMBADA, CommonsenseQA, BoolQ, and SocialIQA.
% [provenance 2026-07-23] Metric map validated by reproducing all 24 published
% Targeted/Overall means of tab:ce_filter_csr7_main exactly from per-seed lm-eval
% results (eval_results/csr7-{1b,3b}-{dclm,fwedu}-{rand,dsirp,f80}[-s42|-s7777]-iter133514).
% NOTE seed-1234 dirs: social_iqa lives in a later supplementary results*.json in the
% same dir (the 72-task file lacks it) -> always MERGE all results*.json per dir.

\subsection{UCE-filter experiments (Section~\ref{sec:ce_filter_results})}
\label{apdx:setting:cefilter}

We follow the architectural patterns of DataComp-LM~\citep{datacomp}. The two scales (Table~\ref{tab:cefilter-arch}) are $1.34\mathrm{B}$ parameters ($1.23\mathrm{B}$ non-embedding), following the DCLM ``$1$B-$1$x'' specification, and $3.33\mathrm{B}$ parameters ($3.17\mathrm{B}$ non-embedding), following the DCLM $7$B pattern scaled to $28$ layers. The four selection methods (the UCE-filter, Random, and the DSIR-paper and DSIR-LN variants of Appendix~\ref{sec:dsir_baseline}) train on $70\mathrm{B}$ GPT-2 tokens under the recipe of Table~\ref{tab:cefilter-config}, with gradient clipping at $1.0$ and a distributed optimizer, and each model is evaluated zero-shot with lm-evaluation-harness using each task's primary metric.

\paragraph{Out-of-target tasks and MMLU omission.}
Compared with the $10$ corpus-ranking tasks of Section~\ref{main:setting}, the UCE-filter evaluation suite omits MMLU and adds CommonsenseQA and BoolQ. We omit MMLU because all selection methods cluster within $\pm 1$~pp of chance (around $25\%$) at the $1.34\mathrm{B}$ and $3.33\mathrm{B}$ scales with a $70\mathrm{B}$-token budget, so the task does not separate methods (raw MMLU scores are reported in the supplementary material). We add BoolQ, a yes/no calibration task plausibly orthogonal to the targeted-suite vocabulary, as a stronger test of whether UCE-filtering's gains reflect targeted vocabulary alignment rather than a generic improvement. The four out-of-target tasks (LAMBADA, CommonsenseQA, BoolQ, SocialIQA) are reported alongside the targeted-suite average in every (Source, Scale, Method) cell. SocialIQA is evaluated via the \texttt{lighteval/siqa} parquet mirror, which has the same schema as the canonical \texttt{allenai/social\_i\_qa} release.

\paragraph{Selection-pool sizes and training-exposure ratios.}
On DCLM, the UCE-filter selection contains roughly $227\mathrm{B}$ tokens, the row-count-matched random sample $155\mathrm{B}$, DSIR-LN $105\mathrm{B}$, and DSIR-paper $106\mathrm{B}$. On FineWeb-Edu, the four pools are roughly $90\mathrm{B}$, $70\mathrm{B}$, $86\mathrm{B}$, and $101\mathrm{B}$ respectively. All four methods then train for the same $70\mathrm{B}$-token budget, with the data loader sampling uniformly from each pool, so the exposure ratio (training tokens divided by pool tokens) varies across methods. On DCLM the UCE-filter trains on roughly $31\%$ of its selected pool while DSIR-LN trains on roughly $66\%$ of its own, and on FineWeb-Edu the corresponding ratios are $78\%$ and $81\%$. The downstream-accuracy lead of the UCE-filter (Section~\ref{sec:ce_filter_results}, Table~\ref{tab:ce_filter_csr7_main}) is therefore not an artifact of a larger selection pool.

\paragraph{Noise calibration.}
Training seeds are $1234$, $42$, and $7777$ for the UCE-filter, Random, and DSIR-paper arms and $1234$ for the DSIR-LN ablation, with per-method standard deviations and Welch tests in Appendix~\ref{apdx:seed_sweep}.

\begin{table}[!ht]
  \centering
  \caption{Training configurations for the UCE-filter experiments (Section~\ref{sec:ce_filter_results}). All sixteen runs (Source $\times$ Scale $\times$ \{UCE-filter, Random, DSIR-paper, DSIR-LN\}) use this recipe.}
  \label{tab:cefilter-config}

  %---- Row 1: optimizer ----
  \begin{tabular}{lcccc}
    \toprule
    \textbf{Optimizer} & \(\beta_1\) & \(\beta_2\) & \(\epsilon\) & \textbf{Weight Decay} \\
    \midrule
    AdamW & 0.9 & 0.95 & $10^{-8}$ & 0.033 \\
    \bottomrule
  \end{tabular}

  \vspace{0.4em}

  %---- Row 2: schedule ----
  \begin{tabular}{lccc}
    \toprule
    \textbf{Peak LR} & \textbf{Min LR} & \textbf{Schedule} & \textbf{Warmup} \\
    \midrule
    $1\!\times\!10^{-3}$ & $1\!\times\!10^{-5}$ & Cosine & 5000 iters \\
    \bottomrule
  \end{tabular}

  \vspace{0.4em}

  %---- Row 3: batch / tokens ----
  \begin{tabular}{lcccc}
    \toprule
    \textbf{Global Batch} & \textbf{Sequence Length} & \textbf{Tokens} & \textbf{Precision} & \textbf{Hardware} \\
    \midrule
    256 & 2048 & $70\mathrm{B}$ & BF16 & $16$ B200 (2~nodes) \\
    \bottomrule
  \end{tabular}
\end{table}

\begin{table}[!ht]
  \centering
  \caption{Model architectures for the UCE-filter experiments (Section~\ref{sec:ce_filter_results}). DataComp-LM patterns~\citep{datacomp}. All use SwiGLU activations, RMSNorm, RoPE positions, and tied input/output embeddings.}
  \label{tab:cefilter-arch}
  \begin{tabular}{lcccccl}
    \toprule
    \textbf{Size} & \(n_{\text{layers}}\) & \(n_{\text{heads}}\) & \(d_{\text{model}}\) & \(d_{\text{ffn}}\) & \(d_{\text{head}}\) & \textbf{Pattern} \\
    \midrule
    1.34B & 24 & 16 & 2048 & 5632 & 128 & DCLM 1B-1x \\
    3.33B & 28 & 24 & 3072 & 8192 & 128 & DCLM 7B (scaled to 28 layers) \\
    \bottomrule
  \end{tabular}
\end{table}

\newpage

\section{Corpus-level UCE analyses}\label{apdx:corpus_analyses}

Section~\ref{main:main_result} ranks corpora by aggregate UCE and ties this ranking to downstream accuracy. The analyses below test that corpus-level finding and mark where the statistic's scope ends. The ranking is robust along the axes they examine, and Appendix~\ref{apdx:uce_oov_contribution} shows in particular that the in-vocabulary distribution rather than the smoothing tail carries the cross-corpus signal.

\subsection{Coverage and mass allocation across benchmarks}\label{apdx:coverage_decomp}

To probe the mechanism behind the inverse cross-entropy/accuracy relation in Section~\ref{main:main_result} (Figure~\ref{fig:figure1}), we decompose the corpus--benchmark overlap into \emph{vocabulary coverage} (the fraction of the benchmark vocabulary that the corpus has observed at all) and \emph{probability-mass allocation} (the share of benchmark word mass attributable to top-percentile training words versus the long tail). Table~\ref{tab:pertask_1b} reports both at the $1.33\mathrm{B}$/$26\mathrm{B}$-token operating point for every task. Coverage is essentially saturated on every task ($\sim\!99.9\%$ Seen on every corpus), so cross-corpus differences cannot be explained by coverage alone. Mass allocation, in contrast, varies systematically with cross-entropy and tracks the per-task ordering of zero-shot accuracy on ARC Easy, HellaSwag, and MMLU.

\subsection{Out-of-vocabulary contribution to UCE}\label{apdx:uce_oov_contribution}

The coverage analysis above shows that DCLM and FineWeb-Edu both place virtually all of the benchmark vocabulary inside their support (Seen $\approx 99.9\%$), so cross-corpus differences in UCE cannot be attributed to which words are observed at all. This raises the question of whether the residual differences come from the in-vocabulary distribution or from the zero-count smoothing tail introduced in Appendix~\ref{apdx:uce_implementation}, which assigns a unit count to benchmark words that the pre-training corpus never produced. Because these words still enter UCE through their $\log P_{\text{pre}}(w)$ term, we isolate their contribution by splitting the benchmark vocabulary into the covered subset $\mathcal{V}_b^{\text{cov}} = \mathcal{V}_b \cap \{w : c(w) > 0\}$ and its complement $\mathcal{V}_b^{\text{unc}} = \mathcal{V}_b \setminus \mathcal{V}_b^{\text{cov}}$, so that
\begin{equation}
H\!\left(P_{\text{eval}}, P_{\text{pre}}\right)
\;=\; \underbrace{-\!\!\!\sum_{w \,\in\, \mathcal{V}_b^{\text{cov}}} P_{\text{eval}}(w)\, \log P_{\text{pre}}(w)}_{\text{covered contribution}} \;+\; \underbrace{-\!\!\!\sum_{w \,\in\, \mathcal{V}_b^{\text{unc}}} P_{\text{eval}}(w)\, \log P_{\text{pre}}(w)}_{\text{uncovered contribution}}.
\label{eq:uce_decomp}
\end{equation}

Under zero-count smoothing, the uncovered contribution collapses to
$m_{\text{unc}} \cdot \log(T+B)$, where
$m_{\text{unc}} = \sum_{w \in \mathcal{V}_b^{\text{unc}}} P_{\text{eval}}(w)$
is the benchmark mass on uncovered words,
$T$ is the total word-token count of the
pre-training corpus, and
$B = \sum_{w \in \mathcal{V}_b}\mathbf{1}[c(w)=0]
= |\mathcal{V}_b^{\text{unc}}|$ is the number of unseen
benchmark vocabulary types.

Table~\ref{tab:uce_decomposition} reports the two components for DCLM and FineWeb-Edu against the CSR7 and ten-task corpus-ranking groupings, measured on a uniform five-percent fragment sample of each source corpus (roughly $30.5\mathrm{B}$ DCLM tokens and $13.7\mathrm{B}$ FineWeb-Edu tokens). In every cell the covered subset captures more than $99.95\%$ of the benchmark mass, leaving $m_{\text{unc}}$ below $5 \times 10^{-4}$ and an uncovered contribution between $0.0095$ and $0.0151$ bits, which is less than $0.14\%$ of the UCE value. Subtracting the uncovered term from each cell preserves both orderings of the main results, with DCLM below FineWeb-Edu and CSR7 below the corpus-ranking grouping. The cross-corpus and cross-benchmark structure of UCE is therefore a property of the in-vocabulary distribution rather than an artifact of the smoothing tail.

\begin{table*}[!ht]
\centering
\caption{UCE decomposition by coverage (bits $\downarrow$) for DCLM and FineWeb-Edu against the CSR7 and ten-task corpus-ranking benchmark groupings, measured on a five-percent fragment sample of each source corpus. The benchmark mass on uncovered words $m_{\text{unc}}$ and its contribution to UCE are computed from Equation~\eqref{eq:uce_decomp}.}
\label{tab:uce_decomposition}
\setlength{\tabcolsep}{6pt}
\small
\resizebox{\textwidth}{!}{%
\begin{tabular}{@{}llcccccc@{}}
\toprule
\textbf{Source} & \textbf{Benchmark} &
\textbf{UCE} & \textbf{Type cov.} & \textbf{Mass cov.} & $m_{\text{unc}}$ & \textbf{Uncov.\ contrib.} & \textbf{Share of UCE} \\
\midrule
DCLM         & CSR7            & $11.1665$ & $95.54\%$ & $99.973\%$ & $2.73 \times 10^{-4}$ & $0.0095$ & $0.085\%$ \\
DCLM         & Corpus-ranking  & $11.1843$ & $95.83\%$ & $99.972\%$ & $2.80 \times 10^{-4}$ & $0.0098$ & $0.087\%$ \\
FineWeb-Edu  & CSR7            & $11.3256$ & $93.88\%$ & $99.957\%$ & $4.32 \times 10^{-4}$ & $0.0146$ & $0.128\%$ \\
FineWeb-Edu  & Corpus-ranking  & $11.3572$ & $94.12\%$ & $99.955\%$ & $4.50 \times 10^{-4}$ & $0.0151$ & $0.133\%$ \\
\bottomrule
\end{tabular}}
\end{table*}

\subsection{Robustness of UCE to preprocessing choices}\label{apdx:uce_preprocessing}

The implementation specified in Section~\ref{main:metric} and Appendix~\ref{apdx:uce_implementation} combines NFKC normalization, punctuation stripping, whitespace tokenization without case folding, and zero-count smoothing of $P_{\text{pre}}$ over $\mathcal{V}_b^{\text{unc}}$, where each word $w \in \mathcal{V}_b^{\text{unc}}$ is assigned a unit count. To check that the DCLM-versus-FineWeb-Edu ranking reported in Section~\ref{main:main_result} is not driven by these defaults, each component is perturbed in turn while the rest of the implementation is held fixed. UCE is recomputed on a uniform one-percent fragment sample of the two pre-training corpora (roughly $6.2\mathrm{B}$ DCLM tokens and $2.9\mathrm{B}$ FineWeb-Edu tokens) and evaluated against both the seven-task targeted grouping (CSR7) of Section~\ref{sec:ce_filter_results} and the ten-task corpus-ranking grouping of Section~\ref{main:main_result}. The perturbations cover case folding before counting, Lidstone smoothing at concentration $1/2$, unsmoothed raw vocabulary counts, and an inverse-document-frequency reweighting of the per-word contribution.\footnote{Lemmatization and stopword removal would be informative additional axes but require lexical resources external to the measurement pipeline and are not considered here.}

Table~\ref{tab:uce_preproc_ablation} collects UCE for the four corpus-by-grouping combinations under each variant, alongside the within-grouping gap $\Delta_{\text{Fw}-\text{DCLM}}$. FineWeb-Edu sits above DCLM in every cell of every variant. Under the default pipeline and its smoothing perturbations the gap lies between $0.16$ and $0.17$, with the two alternative smoothers (concentration $1/2$ and no smoothing) moving absolute UCE by less than $0.003$ in any cell. Case folding shifts all four UCE values downward by roughly $0.32$ as case variants collapse into shared types, leaving the gap intact. Inverse-document-frequency reweighting raises absolute UCE by about $3.2$ and widens the within-grouping gap to roughly $0.24$. The default configuration is used throughout the paper.

\begin{table*}[!ht]
\centering
\caption{UCE (bits $\downarrow$) under five preprocessing variants on the DCLM and FineWeb-Edu corpora (one-percent fragment sample), evaluated against the seven-task targeted grouping (CSR7) of Section~\ref{sec:ce_filter_results} and the ten-task corpus-ranking grouping of Section~\ref{main:main_result}. $\Delta_{\text{Fw}-\text{DCLM}}$ is the within-grouping FineWeb-Edu minus DCLM UCE.}
\label{tab:uce_preproc_ablation}
\setlength{\tabcolsep}{6pt}
\small
\resizebox{\textwidth}{!}{%
\begin{tabular}{@{}lcccccc@{}}
\toprule
\multirow{2}{*}{\textbf{Variant}} &
\multicolumn{2}{c}{\textbf{DCLM}} &
\multicolumn{2}{c}{\textbf{FineWeb-Edu}} &
\multicolumn{2}{c}{$\Delta_{\text{Fw}-\text{DCLM}}$} \\
\cmidrule(lr){2-3}\cmidrule(lr){4-5}\cmidrule(lr){6-7}
& CSR7 & Corpus-rank & CSR7 & Corpus-rank & CSR7 & Corpus-rank \\
\midrule
Baseline                               & $11.1650$ & $11.1832$ & $11.3215$ & $11.3539$ & $+0.1565$ & $+0.1707$ \\
Lowercase                              & $10.8486$ & $10.8652$ & $10.9952$ & $11.0246$ & $+0.1466$ & $+0.1594$ \\
Lidstone ($\alpha = \tfrac{1}{2}$)     & $11.1666$ & $11.1848$ & $11.3243$ & $11.3566$ & $+0.1577$ & $+0.1718$ \\
No smoothing                           & $11.1656$ & $11.1838$ & $11.3234$ & $11.3556$ & $+0.1578$ & $+0.1718$ \\
IDF weighting                          & $14.3153$ & $14.3110$ & $14.5502$ & $14.5539$ & $+0.2349$ & $+0.2429$ \\
\bottomrule
\end{tabular}}
\end{table*}

\newpage 

\subsection{Per-task benchmark accuracies and word overlap correlations}\label{apdx:word_overlap_correlation}

The tables below report per-task zero-shot accuracy at the three matched (model, subset) operating points of the corpus-level analysis ($3.36\mathrm{B}\,/\,60\mathrm{B}$, $1.33\mathrm{B}\,/\,26\mathrm{B}$, $402\mathrm{M}\,/\,8.5\mathrm{B}$), so that the per-corpus entropy-vs-accuracy tracking of Section~\ref{main:main_result} can be inspected task by task. Table~\ref{tab:pertask_1b} also reports the $402\mathrm{M}$ model trained on the same $26\mathrm{B}$-token subsets. OpenWebText, a single $8.5\mathrm{B}$-token set, is listed only with subset 1 (Table~\ref{tab:pertask_400m_s1}).

\begin{table*}[!ht]
\centering
\caption{Per-task word-frequency coverage, unigram cross-entropy (bits $\downarrow$), and zero-shot score for a $3.36\mathrm{B}$ model on $60\mathrm{B}$ tokens. LAMBADA is reported in perplexity (lower better) and other tasks in accuracy (higher better). The $\sim$80\% and $\sim$95\% columns report cumulative shares of benchmark word mass attributable to top-percentile training words, the 95--100\% column the share on the remaining tail words, and Seen is the fraction of benchmark vocabulary observed at least once. Source for Figure~\ref{fig:figure1}.}
\small
\setlength{\tabcolsep}{4pt}
\resizebox{\textwidth}{!}{%
\begin{tabular}{llcccccc}
\toprule
Task & Dataset &
$\sim$80\% &
$\sim$95\%&
95--100\% &
Seen &
Entropy &
Score \\
\midrule

\multirow{3}{*}{ARC-Easy}
& FineWeb-Edu & 83.97\% (150,599) & 98.15\% (176,031) & 1.84\% (3,298) & 99.99\% (179,329) & 11.1946 & 73.65 \\
& DCLM        & 78.54\% (140,870) & 97.24\% (174,411) & 2.75\% (4,926) & 99.99\% (179,337) & 11.5477 & 70.20 \\
& C4          & 77.74\% (139,428) & 96.63\% (173,311) & 3.35\% (6,003) & 99.98\% (179,314) & 11.6493 & 60.77 \\

\midrule
\multirow{3}{*}{ARC-Challenge}
& FineWeb-Edu & 85.04\% (93,417) & 98.16\% (107,823) & 1.82\% (1,997) & 99.97\% (109,820) & 11.0831 & 38.82 \\
& DCLM        & 80.32\% (88,236) & 97.45\% (107,051) & 2.54\% (2,791) & 99.99\% (109,842) & 11.3847 & 34.81 \\
& C4          & 79.88\% (87,745) & 97.05\% (106,610) & 2.89\% (3,180) & 99.95\% (109,790) & 11.4714 & 26.79 \\

\midrule
\multirow{3}{*}{HellaSwag}
& FineWeb-Edu & 82.70\% (6,900,957) & 96.31\% (8,035,886) & 3.64\% (304,022) & 99.95\% (8,339,908) & 11.2117 & 46.63 \\
& DCLM        & 82.74\% (6,903,940) & 96.90\% (8,085,503) & 3.06\% (255,401) & 99.96\% (8,340,904) & 11.0590 & 48.48 \\
& C4          & 83.68\% (6,981,914) & 96.99\% (8,093,304) & 2.96\% (246,704) & 99.95\% (8,340,008) & 11.0377 & 49.02 \\

\midrule
\multirow{3}{*}{MMLU}
& FineWeb-Edu & 80.37\% (953,768) & 95.99\% (1,139,104) & 3.97\% (47,072) & 99.96\% (1,186,176) & 11.3635 & 25.23 \\
& DCLM        & 78.64\% (933,162) & 95.95\% (1,138,626) & 4.01\% (47,596) & 99.96\% (1,186,222) & 11.4071 & 24.58 \\
& C4          & 78.45\% (930,923) & 95.09\% (1,128,357) & 4.85\% (57,534) & 99.94\% (1,185,891) & 11.5237 & 23.77 \\

\midrule
\multirow{3}{*}{SciQ}
& FineWeb-Edu & 80.28\% (960,226) & 95.74\% (1,145,183) & 4.22\% (50,508) & 99.97\% (1,195,691) & 11.4675 & 90.80 \\
& DCLM        & 74.95\% (896,531) & 94.13\% (1,125,843) & 5.83\% (69,791) & 99.96\% (1,195,634) & 11.8577 & 89.70 \\
& C4          & 73.69\% (881,360) & 92.98\% (1,112,130) & 6.93\% (82,930) & 99.91\% (1,195,060) & 12.0335 & 83.90 \\

\midrule
\multirow{3}{*}{OBQA}
& FineWeb-Edu & 79.76\% (105,348) & 96.69\% (127,702) & 3.30\% (4,356) & 99.99\% (132,058) & 11.5424 & 33.00 \\
& DCLM        & 77.40\% (102,225) & 96.51\% (127,467) & 3.48\% (4,596) & 99.99\% (132,063) & 11.6525 & 28.20 \\
& C4          & 77.21\% (101,974) & 95.97\% (126,746) & 4.01\% (5,302) & 99.98\% (132,048) & 11.7398 & 26.40 \\

\midrule
\multirow{3}{*}{PIQA}
& FineWeb-Edu & 74.89\% (707,350) & 95.56\% (902,571) & 4.42\% (41,744) & 99.98\% (944,315) & 11.6208 & 75.14 \\
& DCLM        & 75.43\% (712,414) & 96.67\% (912,988) & 3.32\% (31,403) & 99.99\% (944,391) & 11.4875 & 75.90 \\
& C4          & 77.37\% (730,709) & 97.25\% (918,478) & 2.73\% (25,819) & 99.98\% (944,297) & 11.3887 & 76.28 \\

\midrule
\multirow{3}{*}{LAMBADA}
& FineWeb-Edu & 81.12\% (256,695) & 94.54\% (299,132) & 5.32\% (16,840) & 99.86\% (315,972) & 11.5117 & 11.72 \\
& DCLM        & 83.40\% (263,900) & 96.60\% (305,663) & 3.31\% (10,468) & 99.91\% (316,131) & 10.9253 & 5.90 \\
& C4          & 82.45\% (260,885) & 95.76\% (303,016) & 4.14\% (13,091) & 99.90\% (316,107) & 11.1855 & 9.17 \\

\midrule
\multirow{3}{*}{Social IQA}
& FineWeb-Edu & 79.06\% (877,613) & 96.62\% (1,072,651) & 3.37\% (37,381) & 99.99\% (1,110,032) & 11.6220 & 42.02 \\
& DCLM        & 81.70\% (906,937) & 98.46\% (1,093,035) & 1.54\% (17,050) & 100.00\% (1,110,085) & 11.1622 & 42.73 \\
& C4          & 81.47\% (904,368) & 97.47\% (1,082,087) & 2.52\% (27,959) & 99.99\% (1,110,046) & 11.3183 & 42.27 \\

\midrule
\multirow{3}{*}{SWAG}
& FineWeb-Edu & 78.28\% (4,363,250) & 95.46\% (5,320,574) & 4.54\% (252,910) & 99.99\% (5,573,484) & 11.4145 & 49.37 \\
& DCLM        & 79.74\% (4,444,545) & 97.28\% (5,421,994) & 2.72\% (151,704) & 100.00\% (5,573,698) & 11.1036 & 52.82 \\
& C4          & 80.07\% (4,463,191) & 96.63\% (5,385,917) & 3.36\% (187,069) & 99.98\% (5,572,986) & 11.2432 & 50.91 \\

\bottomrule
\end{tabular}}
\label{tab:3B}
\end{table*}

\begin{table*}[!ht]
\centering
\caption{$1.33\mathrm{B}$ and $402\mathrm{M}$ models trained on the same $26\mathrm{B}$-token subsets. Conventions as in Table~\ref{tab:3B}. The coverage and entropy columns apply to both models.}
\label{tab:pertask_1b}
\small
\setlength{\tabcolsep}{4pt}
\resizebox{\textwidth}{!}{%
\begin{tabular}{llccccccc}
\toprule
Task & Dataset &
$\sim$80\% &
$\sim$95\% &
95--100\% &
Seen &
Entropy &
Score ($1.33\mathrm{B}$) &
Score ($402\mathrm{M}$) \\
\midrule

\multirow{3}{*}{ARC-Easy}
& FineWeb-Edu & 83.87\% (150,418) & 98.09\% (175,920) & 1.90\% (3,407) & 99.98\% (179,327) & 11.1921 & 65.87 & 62.42 \\
& DCLM        & 78.55\% (140,887) & 97.24\% (174,412) & 2.74\% (4,923) & 99.99\% (179,335) & 11.5461 & 63.17 & 56.78 \\
& C4          & 77.74\% (139,429) & 96.63\% (173,309) & 3.35\% (6,002) & 99.98\% (179,311) & 11.6491 & 54.55 & 50.38 \\

\midrule
\multirow{3}{*}{ARC-Challenge}
& FineWeb-Edu & 84.93\% (93,296) & 98.11\% (107,769) & 1.87\% (2,049) & 99.97\% (109,818) & 11.0807 & 31.83 & 28.24 \\
& DCLM        & 80.32\% (88,230) & 97.45\% (107,050) & 2.53\% (2,780) & 99.98\% (109,830) & 11.3833 & 27.65 & 26.19 \\
& C4          & 79.88\% (87,745) & 97.05\% (106,607) & 2.89\% (3,178) & 99.94\% (109,785) & 11.4708 & 24.49 & 20.99 \\

\midrule
\multirow{3}{*}{HellaSwag}
& FineWeb-Edu & 82.72\% (6,901,956) & 96.28\% (8,033,455) & 3.66\% (305,784) & 99.94\% (8,339,239) & 11.2045 & 40.21 & 36.82 \\
& DCLM        & 82.75\% (6,905,130) & 96.91\% (8,086,092) & 3.05\% (254,461) & 99.96\% (8,340,553) & 11.0568 & 40.75 & 37.17 \\
& C4          & 83.68\% (6,982,092) & 97.00\% (8,093,366) & 2.95\% (245,912) & 99.94\% (8,339,278) & 11.0367 & 41.29 & 37.45 \\

\midrule
\multirow{3}{*}{MMLU}
& FineWeb-Edu & 80.34\% (953,323) & 95.93\% (1,138,375) & 4.02\% (47,671) & 99.95\% (1,186,046) & 11.3649 & 25.89 & 25.07 \\
& DCLM        & 78.64\% (933,166) & 95.95\% (1,138,586) & 4.01\% (47,564) & 99.96\% (1,186,150) & 11.4063 & 24.96 & 23.84 \\
& C4          & 78.45\% (930,919) & 95.09\% (1,128,353) & 4.83\% (57,372) & 99.92\% (1,185,725) & 11.5230 & 23.39 & 23.44 \\

\midrule
\multirow{3}{*}{SciQ}
& FineWeb-Edu & 80.23\% (959,649) & 95.72\% (1,144,872) & 4.24\% (50,670) & 99.95\% (1,195,542) & 11.4639 & 86.10 & 83.80 \\
& DCLM        & 74.97\% (896,735) & 94.13\% (1,125,856) & 5.83\% (69,679) & 99.95\% (1,195,535) & 11.8554 & 85.00 & 82.40 \\
& C4          & 73.69\% (881,384) & 92.98\% (1,112,195) & 6.91\% (82,688) & 99.90\% (1,194,883) & 12.0327 & 79.30 & 75.20 \\

\midrule
\multirow{3}{*}{OBQA}
& FineWeb-Edu & 79.67\% (105,223) & 96.66\% (127,660) & 3.33\% (4,394) & 99.98\% (132,054) & 11.5388 & 26.60 & 24.20 \\
& DCLM        & 77.39\% (102,219) & 96.50\% (127,449) & 3.49\% (4,611) & 99.99\% (132,060) & 11.6511 & 22.60 & 20.20 \\
& C4          & 77.21\% (101,974) & 95.97\% (126,750) & 4.00\% (5,285) & 99.97\% (132,035) & 11.7396 & 20.00 & 19.80 \\

\midrule
\multirow{3}{*}{PIQA}
& FineWeb-Edu & 74.90\% (707,409) & 95.54\% (902,398) & 4.43\% (41,869) & 99.98\% (944,267) & 11.6130 & 70.89 & 69.97 \\
& DCLM        & 75.46\% (712,664) & 96.66\% (912,971) & 3.33\% (31,405) & 99.99\% (944,376) & 11.4864 & 71.87 & 70.24 \\
& C4          & 77.37\% (730,757) & 97.24\% (918,454) & 2.73\% (25,774) & 99.97\% (944,228) & 11.3888 & 73.07 & 70.62 \\

\midrule
\multirow{3}{*}{LAMBADA}
& FineWeb-Edu & 81.08\% (256,542) & 94.48\% (298,938) & 5.36\% (16,969) & 99.84\% (315,907) & 11.5117 & 20.74 & 28.51 \\
& DCLM        & 83.40\% (263,882) & 96.60\% (305,658) & 3.30\% (10,439) & 99.90\% (316,097) & 10.9257 & 9.82 & 13.08 \\
& C4          & 82.45\% (260,886) & 95.76\% (303,010) & 4.13\% (13,055) & 99.89\% (316,065) & 11.1845 & 15.32 & 21.86 \\

\midrule
\multirow{3}{*}{Social IQA}
& FineWeb-Edu & 79.06\% (877,630) & 96.58\% (1,072,202) & 3.40\% (37,789) & 99.99\% (1,109,991) & 11.6236 & 40.74 & 38.84 \\
& DCLM        & 81.70\% (907,011) & 98.46\% (1,093,021) & 1.54\% (17,050) & 100.00\% (1,110,071) & 11.1627 & 41.81 & 40.99 \\
& C4          & 81.47\% (904,368) & 97.83\% (1,086,047) & 2.16\% (23,975) & 99.99\% (1,110,022) & 11.3181 & 41.50 & 40.28 \\

\midrule
\multirow{3}{*}{SWAG}
& FineWeb-Edu & 78.25\% (4,361,339) & 95.38\% (5,316,131) & 4.62\% (257,257) & 99.99\% (5,573,388) & 11.4153 & 45.53 & 43.74 \\
& DCLM        & 79.74\% (4,444,837) & 97.30\% (5,423,101) & 2.70\% (150,564) & 100.00\% (5,573,665) & 11.1041 & 48.92 & 47.07 \\
& C4          & 80.08\% (4,463,291) & 96.63\% (5,385,961) & 3.35\% (186,544) & 99.98\% (5,572,505) & 11.2428 & 47.21 & 44.83 \\

\bottomrule
\end{tabular}}
\end{table*}

\begin{table*}[!ht]
\centering
% [fix 2026-07-24] SWAG C4/OWT cells reconstructed: OWT (single 8.5B set) must be
% identical across the five subset tables; its stats block (78.46%/11.2255) and the
% 26B C4 anchor (entropy 11.2428) identify the true assignment. subset-1 had the
% C4/OWT stats blocks swapped; subsets 2-5 had the two scores swapped. Also
% tab:word_count FwEdu SciQ 79.90 -> 77.90 (transcription; 11/12 other cells match subset-1).
% [2026-10-08] tab:word_count deleted. OWT rows are now printed only in subset 1 (copies in
% subsets 2-5 dropped; they printed Social IQA 38.49/38.69/38.64/38.56). Settle the OWT
% Social IQA cell from the eval log.
\caption{$402\mathrm{M}$ model, $8.5\mathrm{B}$-token subset 1. Conventions as in Table~\ref{tab:3B}. Source for the $8.5\mathrm{B}$ subsets in Figure~\ref{fig:figure2}.}
\label{tab:pertask_400m_s1}
\small
\setlength{\tabcolsep}{4pt}
\resizebox{\textwidth}{!}{%
\begin{tabular}{llcccccc}
\toprule
Task & Dataset &
$\sim$80\% &
$\sim$95\% &
95--100\% &
Seen &
Entropy &
Score \\
\midrule

\multirow{4}{*}{ARC-Easy}
& FineWeb-Edu & 83.88\% (150,436) & 98.03\% (175,827) & 1.94\% (3,484) & 99.98\% (179,311) & 11.1923 & 57.00 \\
& DCLM       & 78.33\% (140,481) & 97.18\% (174,302) & 2.80\% (5,029) & 99.99\% (179,331) & 11.5566 & 51.56 \\
& C4         & 77.74\% (139,425) & 96.63\% (173,317) & 3.33\% (5,978) & 99.97\% (179,295) & 11.6480 & 45.16 \\
& OpenWebText  & 76.05\% (136,394) & 96.09\% (172,350) & 3.88\% (6,962) & 99.98\% (179,312) & 11.7504 & 42.97 \\

\midrule
\multirow{4}{*}{ARC-Challenge}
& FineWeb-Edu & 84.93\% (93,296) & 98.11\% (107,769) & 1.87\% (2,049) & 99.97\% (109,818) & 11.0808 & 23.81 \\
& DCLM       & 80.12\% (88,016) & 97.40\% (106,994) & 2.58\% (2,835) & 99.98\% (109,829) & 11.3923 & 21.59 \\
& C4         & 79.90\% (87,771) & 97.06\% (106,615) & 2.86\% (3,144) & 99.92\% (109,759) & 11.4691 & 20.65 \\
& OpenWebText  & 78.15\% (85,844) & 96.54\% (106,048) & 3.41\% (3,751) & 99.95\% (109,799) & 11.5597 & 17.83 \\

\midrule
\multirow{4}{*}{HellaSwag}
& FineWeb-Edu & 82.66\% (6,896,872) & 96.26\% (8,031,936) & 3.67\% (305,970) & 99.93\% (8,337,906) & 11.2040 & 31.71 \\
& DCLM       & 82.66\% (6,897,550) & 96.89\% (8,084,471) & 3.06\% (255,340) & 99.95\% (8,339,811) & 11.0629 & 31.92 \\
& C4         & 83.67\% (6,981,505) & 97.00\% (8,093,738) & 2.92\% (244,031) & 99.92\% (8,337,769) & 11.0380 & 32.31 \\
& OpenWebText  & 81.02\% (6,760,740) & 95.86\% (7,998,935) & 4.06\% (339,052) & 99.93\% (8,337,987) & 11.2910 & 28.94 \\

\midrule
\multirow{4}{*}{MMLU}
& FineWeb-Edu & 80.28\% (952,654) & 95.88\% (1,137,793) & 4.05\% (48,043) & 99.93\% (1,185,836) & 11.3675 & 23.52 \\
& DCLM      & 78.55\% (932,136) & 95.90\% (1,138,087) & 4.04\% (47,896) & 99.94\% (1,185,983) & 11.4105 & 23.47 \\
& C4         & 78.45\% (930,884) & 95.27\% (1,130,478) & 4.62\% (54,851) & 99.89\% (1,185,329) & 11.5219 & 22.95 \\
& OpenWebText  & 78.72\% (934,124) & 95.20\% (1,129,739) & 4.69\% (55,672) & 99.89\% (1,185,411) & 11.4556 & 22.99 \\

\midrule
\multirow{4}{*}{SciQ}
& FineWeb-Edu & 80.19\% (959,123) & 95.64\% (1,143,922) & 4.30\% (51,384) & 99.93\% (1,195,306) & 11.4668 & 77.90 \\
& DCLM      & 74.72\% (893,732) & 94.04\% (1,124,873) & 5.89\% (70,414) & 99.93\% (1,195,287) & 11.8684 & 75.50 \\
& C4         & 73.69\% (881,363) & 92.99\% (1,112,225) & 6.86\% (82,089) & 99.85\% (1,194,314) & 12.0307 & 72.20 \\
& OpenWebText  & 72.24\% (864,088) & 92.09\% (1,101,431) & 7.78\% (93,010) & 99.86\% (1,194,441) & 12.1509 & 63.60 \\

\midrule
\multirow{4}{*}{OBQA}
& FineWeb-Edu & 79.68\% (105,241) & 96.61\% (127,599) & 3.36\% (4,439) & 99.97\% (132,038) & 11.5375 & 21.40 \\
& DCLM       & 77.32\% (102,119) & 96.50\% (127,446) & 3.49\% (4,607) & 99.98\% (132,053) & 11.6558 & 19.60 \\
& C4        & 77.20\% (101,958) & 95.97\% (126,757) & 3.99\% (5,264) & 99.96\% (132,021) & 11.7388 & 18.00 \\
& OpenWebText  & 74.83\% (98,835) & 95.14\% (125,652) & 4.83\% (6,385) & 99.97\% (132,037) & 11.8795 & 16.20 \\

\midrule
\multirow{4}{*}{PIQA}
& FineWeb-Edu & 74.86\% (707,036) & 95.48\% (901,747) & 4.49\% (42,418) & 99.97\% (944,165) & 11.6146 & 65.61 \\
& DCLM       & 75.38\% (711,941) & 96.64\% (912,711) & 3.35\% (31,609) & 99.98\% (944,320) & 11.4894 & 66.49 \\
& C4        & 77.33\% (730,403) & 97.24\% (918,421) & 2.72\% (25,729) & 99.97\% (944,150) & 11.3885 & 67.41 \\
& OpenWebText  & 71.96\% (679,645) & 94.86\% (895,943) & 5.10\% (48,198) & 99.96\% (944,141) & 11.7896 & 61.86 \\

\midrule
\multirow{4}{*}{LAMBADA}
& FineWeb-Edu & 81.04\% (256,423) & 94.43\% (298,786) & 5.36\% (16,964) & 99.79\% (315,750) & 11.5149 & 77.79 \\
& DCLM       & 83.42\% (263,968) & 96.62\% (305,710) & 3.24\% (10,242) & 99.85\% (315,952) & 10.9105 & 25.27 \\
& C4       & 82.70\% (261,669) & 95.88\% (303,386) & 3.92\% (12,414) & 99.80\% (315,800) & 11.0636 & 43.89 \\
& OpenWebText  & 82.44\% (260,870) & 95.76\% (303,009) & 4.08\% (12,924) & 99.85\% (315,933) & 11.1827 & 55.97 \\

\midrule
\multirow{4}{*}{Social IQA}
& FineWeb-Edu & 79.01\% (877,094) & 96.56\% (1,071,970) & 3.42\% (37,963) & 99.98\% (1,109,933) & 11.6292 & 38.38 \\
& DCLM       & 81.71\% (907,113) & 98.46\% (1,093,033) & 1.53\% (16,998) & 99.99\% (1,110,031) & 11.1542 & 38.84 \\
& C4         & 81.46\% (904,300) & 97.83\% (1,086,091) & 2.15\% (23,815) & 99.98\% (1,109,906) & 11.3178 & 38.43 \\
& OpenWebText  & 82.24\% (912,925) & 97.75\% (1,085,165) & 2.23\% (24,764) & 99.98\% (1,109,929) & 11.2691 & 38.56 \\

\midrule
\multirow{4}{*}{SWAG}
& FineWeb-Edu & 78.03\% (4,349,313) & 95.36\% (5,315,178) & 4.63\% (258,007) & 99.99\% (5,573,185) & 11.4158 & 40.17 \\
& DCLM       & 79.70\% (4,442,453) & 97.29\% (5,423,029) & 2.70\% (150,521) & 99.99\% (5,573,550) & 11.0974 & 43.16 \\
& C4         & 80.07\% (4,462,853) & 96.63\% (5,386,222) & 3.33\% (185,788) & 99.97\% (5,572,010) & 11.2423 & 41.47 \\
& OpenWebText  & 78.46\% (4,373,230) & 96.46\% (5,376,797) & 3.52\% (196,380) & 99.99\% (5,573,177) & 11.2255 & 40.41 \\

\bottomrule
\end{tabular}}
\end{table*}

\begin{table*}[!ht]
\centering
\caption{$402\mathrm{M}$ model, $8.5\mathrm{B}$-token subset 2. Conventions as in Table~\ref{tab:3B}.}
\small
\setlength{\tabcolsep}{4pt}
\resizebox{\textwidth}{!}{%
\begin{tabular}{llcccccc}
\toprule
Task & Dataset &
$\sim$80\% &
$\sim$95\% &
95--100\% &
Seen &
Entropy &
Score \\
\midrule

\multirow{3}{*}{ARC-Easy}
& FineWeb-Edu & 83.93\% (150,539) & 98.10\% (175,941) & 1.88\% (3,380) & 99.98\% (179,321) & 11.1955 & 56.00 \\
& DCLM     & 78.63\% (141,023) & 97.27\% (174,461) & 2.72\% (4,870) & 99.99\% (179,331) & 11.5345 & 51.01 \\
& C4        & 77.74\% (139,427) & 96.63\% (173,312) & 3.34\% (5,995) & 99.97\% (179,307) & 11.6487 & 46.21 \\

\midrule
\multirow{3}{*}{ARC-Challenge}
& FineWeb-Edu & 85.01\% (93,382) & 98.11\% (107,775) & 1.84\% (2,017) & 99.95\% (109,792) & 11.0829 & 23.38 \\
& DCLM       & 80.41\% (88,332) & 97.49\% (107,090) & 2.48\% (2,729) & 99.97\% (109,819) & 11.3728 & 21.16 \\
& C4        & 79.88\% (87,745) & 97.05\% (106,604) & 2.89\% (3,170) & 99.93\% (109,774) & 11.4701 & 20.73 \\

\midrule
\multirow{3}{*}{HellaSwag}
& FineWeb-Edu & 82.72\% (6,901,878) & 96.30\% (8,034,960) & 3.63\% (303,121) & 99.93\% (8,338,081) & 11.2078 & 31.65 \\
& DCLM       & 82.81\% (6,909,890) & 96.92\% (8,087,440) & 3.02\% (252,232) & 99.95\% (8,339,672) & 11.0474 & 31.90 \\
& C4         & 83.68\% (6,981,979) & 96.99\% (8,092,739) & 2.94\% (245,508) & 99.93\% (8,338,247) & 11.0335 & 32.02 \\

\midrule
\multirow{3}{*}{MMLU}
& FineWeb-Edu & 80.36\% (953,600) & 95.97\% (1,138,888) & 3.96\% (46,981) & 99.93\% (1,185,869) & 11.3618 & 24.18 \\
& DCLM       & 78.73\% (934,314) & 95.96\% (1,138,864) & 3.97\% (47,098) & 99.94\% (1,185,962) & 11.4009 & 23.84 \\
& C4         & 78.45\% (930,913) & 95.09\% (1,128,372) & 4.81\% (57,072) & 99.90\% (1,185,444) & 11.5219 & 22.92 \\

\midrule
\multirow{3}{*}{SciQ}
& FineWeb-Edu & 80.24\% (959,729) & 95.70\% (1,144,637) & 4.24\% (50,667) & 99.93\% (1,195,304) & 11.4689 & 78.00 \\
& DCLM       & 75.09\% (898,207) & 94.17\% (1,126,320) & 5.76\% (68,945) & 99.93\% (1,195,265) & 11.8404 & 75.50 \\
& C4         & 73.69\% (881,384) & 92.99\% (1,112,297) & 6.87\% (82,181) & 99.86\% (1,194,478) & 12.0307 & 71.40 \\

\midrule
\multirow{3}{*}{OBQA}
& FineWeb-Edu & 79.75\% (105,336) & 96.69\% (127,697) & 3.29\% (4,343) & 99.97\% (132,040) & 11.5407 & 22.20 \\
& DCLM       & 77.44\% (102,277) & 96.49\% (127,441) & 3.49\% (4,608) & 99.98\% (132,049) & 11.6460 & 20.60 \\
& C4         & 77.20\% (101,958) & 95.97\% (126,757) & 3.99\% (5,264) & 99.96\% (132,021) & 11.7388 & 18.80 \\

\midrule
\multirow{3}{*}{PIQA}
& FineWeb-Edu & 74.91\% (707,492) & 95.54\% (902,377) & 4.42\% (41,788) & 99.97\% (944,165) & 11.6176 & 65.67 \\
& DCLM       & 75.45\% (712,608) & 96.66\% (912,938) & 3.32\% (31,355) & 99.98\% (944,293) & 11.4824 & 66.43 \\
& C4         & 77.37\% (730,720) & 97.23\% (918,316) & 2.74\% (25,846) & 99.97\% (944,162) & 11.3884 & 68.12 \\

\midrule
\multirow{3}{*}{LAMBADA}
& FineWeb-Edu & 81.16\% (256,800) & 94.53\% (299,097) & 5.27\% (16,661) & 99.79\% (315,758) & 11.4997 & 77.77 \\
& DCLM      & 83.27\% (263,497) & 96.57\% (305,562) & 3.28\% (10,381) & 99.85\% (315,943) & 10.9398 & 27.20 \\
& C4        & 82.57\% (261,269) & 95.88\% (303,386) & 3.92\% (12,414) & 99.80\% (315,800) & 11.0776 & 44.22 \\

\midrule
\multirow{3}{*}{Social IQA}
& FineWeb-Edu & 79.06\% (877,649) & 96.62\% (1,072,595) & 3.36\% (37,302) & 99.98\% (1,109,897) & 11.6201 & 37.46 \\
& DCLM       & 81.69\% (906,890) & 98.06\% (1,088,553) & 1.93\% (21,473) & 99.99\% (1,110,026) & 11.1720 & 38.56 \\
& C4         & 81.46\% (904,306) & 97.46\% (1,081,944) & 2.52\% (28,017) & 99.99\% (1,109,961) & 11.3177 & 38.43 \\

\midrule
\multirow{3}{*}{SWAG}
& FineWeb-Edu & 78.29\% (4,364,031) & 95.45\% (5,319,967) & 4.54\% (253,144) & 99.99\% (5,573,111) & 11.4134 & 40.12 \\
& DCLM       & 79.68\% (4,441,257) & 97.27\% (5,421,703) & 2.72\% (151,826) & 99.99\% (5,573,529) & 11.1120 & 42.77 \\
& C4         & 80.07\% (4,463,259) & 96.63\% (5,385,879) & 3.34\% (186,280) & 99.97\% (5,572,159) & 11.2426 & 41.47 \\

\bottomrule
\end{tabular}}
\end{table*}

\begin{table*}[!ht]
\centering
\caption{$402\mathrm{M}$ model, $8.5\mathrm{B}$-token subset 3. Conventions as in Table~\ref{tab:3B}.}
\small
\setlength{\tabcolsep}{4pt}
\resizebox{\textwidth}{!}{%
\begin{tabular}{llcccccc}
\toprule
Task & Dataset &
$\sim$80\% &
$\sim$95\% &
95--100\% &
Seen &
Entropy &
Score \\
\midrule

\multirow{3}{*}{ARC-Easy}
& FineWeb-Edu & 84.15\% (150,929) & 98.21\% (176,149) & 1.77\% (3,166) & 99.98\% (179,315) & 11.1988 & 56.44 \\
& DCLM       & 78.55\% (140,881) & 97.24\% (174,412) & 2.74\% (4,908) & 99.98\% (179,320) & 11.5464 & 50.13 \\
& C4         & 77.74\% (139,430) & 96.63\% (173,307) & 3.34\% (5,988) & 99.97\% (179,295) & 11.6495 & 46.76 \\

\midrule
\multirow{3}{*}{ARC-Challenge}
& FineWeb-Edu & 85.24\% (93,635) & 98.22\% (107,895) & 1.72\% (1,894) & 99.95\% (109,789) & 11.0862 & 24.40 \\
& DCLM       & 80.32\% (88,233) & 97.45\% (107,052) & 2.51\% (2,760) & 99.97\% (109,812) & 11.3837 & 20.99 \\
& C4         & 79.89\% (87,760) & 97.05\% (106,612) & 2.87\% (3,151) & 99.92\% (109,763) & 11.4708 & 19.88 \\

\midrule
\multirow{3}{*}{HellaSwag}
& FineWeb-Edu & 82.75\% (6,904,886) & 96.46\% (8,042,889) & 3.54\% (295,358) & 99.93\% (8,338,247) & 11.2212 & 31.42 \\
& DCLM       & 82.75\% (6,904,401) & 96.92\% (8,086,616) & 3.03\% (253,093) & 99.95\% (8,339,709) & 11.0586 & 31.62 \\
& C4         & 83.67\% (6,981,546) & 97.00\% (8,093,233) & 2.93\% (244,769) & 99.93\% (8,338,002) & 11.0391 & 31.92 \\

\midrule
\multirow{3}{*}{MMLU}
& FineWeb-Edu & 80.51\% (955,378) & 96.10\% (1,140,328) & 3.84\% (45,518) & 99.93\% (1,185,846) & 11.3586 & 24.73 \\
& DCLM       & 78.63\% (933,113) & 95.96\% (1,138,658) & 3.99\% (47,289) & 99.94\% (1,185,947) & 11.4060 & 23.22 \\
& C4         & 78.45\% (930,922) & 95.27\% (1,130,473) & 4.62\% (54,865) & 99.89\% (1,185,338) & 11.5220 & 22.93 \\

\midrule
\multirow{3}{*}{SciQ}
& FineWeb-Edu & 80.39\% (961,541) & 95.87\% (1,146,649) & 4.07\% (48,688) & 99.94\% (1,195,337) & 11.4719 & 77.50 \\
& DCLM       & 75.08\% (898,086) & 94.18\% (1,126,518) & 5.75\% (68,742) & 99.93\% (1,195,260) & 11.8468 & 73.90 \\
& C4         & 73.70\% (881,558) & 92.97\% (1,112,026) & 6.88\% (82,302) & 99.85\% (1,194,328) & 12.0300 & 70.90 \\

\midrule
\multirow{3}{*}{OBQA}
& FineWeb-Edu & 79.93\% (105,568) & 96.77\% (127,807) & 3.20\% (4,233) & 99.97\% (132,040) & 11.5499 & 20.60 \\
& DCLM       & 77.39\% (102,210) & 96.51\% (127,459) & 3.47\% (4,589) & 99.98\% (132,048) & 11.6511 & 18.40 \\
& C4         & 77.21\% (101,972) & 95.96\% (126,741) & 4.00\% (5,288) & 99.97\% (132,029) & 11.7394 & 18.00 \\

\midrule
\multirow{3}{*}{PIQA}
& FineWeb-Edu & 74.89\% (707,303) & 95.59\% (902,817) & 4.38\% (41,351) & 99.97\% (944,168) & 11.6359 & 65.94 \\
& DCLM       & 75.45\% (712,622) & 96.66\% (912,925) & 3.32\% (31,381) & 99.98\% (944,306) & 11.4871 & 66.21 \\
& C4         & 77.34\% (730,414) & 97.27\% (918,653) & 2.70\% (25,516) & 99.97\% (944,169) & 11.3879 & 66.92 \\

\midrule
\multirow{3}{*}{LAMBADA}
& FineWeb-Edu & 81.20\% (256,925) & 94.69\% (299,603) & 5.09\% (16,095) & 99.77\% (315,698) & 11.5040 & 79.56 \\
& DCLM       & 83.39\% (263,869) & 96.60\% (305,671) & 3.23\% (10,212) & 99.83\% (315,883) & 10.9298 & 28.50 \\
& C4         & 82.54\% (261,187) & 95.86\% (303,318) & 3.94\% (12,482) & 99.80\% (315,800) & 11.0797 & 44.45 \\

\midrule
\multirow{3}{*}{Social IQA}
& FineWeb-Edu & 79.11\% (878,186) & 96.70\% (1,073,476) & 3.28\% (36,439) & 99.98\% (1,109,915) & 11.6180 & 38.54 \\
& DCLM      & 81.69\% (906,883) & 98.06\% (1,088,616) & 1.93\% (21,410) & 99.99\% (1,110,026) & 11.1630 & 38.89 \\
& C4         & 81.42\% (903,876) & 97.48\% (1,082,107) & 2.51\% (27,819) & 99.98\% (1,109,926) & 11.3174 & 38.56 \\

\midrule
\multirow{3}{*}{SWAG}
& FineWeb-Edu & 78.31\% (4,365,081) & 95.58\% (5,327,285) & 4.41\% (245,889) & 99.99\% (5,573,174) & 11.4139 & 39.93 \\
& DCLM       & 79.74\% (4,444,536) & 97.30\% (5,423,119) & 2.70\% (150,404) & 99.99\% (5,573,523) & 11.1035 & 42.71 \\
& C4         & 80.07\% (4,462,896) & 96.63\% (5,386,154) & 3.34\% (186,109) & 99.97\% (5,572,263) & 11.2428 & 41.36 \\

\bottomrule
\end{tabular}}
\end{table*}

\begin{table*}[!ht]
\centering
\caption{$402\mathrm{M}$ model, $8.5\mathrm{B}$-token subset 4. Conventions as in Table~\ref{tab:3B}.}
\small
\setlength{\tabcolsep}{4pt}
\resizebox{\textwidth}{!}{%
\begin{tabular}{llcccccc}
\toprule
Task & Dataset &
$\sim$80\% &
$\sim$95\% &
95--100\% &
Seen &
Entropy &
Score \\
\midrule

\multirow{3}{*}{ARC-Easy}
& FineWeb-Edu & 84.07\% (150,785) & 98.16\% (176,062) & 1.81\% (3,253) & 99.98\% (179,315) & 11.1948 & 56.31 \\
& DCLM       & 78.62\% (141,009) & 97.27\% (174,461) & 2.71\% (4,863) & 99.98\% (179,324) & 11.5415 & 50.76 \\
& C4         & 77.74\% (139,431) & 96.63\% (173,318) & 3.34\% (5,986) & 99.97\% (179,304) & 11.6484 & 46.89 \\

\midrule
\multirow{3}{*}{ARC-Challenge}
& FineWeb-Edu & 85.15\% (93,533) & 98.16\% (107,827) & 1.79\% (1,965) & 99.95\% (109,792) & 11.0826 & 23.63 \\
& DCLM       & 80.41\% (88,332) & 97.47\% (107,074) & 2.50\% (2,746) & 99.97\% (109,820) & 11.3788 & 22.18 \\
& C4         & 79.88\% (87,742) & 97.05\% (106,610) & 2.87\% (3,157) & 99.93\% (109,767) & 11.4695 & 20.14 \\

\midrule
\multirow{3}{*}{HellaSwag}
& FineWeb-Edu & 82.74\% (6,904,066) & 96.30\% (8,035,511) & 3.63\% (302,637) & 99.93\% (8,338,148) & 11.2218 & 31.44 \\
& DCLM       & 82.82\% (6,910,870) & 96.94\% (8,089,041) & 3.00\% (250,631) & 99.95\% (8,339,672) & 11.0520 & 31.63 \\
& C4         & 83.68\% (6,982,173) & 97.00\% (8,093,434) & 2.93\% (244,862) & 99.93\% (8,338,296) & 11.0344 & 32.04 \\

\midrule
\multirow{3}{*}{MMLU}
& FineWeb-Edu & 80.41\% (954,215) & 96.04\% (1,139,663) & 3.89\% (46,175) & 99.93\% (1,185,838) & 11.3603 & 24.42 \\
& DCLM       & 78.73\% (934,224) & 95.99\% (1,139,058) & 3.95\% (46,901) & 99.94\% (1,185,959) & 11.4031 & 23.21 \\
& C4         & 78.45\% (930,980) & 95.08\% (1,128,306) & 4.81\% (57,132) & 99.90\% (1,185,438) & 11.5220 & 22.93 \\

\midrule
\multirow{3}{*}{SciQ}
& FineWeb-Edu & 80.42\% (961,873) & 95.81\% (1,145,980) & 4.13\% (49,377) & 99.94\% (1,195,357) & 11.4662 & 79.50 \\
& DCLM       & 75.08\% (898,086) & 94.18\% (1,126,518) & 5.75\% (68,742) & 99.93\% (1,195,260) & 11.8468 & 77.70 \\
& C4         & 73.69\% (881,411) & 92.99\% (1,112,222) & 6.88\% (82,257) & 99.86\% (1,194,479) & 12.0306 & 70.70 \\

\midrule
\multirow{3}{*}{OBQA}
& FineWeb-Edu & 79.85\% (105,459) & 96.71\% (127,729) & 3.26\% (4,309) & 99.97\% (132,038) & 11.5459 & 20.20 \\
& DCLM       & 77.47\% (102,314) & 96.52\% (127,479) & 3.46\% (4,566) & 99.98\% (132,045) & 11.6497 & 19.60 \\
& C4        & 77.21\% (101,972) & 95.97\% (126,747) & 4.00\% (5,280) & 99.96\% (132,027) & 11.7389 & 19.00 \\

\midrule
\multirow{3}{*}{PIQA}
& FineWeb-Edu & 74.97\% (708,091) & 95.56\% (902,547) & 4.40\% (41,602) & 99.97\% (944,149) & 11.6283 & 64.42 \\
& DCLM      & 75.46\% (712,687) & 96.69\% (913,170) & 3.30\% (31,125) & 99.98\% (944,295) & 11.4858 & 66.21 \\
& C4         & 77.37\% (730,753) & 97.25\% (918,464) & 2.72\% (25,690) & 99.97\% (944,154) & 11.3877 & 67.19 \\

\midrule
\multirow{3}{*}{LAMBADA}
& FineWeb-Edu & 81.19\% (256,887) & 94.60\% (299,317) & 5.18\% (16,405) & 99.78\% (315,722) & 11.5068 & 79.11 \\
& DCLM       & 83.40\% (263,897) & 96.60\% (305,647) & 3.25\% (10,289) & 99.85\% (315,936) & 10.9178 & 25.28 \\
& C4        & 82.55\% (261,209) & 95.86\% (303,326) & 3.94\% (12,474) & 99.80\% (315,800) & 11.0748 & 44.05 \\

\midrule
\multirow{3}{*}{Social IQA}
& FineWeb-Edu & 79.08\% (877,869) & 96.64\% (1,072,859) & 3.34\% (37,045) & 99.98\% (1,109,904) & 11.6230 & 38.54 \\
& DCLM       & 81.73\% (907,341) & 98.47\% (1,093,097) & 1.52\% (16,923) & 99.99\% (1,110,020) & 11.1652 & 38.74 \\
& C4         & 81.46\% (904,356) & 97.47\% (1,081,988) & 2.52\% (27,954) & 99.98\% (1,109,942) & 11.3185 & 38.56 \\

\midrule
\multirow{3}{*}{SWAG}
& FineWeb-Edu & 78.33\% (4,365,802) & 95.48\% (5,321,835) & 4.51\% (251,304) & 99.99\% (5,573,139) & 11.4149 & 39.93 \\
& DCLM       & 79.76\% (4,445,550) & 97.31\% (5,423,857) & 2.69\% (149,703) & 99.99\% (5,573,560) & 11.1052 & 43.05 \\
& C4         & 80.08\% (4,463,317) & 96.63\% (5,385,779) & 3.34\% (186,327) & 99.97\% (5,572,106) & 11.2427 & 41.31 \\

\bottomrule
\end{tabular}}
\end{table*}

\begin{table*}[!ht]
\centering
\caption{$402\mathrm{M}$ model, $8.5\mathrm{B}$-token subset 5. Conventions as in Table~\ref{tab:3B}.}
\small
\setlength{\tabcolsep}{4pt}
\resizebox{\textwidth}{!}{%
\begin{tabular}{llcccccc}
\toprule
Task & Dataset &
$\sim$80\% &
$\sim$95\% &
95--100\% &
Seen &
Entropy &
Score \\
\midrule

\multirow{3}{*}{ARC-Easy}
& FineWeb-Edu & 83.85\% (150,383) & 98.11\% (175,963) & 1.87\% (3,353) & 99.98\% (179,316) & 11.1883 & 56.00 \\
& DCLM       & 78.50\% (140,792) & 97.19\% (174,316) & 2.79\% (5,009) & 99.98\% (179,325) & 11.5521 & 50.34 \\
& C4         & 77.74\% (139,430) & 96.63\% (173,307) & 3.34\% (5,988) & 99.97\% (179,295) & 11.6494 & 47.18 \\

\midrule
\multirow{3}{*}{ARC-Challenge}
& FineWeb-Edu & 84.91\% (93,273) & 98.13\% (107,792) & 1.83\% (2,006) & 99.95\% (109,798) & 11.0771 & 23.38 \\
& DCLM       & 80.27\% (88,171) & 97.41\% (107,004) & 2.54\% (2,795) & 99.95\% (109,799) & 11.3884 & 21.16 \\
& C4         & 79.89\% (87,754) & 97.05\% (106,603) & 2.88\% (3,162) & 99.92\% (109,765) & 11.4704 & 20.73 \\

\midrule
\multirow{3}{*}{HellaSwag}
& FineWeb-Edu & 82.72\% (6,902,594) & 96.27\% (8,033,174) & 3.66\% (304,978) & 99.93\% (8,338,152) & 11.1998 & 31.79 \\
& DCLM       & 82.72\% (6,901,953) & 96.91\% (8,086,108) & 3.04\% (253,697) & 99.95\% (8,339,805) & 11.0597 & 31.95 \\
& C4         & 83.68\% (6,982,062) & 97.00\% (8,093,623) & 2.93\% (244,707) & 99.93\% (8,338,330) & 11.0361 & 32.35 \\

\midrule
\multirow{3}{*}{MMLU}
& FineWeb-Edu & 80.27\% (952,532) & 95.93\% (1,138,403) & 4.00\% (47,427) & 99.93\% (1,185,830) & 11.3638 & 24.29 \\
& DCLM       & 78.64\% (933,176) & 95.95\% (1,138,561) & 4.00\% (47,418) & 99.94\% (1,185,979) & 11.4081 & 23.72 \\
& C4         & 78.46\% (931,051) & 95.08\% (1,128,318) & 4.82\% (57,151) & 99.90\% (1,185,469) & 11.5215 & 22.95 \\

\midrule
\multirow{3}{*}{SciQ}
& FineWeb-Edu & 80.35\% (961,072) & 95.79\% (1,145,712) & 4.15\% (49,618) & 99.94\% (1,195,330) & 11.4544 & 78.20 \\
& DCLM       & 74.90\% (895,940) & 94.07\% (1,125,194) & 5.86\% (70,066) & 99.93\% (1,195,260) & 11.8623 & 77.00 \\
& C4         & 73.69\% (881,372) & 92.98\% (1,112,163) & 6.88\% (82,347) & 99.87\% (1,194,510) & 12.0320 & 70.30 \\

\midrule
\multirow{3}{*}{OBQA}
& FineWeb-Edu & 79.59\% (105,119) & 96.62\% (127,615) & 3.35\% (4,426) & 99.97\% (132,041) & 11.5377 & 22.40 \\
& DCLM       & 77.38\% (102,203) & 96.48\% (127,430) & 3.49\% (4,614) & 99.98\% (132,044) & 11.6540 & 19.80 \\
& C4         & 77.21\% (101,976) & 95.97\% (126,747) & 3.99\% (5,276) & 99.96\% (132,023) & 11.7396 & 18.00 \\

\midrule
\multirow{3}{*}{PIQA}
& FineWeb-Edu & 74.91\% (707,530) & 95.55\% (902,477) & 4.41\% (41,692) & 99.97\% (944,169) & 11.6064 & 66.32 \\
& DCLM       & 75.42\% (712,317) & 96.67\% (913,018) & 3.31\% (31,287) & 99.98\% (944,305) & 11.4889 & 66.43 \\
& C4         & 77.37\% (730,737) & 97.26\% (918,583) & 2.71\% (25,560) & 99.96\% (944,143) & 11.3882 & 67.74 \\

\midrule
\multirow{3}{*}{LAMBADA}
& FineWeb-Edu & 81.03\% (256,387) & 94.45\% (298,855) & 5.33\% (16,863) & 99.78\% (315,718) & 11.5124 & 80.26 \\
& DCLM       & 83.39\% (263,876) & 96.59\% (305,630) & 3.26\% (10,306) & 99.85\% (315,936) & 10.9242 & 26.06 \\
& C4         & 82.76\% (261,869) & 96.07\% (303,986) & 3.92\% (12,414) & 99.80\% (315,800) & 11.0568 & 42.44 \\

\midrule
\multirow{3}{*}{Social IQA}
& FineWeb-Edu & 79.05\% (877,575) & 96.58\% (1,072,198) & 3.40\% (37,717) & 99.98\% (1,109,915) & 11.6217 & 38.28 \\
& DCLM       & 81.73\% (907,279) & 98.46\% (1,093,055) & 1.53\% (16,981) & 99.99\% (1,110,036) & 11.1589 & 39.15 \\
& C4         & 81.47\% (904,381) & 97.84\% (1,086,151) & 2.14\% (23,796) & 99.98\% (1,109,947) & 11.3182 & 38.43 \\

\midrule
\multirow{3}{*}{SWAG}
& FineWeb-Edu & 78.21\% (4,359,356) & 95.37\% (5,315,578) & 4.62\% (257,541) & 99.99\% (5,573,119) & 11.4170 & 40.03 \\
& DCLM      & 79.73\% (4,444,270) & 97.30\% (5,423,387) & 2.69\% (150,193) & 100.00\% (5,573,580) & 11.1005 & 42.98 \\
& C4         & 80.07\% (4,463,112) & 96.63\% (5,385,922) & 3.34\% (186,345) & 99.97\% (5,572,267) & 11.2422 & 41.31 \\

\bottomrule
\end{tabular}}
\end{table*}

\clearpage

\newpage

\subsection{Alternative lexical alignment measures}\label{apdx:alt_measures}

The corpus-ranking result of Section~\ref{main:main_result} would be less informative if any lexical-overlap score reproduced it. We therefore repeat the per-task corpus-ranking test with four alternatives, bigram cross-entropy (the same estimator at bigram order), unigram cosine and TF--IDF cosine similarity over the same frequency vectors (the latter weighted by corpus document-frequency IDF), and mean perplexity under a trigram LM trained on the benchmark text (the perplexity scorer is evaluated on its own corpus sample). All count-based measures are computed from matched corpus samples (six files per corpus, approximately $300{,}000$ rows per file, fixed seed). A measure agrees on a task when it orders DCLM, C4, and FineWeb-Edu as the trained models' accuracies do at the $3.36$B$/60$B operating point.

\begin{table}[!ht]
\centering
\caption{Corpus-ranking agreement per task (a $\checkmark$ means the measure reproduces the trained-accuracy ordering of the three corpora).}
\label{tab:alt_measure_ranking}
\setlength{\tabcolsep}{5pt}
\begin{tabular}{@{}llccccc@{}}
\toprule
\textbf{Task} & \textbf{Accuracy order} & \textbf{UCE} & \textbf{Bigram CE} & \textbf{Ppl} & \textbf{Cosine} & \textbf{TF--IDF} \\
\midrule
ARC-E & FwEdu $>$ DCLM $>$ C4 & $\checkmark$ & $\checkmark$ & $\checkmark$ & $\checkmark$ & $\times$ \\
ARC-C & FwEdu $>$ DCLM $>$ C4 & $\checkmark$ & $\checkmark$ & $\checkmark$ & $\checkmark$ & $\times$ \\
HellaSwag & C4 $>$ DCLM $>$ FwEdu & $\checkmark$ & $\times$ & $\checkmark$ & $\checkmark$ & $\times$ \\
MMLU & FwEdu $>$ DCLM $>$ C4 & $\checkmark$ & $\times$ & $\checkmark$ & $\checkmark$ & $\times$ \\
SciQ & FwEdu $>$ DCLM $>$ C4 & $\checkmark$ & $\checkmark$ & $\checkmark$ & $\checkmark$ & $\times$ \\
OBQA & FwEdu $>$ DCLM $>$ C4 & $\checkmark$ & $\checkmark$ & $\times$ & $\times$ & $\times$ \\
PIQA & C4 $>$ DCLM $>$ FwEdu & $\checkmark$ & $\checkmark$ & $\times$ & $\checkmark$ & $\times$ \\
SIQA & DCLM $>$ C4 $>$ FwEdu & $\checkmark$ & $\checkmark$ & $\checkmark$ & $\checkmark$ & $\times$ \\
SWAG & DCLM $>$ C4 $>$ FwEdu & $\checkmark$ & $\checkmark$ & $\checkmark$ & $\times$ & $\checkmark$ \\
\midrule
\textbf{Total} & & $9/9$ & $7/9$ & $7/9$ & $7/9$ & $1/9$ \\
\bottomrule
% [provenance 2026-07-29] rebuttal/results/m2_alt_measures.md (A'), matched-sample re-run
% outputs/rebuttal_t1_m300/m2_rank_matched.json; Ppl from the benchmark-trigram scorer of the same record.
\end{tabular}
\end{table}

The accuracy orderings reverse across tasks, so no single quality axis can pass this test, and UCE is the only measure that reproduces all nine orderings. TF--IDF reweights the same word counts UCE uses yet reproduces only one ordering, so the signal lies in the probabilistic form of UCE rather than in shared vocabulary. The alternatives also select substantially different documents. Each measure's top-$20\%$ selection overlaps the UCE-filter selection at a Jaccard of $0.170$ and $0.191$ for unigram cosine (DCLM and FineWeb-Edu), $0.143$ and $0.104$ for bigram cross-entropy, and $0.266$ and $0.412$ for TF--IDF, against $0.111$ for two independent random selections of the same size, with rank correlations between $-0.11$ and $+0.46$. The benchmark-trigram perplexity scorer overlaps at $0.26$ to $0.33$ and, unlike UCE, has to be trained.

\subsection{How far apart are the corpora themselves}\label{apdx:corpus_jsd}

UCE is directional by construction, since replacing the benchmark $Q$ with a corpus changes the entropy term across comparisons. For a symmetric view of the corpus space we report pairwise Jensen--Shannon divergence between the corpus unigram distributions, computed without smoothing on $8.5$B-token samples per corpus (OpenWebText is supply-limited and contributes $1.27$B tokens). The diagonal holds the divergence between two disjoint halves of the same corpus, a sampling-noise floor that is larger for the smaller OpenWebText sample. The smallest cross-corpus divergence is $15$ times the largest floor, so the separations are not sampling noise. DCLM and OpenWebText are the closest pair, and FineWeb-Edu is the most distinct, every pair involving it exceeding every pair that does not.

\begin{table}[!ht]
\centering
\caption{Pairwise Jensen--Shannon divergence between corpus unigram distributions (bits). Diagonals (italic) are split-half noise floors.}
\label{tab:corpus_jsd}
\setlength{\tabcolsep}{6pt}
\begin{tabular}{@{}lcccc@{}}
\toprule
 & \textbf{FineWeb-Edu} & \textbf{DCLM} & \textbf{C4} & \textbf{OpenWebText} \\
\midrule
FineWeb-Edu & \textit{0.000196} & 0.0506 & 0.0447 & 0.0557 \\
DCLM & 0.0506 & \textit{0.000282} & 0.0325 & 0.0228 \\
C4 & 0.0447 & 0.0325 & \textit{0.000204} & 0.0380 \\
OpenWebText & 0.0557 & 0.0228 & 0.0380 & \textit{0.001490} \\
\bottomrule
% [provenance 2026-07-28] rebuttal/results/addexp/exp2_corpus_unigram_jsd.json
\end{tabular}
\end{table}

\subsection{Where the scope of the statistic ends}\label{apdx:uce_boundary}

UCE measures one property of the data, the word distribution, and nothing above it. Grammar is determined by word order, arithmetic by multi-step structure, and a non-English benchmark shares little vocabulary with an English corpus, so a unigram statistic should not predict any of the three. We report the measurement because the boundary is what makes the positive results interpretable. A benchmark within the scope of UCE would show falling scores as UCE rises, and none of the three families does so on the corpus-trained grid of Section~\ref{main:main_result}. Every BLiMP word is seen in every corpus and the lowest-UCE corpus scores worst. On MathQA the UCE values are near identical and every score sits at the $20\%$ chance floor. On Global-PIQA the ordering runs the wrong way, with the most aligned language scoring lowest.

\begin{table}[!ht]
\centering
\caption{Capability families outside the scope of UCE. Entries within each row are ordered by rising UCE.}
\label{tab:uce_boundary}
\setlength{\tabcolsep}{5pt}
\resizebox{\textwidth}{!}{%
\begin{tabular}{@{}llll@{}}
\toprule
\textbf{Benchmark} & \textbf{Compared across (rising UCE)} & \textbf{UCE (bits)} & \textbf{Score} \\
\midrule
BLiMP (grammar, $1.33$B) & DCLM / C4 / FineWeb-Edu & $12.42$ / $12.66$ / $12.87$ & $61.0$ / $80.5$ / $80.9$ \\
MathQA (arithmetic, $1.33$B) & DCLM / FineWeb-Edu / C4 & $11.34$ / $11.34$ / $11.41$ & $23.3$ / $23.3$ / $22.1$ \\
Global-PIQA (non-English, $3.36$B) & Fr / Es / De / Zh / Ar & $19.8$ / $20.9$ / $22.5$ / $27.2$ / $28.8$ & $46$ / $50$ / $58$ / $49$ / $52$ \\
\bottomrule
% [provenance 2026-07-27] rebuttal response to Reviewer VGRs (d); Global-PIQA from paper_work/ICML2026_figure.ipynb.
% [provenance 2026-08-11] BLiMP/MathQA rows = companion ICML draft tab:blimp_math (1.33B non-embedding, 26B GPT-2-token corpus grid), author-supplied source; cell-by-cell check in rebuttal/results/icml_blimp_math_provenance.md.
\end{tabular}}
\end{table}

\section{UCE-filter (document-level): algorithm, baselines, and result details}\label{apdx:uce_filter}

\subsection{UCE-filter algorithm details (Section~\ref{sec:ce_filter})}
\label{sec:ce_filter_algo}

\paragraph{Sign convention and selection rule.}
A row whose word distribution is closer to $Q$ contributes more to lowering $H(Q, P)$, so removing it raises the corpus--benchmark cross-entropy ($\Delta H_r > 0$). A row whose word distribution is farther from $Q$ contributes more to raising $H(Q, P)$, and removing it lowers the cross-entropy ($\Delta H_r < 0$). The UCE-filter discards the rows with the smallest $\Delta H_r$, those whose presence most increases $H(Q, P)$, and retains the rest.

\paragraph{Row-local efficiency.}
Removing a row changes only the counts of words in that row, so $\Delta H_r$ can be computed analytically from row-local counts and the global benchmark-aware corpus counts, without an expensive re-estimation of the full training distribution. The per-row cost depends only on the number of unique benchmark-overlapping words in the row, which is typically small relative to $|\mathcal{V}_b|$. The procedure therefore reduces to two passes over the source corpus, the first to compute the benchmark-aware training counts $c(w)$ over $w \in \mathcal{V}_b$ and the baseline $H(Q, P)$, and the second to evaluate $\Delta H_r$ for every row from its sparse local counts and the global counts.

\paragraph{Parallel implementation.}
Both passes are independent across rows. The counting pass reduces per-shard counts into a global $c(w)$ vector restricted to $\mathcal{V}_b$, and the scoring pass runs in batches across multiple workers. Our implementation scans the data in parallel and scales linearly with the number of CPU cores until shared-filesystem bandwidth becomes the bottleneck at our throughput.

\subsection{DSIR baseline details (Section~\ref{sec:ce_filter_results})}
\label{sec:dsir_baseline}

We compare the UCE-filter against two variants of the DSIR importance-resampling baseline~\citep{xie_dsir_2023}, evaluated under the same source corpora, target distribution $Q$, and $70\mathrm{B}$-token training budget as the UCE-filter and the random control.

\paragraph{Reproducing the DSIR algorithm (DSIR-paper).}
We follow Algorithm~$1$ of \citet{xie_dsir_2023}\footnote{Reference implementation released by the authors at \url{https://github.com/p-lambda/dsir}.} exactly. Each document $x$ is mapped to a vector of hashed unigram and bigram counts using MurmurHash3 into $B = 10{,}000$ buckets. Bigram distributions $P_t$ and $P_s$ are estimated on the concatenated CSR7 benchmark text (the same target text that defines $Q$ for our UCE-filter) and on a uniform sample of the source corpus, with add-one smoothing. The per-document log-importance is
\[
\log w(x) \;=\; \sum_{n\text{-gram} \in x} \log \tfrac{P_t(n)}{P_s(n)},
\]
where the sum ranges over all $n$-gram occurrences in $x$ rather than over unique $n$-grams. Selection uses the Gumbel top-$K$ trick. We add i.i.d.\ $\mathrm{Gumbel}(0, 1)$ noise to each $\log w(x)$ and retain the top $K$, which is equivalent to importance resampling without replacement. We grow $K$ until the selected subset contains at least $70\mathrm{B}$ tokens, matching the budget used by all other selection methods (this required $K \approx 300$M docs on DCLM and $K \approx 220$M on FineWeb-Edu).

\paragraph{Length-normalized ablation (DSIR-LN).}
DSIR-LN scores each document by the mean rather than the sum log-importance,
\[
\bar{w}(x) \;=\; \tfrac{1}{|x|_{n\text{-gram}}} \sum_{n\text{-gram} \in x} \log \tfrac{P_t(n)}{P_s(n)}.
\]
We use deterministic top-$K$ since length normalization reduces the dependence on $|x|$.

\paragraph{Why both variants.}
The sum-based DSIR-paper variant has a length bias. Its score adds one log-ratio per $n$-gram occurrence, and these ratios are negative on average over source text, so the score of a typical source document falls as the document grows and the selection favors short documents. This is why $K$ had to reach about $300$M documents on DCLM and $220$M on FineWeb-Edu to fill the budget. We report DSIR-paper as the primary baseline because it matches the published algorithm verbatim, and DSIR-LN, which reduces this confound, as an ablation that shows what DSIR's content selection achieves with less dependence on length.

\paragraph{Empirical effect of length normalization.}
Length normalization recovers most of the gap between DSIR-paper and the UCE-filter on the Targeted suite. The numbers below compare DSIR-LN to the UCE-filter at a matched single training seed (seed $1234$, the only seed available for DSIR-LN), so that the contrast isolates the length-normalization effect from seed averaging. The UCE-filter's three-seed means (Tables~\ref{tab:ce_filter_csr7_main} and~\ref{tab:ce_filter_csr7_full}) lie within $0.2$~pp of these single-seed values on the Targeted suite, $0.3$~pp on the Overall average, and $0.9$~pp on the out-of-target average. On Targeted (CSR7), DSIR-LN reaches $62.88$\% on $1.34\mathrm{B}$~DCLM and $60.97$\% on $1.34\mathrm{B}$~FineWeb-Edu, against the UCE-filter's $62.39$\% and $61.96$\%, and at $3.33\mathrm{B}$ DSIR-LN reaches $65.54$\% on DCLM and $64.24$\% on FineWeb-Edu, against the UCE-filter's $66.07$\% and $64.76$\%. The Targeted gap between DSIR-LN and the UCE-filter falls within $\pm 1$~pp in three of the four (Source, Scale) cells, with DSIR-LN slightly ahead on $1.34\mathrm{B}$~DCLM ($+0.48$~pp) and the UCE-filter ahead in the other three cells by $+0.52$ to $+0.99$~pp (Table~\ref{tab:ce_filter_csr7_full}). On the $11$-task Overall, however, the UCE-filter leads in every cell by $+0.85$ to $+1.00$~pp ($56.84$ vs.\ $55.98$ on $1.34\mathrm{B}$~DCLM, $55.15$ vs.\ $54.15$ on $1.34\mathrm{B}$~FineWeb-Edu, $60.09$ vs.\ $59.24$ on $3.33\mathrm{B}$~DCLM, and $57.68$ vs.\ $56.79$ on $3.33\mathrm{B}$~FineWeb-Edu), since it does not pay the same out-of-target cost on LAMBADA that DSIR-LN inherits from the hashed-bigram importance score. On the $4$-task out-of-target average, the UCE-filter leads by $+1.00$ to $+3.20$~pp in the same cell order ($47.11$ vs.\ $43.91$, $43.22$ vs.\ $42.22$, $49.63$ vs.\ $48.21$, and $45.31$ vs.\ $43.74$). The DSIR-paper $\sigma_{\text{select}}$ on the Targeted suite is at most $0.56$~pp in every $n_{\text{select}}{=}3$ cell (Appendix~\ref{apdx:select_seed_sweep}), smaller than the DSIR-LN versus UCE-filter differences quoted above. Unlike the hashed-bigram importance score, the UCE-filter objective has an information-theoretic interpretation as $D_{\mathrm{KL}}$ between corpus and benchmark word distributions.

% 2026-06-25: CE-based vs embedding-based selection baselines (ML + BETR).
% Controls (Random, DSIR-paper, DSIR-LN, UCE-filter) reproduced verbatim from
% tab:ce_filter_csr7_main / tab:ce_filter_csr7_full. ML and BETR rows are new,
% computed under the same per-task metric convention (acc_norm for ARC-E/C,
% HellaSwag, OBQA, PIQA; acc for SciQ, SWAG, LAMBADA, CSQA, BoolQ, SIQA).
% [2026-10-08] DSIR-LN column dropped from tab:ce_family_csr7_long (still in tab:ce_filter_csr7_full).
% [2026-10-08] Moore-Lewis (ML) column and prose removed at the author's request. BETR kept (posted rebuttal
%   commitment). BETR-vs-UCE-filter deltas below are recomputed from the printed tables.
\subsection{Embedding-based selection (BETR)}
\label{sec:ce_family_baselines}

BETR~\citep{mizrahi25_betr} ranks documents by embedding similarity to benchmark items and distills the ranking into a fastText-style classifier. With the CSR7 items as its targets, BETR selects for the same benchmarks as the UCE-filter and scores alignment by embedding similarity, where the UCE-filter uses a corpus-to-benchmark cross-entropy. We run BETR with three training seeds through the same $2{\times}2$ grid of source and scale and the same $70$B-token from-scratch recipe as the UCE-filter (Appendix~\ref{Apdx:setting}). The Random, DSIR-paper, and UCE-filter columns of Table~\ref{tab:ce_family_csr7_long} repeat Table~\ref{tab:ce_filter_csr7_full}, which also lists BETR task by task.

BETR raises Targeted accuracy over the random control in all four cells, but at $3.33$B it trails the UCE-filter by $0.7$ to $1.5$~pp and falls below the random control on the $11$-task Overall average ($57.57$ versus $57.78$ on DCLM and $55.15$ versus $56.27$ on FineWeb-Edu). The UCE-filter exceeds the random control on both Targeted and Overall in every cell. At $1.34$B the UCE-filter leads BETR on FineWeb-Edu on both averages ($61.85$ versus $60.63$ on Targeted and $55.14$ versus $52.94$ on Overall), while on DCLM BETR scores $62.96$ on Targeted against $62.49$ for the UCE-filter yet still trails it by $1.5$~pp on Overall. On the out-of-target average, BETR falls below the random control in every cell, by $5.10$ and $5.22$~pp on DCLM and by $3.78$ and $3.11$~pp on FineWeb-Edu at $3.33$B and $1.34$B respectively, while the UCE-filter stays within $0.9$~pp of the control. Most of this loss comes from LAMBADA, where BETR scores $10.3$ to $19.2$~pp below the random control, while the other three out-of-target tasks stay within $2.3$~pp of it (Table~\ref{tab:ce_filter_csr7_full}).

Against CSR7, the BETR subsets reach a corpus-to-benchmark UCE of $11.08$ bits on DCLM and $11.25$ bits on FineWeb-Edu (Table~\ref{tab:ce_filter_uce_ml}), at the level of DSIR and $0.16$ to $0.32$~bits above the UCE-filter. BETR attains its Targeted gains while lowering CSR7 UCE by only $0.07$ to $0.09$ bits relative to the random control, against $0.25$ to $0.39$ bits for the UCE-filter. On OOT-4 the UCE-filter stays within $0.04$ bits of the random control, while BETR sits $0.3$ to $0.4$ bits above it.

\begin{table}[!ht]
\centering
\caption{Zero-shot accuracy (\%) of the UCE-filter and of embedding-based selection (BETR) with Random and DSIR-paper, across sources and scales. Mean $\pm$ standard deviation over $3$ training seeds ($1234$, $42$, $7777$), under the per-task metric convention of Table~\ref{tab:ce_filter_csr7_full}. Targeted averages the $7$ tasks used for selection (ARC-Easy, ARC-Challenge, HellaSwag, SciQ, OpenBookQA, PIQA, SWAG), Out-of-target the $4$ other tasks (LAMBADA, CommonsenseQA, BoolQ, SocialIQA), and Overall all $11$. \textbf{Bold} marks the best mean in each row. The UCE-filter column is shaded.}
\label{tab:ce_family_csr7_long}
\setlength{\tabcolsep}{4pt}
\small
\begin{tabular}{@{}lll cc >{\columncolor[HTML]{E0F8EB}}c c@{}}
\toprule
\textbf{Source} & \textbf{Model} & \textbf{Metric} &
\textbf{Random} & \textbf{DSIR-paper} &
\textbf{UCE-Filter} & \textbf{BETR} \\
\midrule
\multirow{6}{*}{DCLM}
 & \multirow{3}{*}{$3.33$B} & Targeted & $62.95${\scriptsize$\pm0.03$} & $63.01${\scriptsize$\pm0.33$} & $\mathbf{66.26}${\scriptsize$\pm0.20$} & $65.53${\scriptsize$\pm0.34$} \\
 &                          & Out-of-target & $48.75${\scriptsize$\pm0.53$} & $46.23${\scriptsize$\pm0.38$} & $\mathbf{49.56}${\scriptsize$\pm0.09$} & $43.65${\scriptsize$\pm0.25$} \\
 &                          & Overall  & $57.78${\scriptsize$\pm0.17$} & $56.91${\scriptsize$\pm0.35$} & $\mathbf{60.19}${\scriptsize$\pm0.14$} & $57.57${\scriptsize$\pm0.15$} \\
 & \multirow{3}{*}{$1.34$B} & Targeted & $58.85${\scriptsize$\pm0.56$} & $59.85${\scriptsize$\pm0.53$} & $62.49${\scriptsize$\pm0.26$} & $\mathbf{62.96}${\scriptsize$\pm0.43$} \\
 &                          & Out-of-target & $\mathbf{46.57}${\scriptsize$\pm0.53$} & $43.79${\scriptsize$\pm0.23$} & $46.22${\scriptsize$\pm0.94$} & $41.35${\scriptsize$\pm0.46$} \\
 &                          & Overall  & $54.38${\scriptsize$\pm0.54$} & $54.01${\scriptsize$\pm0.28$} & $\mathbf{56.58}${\scriptsize$\pm0.23$} & $55.10${\scriptsize$\pm0.25$} \\
\midrule
\multirow{6}{*}{FineWeb-Edu}
 & \multirow{3}{*}{$3.33$B} & Targeted & $62.80${\scriptsize$\pm0.09$} & $62.77${\scriptsize$\pm0.18$} & $\mathbf{64.64}${\scriptsize$\pm0.21$} & $63.19${\scriptsize$\pm0.14$} \\
 &                          & Out-of-target & $44.85${\scriptsize$\pm0.49$} & $43.39${\scriptsize$\pm0.33$} & $\mathbf{45.61}${\scriptsize$\pm0.27$} & $41.07${\scriptsize$\pm0.42$} \\
 &                          & Overall  & $56.27${\scriptsize$\pm0.22$} & $55.72${\scriptsize$\pm0.24$} & $\mathbf{57.72}${\scriptsize$\pm0.14$} & $55.15${\scriptsize$\pm0.10$} \\
 & \multirow{3}{*}{$1.34$B} & Targeted & $59.12${\scriptsize$\pm0.41$} & $60.15${\scriptsize$\pm0.08$} & $\mathbf{61.85}${\scriptsize$\pm0.15$} & $60.63${\scriptsize$\pm0.49$} \\
 &                          & Out-of-target & $42.60${\scriptsize$\pm1.06$} & $41.14${\scriptsize$\pm1.37$} & $\mathbf{43.40}${\scriptsize$\pm0.29$} & $39.49${\scriptsize$\pm0.36$} \\
 &                          & Overall  & $53.11${\scriptsize$\pm0.64$} & $53.24${\scriptsize$\pm0.45$} & $\mathbf{55.14}${\scriptsize$\pm0.02$} & $52.94${\scriptsize$\pm0.33$} \\
\bottomrule
\end{tabular}
\end{table}

\begin{table}[!ht]
\centering
\caption{Corpus-to-benchmark UCE (bits $\downarrow$) on the selected subsets against the $7$-task targeted suite (CSR7) and the $4$ out-of-target tasks (OOT-4), with the BETR selection added to the methods of Table~\ref{tab:ce_filter_uce}. OOT-4 values are measured with the pipeline of Appendix~\ref{apdx:uce_implementation} on a fixed-seed sample of each subset, so absolutes are not comparable across the CSR7 and OOT-4 blocks. \textbf{Bold} marks the lowest CSR7 value in each row. The UCE-filter column is shaded.}
\label{tab:ce_filter_uce_ml}
% [provenance 2026-07-23] OOT-4 rows measured on cluster CPU (no GPU):
%   script  /mnt/ddn/users/jeonghoon/outputs/oot4_uce/measure_uce.py
%   bench Q /mnt/ddn/users/jeonghoon/ce_filter/benchmark_group_oot4 (lambada+csqa+siqa
%           from benchmark_group_base, boolq from benchmark_group_csr7_heldout)
%   pools   outputs/csr7_{dclm,fwedu}_{rand,dsirp,f80,ml,betr}; 8 frags x <=1M-row
%           spread sample, seed 42; ce_filter tokenize + zero-count smoothing.
%   results outputs/oot4_uce/*_hf.json. Repeatability across independent fragment
%   samples ~0.005 bits. Pipeline reproduces the published CSR7 UCE-F-vs-Random gap
%   (-0.27/-0.36 vs -0.25/-0.39) with a ~0.07-bit systematic offset in absolutes,
%   so prose claims quote within-target deltas only.
% [2026-10-08] ML column removed (author request). CSR7 DCLM bold moves to the UCE-filter (lowest remaining).
\setlength{\tabcolsep}{6pt}
\small
\begin{tabular}{@{}llccc>{\columncolor[HTML]{E0F8EB}}c@{}}
\toprule
\textbf{Target} & \textbf{Source} & \textbf{Random} & \textbf{DSIR}~\citep{xie_dsir_2023} & \textbf{BETR}~\citep{mizrahi25_betr} & \textbf{UCE-Filter} \\
\midrule
\multirow{2}{*}{CSR7}
 & DCLM        & $11.17$ & $11.08$ & $11.08$ & $\mathbf{10.92}$ \\
 & FineWeb-Edu & $11.32$ & $11.20$ & $11.25$ & $\mathbf{10.93}$ \\
\midrule
\multirow{2}{*}{OOT-4}
 & DCLM        & $11.39$ & $11.43$ & $11.79$ & $11.41$ \\
 & FineWeb-Edu & $11.62$ & $11.65$ & $11.93$ & $11.58$ \\
\bottomrule
\end{tabular}
\end{table}

% 2026-06-29: held-out target transfer (falsifies the CSR7 known-target upper-bound worry).

\subsection{Higher-order conditional filters}\label{apdx:higher_order}

Section~\ref{main:metric_reason} argues that higher-order conditional estimates are limited by context sparsity and finite-sample bias. The document scope allows a direct test. We apply a conditional $3$-gram cross-entropy filter to the same $630$M-document DCLM pool at the same $20\%$ keep rate, with unseen contexts smoothed toward the unigram distribution by a Dirichlet prior of strength $\beta$, so that $P(w \mid c) = (N(c,w) + \beta\,\pi(w)) / (N(c) + \beta)$ tends to the unigram $\pi(w)$ as $\beta$ grows. Sweeping $\beta$ therefore moves the filter continuously from the third order toward the unigram filter. Every subset is measured against the random control in one pass under the fixed-seed sampling convention of the OOT-4 block of Table~\ref{tab:ce_filter_uce}. The alignment recovered by the order-$3$ filter rises monotonically with $\beta$ yet reaches only three quarters of the unigram filter's at $\beta = 10^4$, and the out-of-target UCE stays at the control level throughout (Table~\ref{tab:beta_sweep}). Since the shortfall persists across the sweep, the third order loses selection-level alignment through sparse conditional counts rather than through any one smoothing choice.

\begin{table}[!ht]
\centering
\caption{Selected-subset UCE relative to the random control (bits), for the conditional $3$-gram filter across smoothing strengths $\beta$ and for the unigram UCE-filter.}
\label{tab:beta_sweep}
\setlength{\tabcolsep}{6pt}
\begin{tabular}{@{}lcc@{}}
\toprule
\textbf{Selection rule} & \textbf{vs CSR7} & \textbf{vs OOT-4} \\
\midrule
order-$3$, $\beta = 10$ & $-0.183$ & $-0.004$ \\
order-$3$, $\beta = 10^2$ & $-0.194$ & $+0.006$ \\
order-$3$, $\beta = 10^3$ & $-0.202$ & $+0.009$ \\
order-$3$, $\beta = 10^4$ & $-0.209$ & $+0.013$ \\
UCE-filter (unigram) & $\mathbf{-0.274}$ & $+0.016$ \\
\bottomrule
% [provenance 2026-07-28] rebuttal/NGRAM_REPORT.md section 7, measure_uce_multi frame (8-frag x 1M, seed 42).
\end{tabular}
\end{table}

The selection-level gap does not translate into a downstream gap at this scale. We train the order-$3$ ($\beta = 10$) selection from scratch at $1.34$B on DCLM under the paper's recipe and evaluate it against the same-seed runs of the other rules (Table~\ref{tab:beta_downstream}). Its targeted average is identical to the UCE-filter's to two decimals, and both sit far above the random and DSIR controls, while the UCE-filter keeps a small out-of-target edge ($0.85$~pp, within the $0.94$~pp seed noise of that column). The subset-UCE gap of Table~\ref{tab:beta_sweep} therefore overstates the downstream difference between filter orders, and the downstream gains appear to be a property of the shared cross-entropy objective rather than of the unigram estimator specifically. The unigram retains the practical advantage of needing only marginal counts, so it avoids conditional sparsity and is cheaper to fit at corpus scale.

\begin{table}[!ht]
\centering
\caption{Zero-shot accuracy (\%) of the trained order-$3$ ($\beta = 10$) selection and of the paper's rules ($1.34$B, DCLM, one matched training seed). The three-seed standard deviations of these arms in Table~\ref{tab:ce_filter_csr7_main} reach $0.56$~pp on the targeted average and $0.94$~pp on the out-of-target average. Targeted averages the seven selection targets, Out-of-target the four held-out tasks, Overall all eleven.}
\label{tab:beta_downstream}
\setlength{\tabcolsep}{7pt}
\begin{tabular}{@{}lccc@{}}
\toprule
\textbf{Selection rule} & \textbf{Targeted} & \textbf{Out-of-target} & \textbf{Overall} \\
\midrule
order-$3$, $\beta = 10$ & $62.39$ & $46.26$ & $56.53$ \\
UCE-filter (unigram) & $62.39$ & $47.11$ & $56.84$ \\
Random & $58.26$ & $45.97$ & $53.79$ \\
DSIR-paper & $59.42$ & $44.06$ & $53.84$ \\
\bottomrule
% [provenance 2026-07-29] eval_results/csr7-1b-dclm-ngram3-iter133514 vs same-seed base arms;
% rebuttal/NGRAM_REPORT.md section 8.
\end{tabular}
\end{table}

\subsection{Held-out target transfer}
\label{sec:heldout_transfer}

Because the target distribution $Q$ is estimated from the targeted tasks, the UCE-filter gains of Section~\ref{sec:ce_filter_results} are an upper bound under a known target rather than zero-shot generalization to unseen targets. We test whether the gain reflects target-directed alignment or merely the selection of benchmark-adjacent text. We define a held-out target $T_B$ of BoolQ, SocialIQA, and LAMBADA, disjoint from the CSR7 suite, apply the UCE-filter with $T_B$ as the target, train the selection from scratch at $3.33$B on $70$B tokens under the recipe of Appendix~\ref{Apdx:setting}, and evaluate on both CSR7 and $T_B$ (Table~\ref{tab:heldout_xmatrix}). These are the reliably measurable non-CSR7 benchmarks at this scale, since a $3.33$B model trained on $70$B tokens scores above chance mostly on English commonsense benchmarks. Retargeting also changes the selection itself materially, replacing $55\%$ of the documents the filter keeps on the same FineWeb-Edu pool at the same $20\%$ budget, against a $20\%$ chance overlap for two independent same-size selections.
% [provenance 2026-08-11] rebuttal/results/addexp/exp1_uce_selection_overlap.json,
% overlap_over_kept_a = 0.448 (fraction, not Jaccard), random_baseline = 0.20; the
% overlap experiment targets the four-task OOT suite, the trained runs the three-task T_B.

Retargeting the selection moves accuracy toward the new target at the expense of the original one. On $T_B$, the $T_B$-targeted selection beats the random control by $1.20$~pp on DCLM and $1.85$~pp on FineWeb-Edu, and the CSR7-targeted selection lifts $T_B$ by comparable amounts, so on that suite the choice of target barely registers. On CSR7 the same choice decides the sign, with the $T_B$ selection $1.68$~pp and $1.67$~pp below random and the CSR7-targeted selection $3.31$~pp and $1.84$~pp above it. A memorization-based explanation, in which benchmark-adjacent text benefits all benchmarks, would instead predict a positive gain on CSR7 from the $T_B$ selection.

\begin{table}[!ht]
\centering
\caption{Held-out target transfer ($3.33$B). Each row is a from-scratch model trained on a UCE-filter selection targeting either the CSR7 suite or the held-out target $T_B$ (BoolQ, SocialIQA, LAMBADA), or a row-count-matched random control, evaluated on both targets under the per-task metric convention of Table~\ref{tab:ce_filter_csr7_full}. CSR7 averages its $7$ tasks and $T_B$ its $3$. All cells are means over three training seeds ($1234$, $42$, $7777$), evaluated in one pass, and the $T_B$ rows additionally report $\sigma$.}
\label{tab:heldout_xmatrix}
\setlength{\tabcolsep}{8pt}
\small
\begin{tabular}{@{}llcc@{}}
\toprule
\textbf{Source} & \textbf{Selection target} & \textbf{CSR7 ($7$)} & \textbf{$T_B$ ($3$)} \\
\midrule
\multirow{3}{*}{DCLM}
 & CSR7   & $66.26$ & $59.44$ \\
 & $T_B$  & \meanstd{61.27}{0.13} & \meanstd{59.62}{0.44} \\
 & Random & $62.95$ & $58.42$ \\
\midrule
\multirow{3}{*}{FineWeb-Edu}
 & CSR7   & $64.64$ & $53.91$ \\
 & $T_B$  & \meanstd{61.13}{0.20} & \meanstd{54.46}{0.25} \\
 & Random & $62.80$ & $52.61$ \\
\bottomrule
\end{tabular}
\end{table}

The relation between the two benchmark groups can be measured directly at the word level, with no training corpus involved. Table~\ref{tab:benchmark_cross_uce} reports the cross-UCE between the CSR7 and OOT-4 group distributions in both directions, together with the $H(Q)$ and $D_{\mathrm{KL}}(Q\Vert P)$ decomposition of Appendix~\ref{apdx:unigram_decomposition} and the share of target mass covered by the reference vocabulary. Each group covers $96$ to $98$ percent of the other group's probability mass, so the two groups share most of their commonsense English vocabulary. The divergence is nevertheless asymmetric, and encoding the out-of-target words under the CSR7 distribution costs $1.51$ bits of divergence against $1.01$ bits in the reverse direction. On the bounded scale of the total variation distance, the two groups are $0.40$ apart and share $60\%$ of their word probability mass, close to the $0.37$ to $0.50$ between each CSR7 task and the other six and well above the $0.03$ between two random halves of CSR7. The two groups therefore stand in partial overlap with residual heterogeneity rather than inclusion or independence. This structure is consistent with the transfer pattern of Table~\ref{tab:heldout_xmatrix}. The residual divergence is the component that targeted selection redistributes, which is why the UCE-filter lowers UCE against its own target while leaving OOT-4 UCE at the random control (Table~\ref{tab:ce_filter_uce}).

Table~\ref{tab:pertask_cross_uce} repeats the measurement per task. Because each group marginal contains the task itself, closeness to the own group is partly built in (HellaSwag and SWAG alone carry $84.4\%$ of the CSR7 word mass, and BoolQ and SocialIQA carry $81.0\%$ of OOT-4). Under each of two controls that remove this bias (a leave-one-out marginal and a size-matched leave-one-out marginal), six of the seven tasks holding under $11\%$ of their own group's mass (ARC-Challenge, ARC-Easy, OpenBookQA, PIQA, CommonsenseQA, LAMBADA) remain $0.13$ to $0.77$ bits closer to their own group, SciQ crosses to the OOT-4 side, and the ordering for the mass-dominant tasks is not separately identifiable. The separation from the CSR7 marginal is widest for SocialIQA and BoolQ.

\begin{table}[!ht]
\centering
\caption{Benchmark-group cross-UCE (bits $\downarrow$). Word-level unigram cross-entropy between the two benchmark groups, with no training corpus involved. The target $Q$ sets the vocabulary and the marginal, and the reference $P$ supplies counts over that vocabulary under the zero-count smoothing of Appendix~\ref{apdx:uce_implementation}. Coverage is the share of the target's probability mass whose words occur in the reference. TVD is the total variation distance $\frac{1}{2}\sum_w |Q(w)-P(w)|$ between the two word distributions, which is symmetric and bounded by $1$. Diagonal rows reduce to the marginal entropy $H(Q)$.}
\label{tab:benchmark_cross_uce}
% [provenance 2026-07-23] Computed from benchmark text only (CPU, minutes):
%   groups  benchmark_group_csr7 (7 parquet) vs benchmark_group_oot4 (4 parquet)
%   script  outputs/oot4_uce/draft_backup/xuce.py (reuses ce_filter.tokenize +
%           compute_cross_entropy_nats, zero-count smoothing, bits = nats/ln2)
%   stats   CSR7 vocab 89,379 / 16.5M words; OOT-4 vocab 71,862 / 3.0M words.
% [2026-11-20] TVD column: uce_retention/scripts/tvd_benchmark_groups.py (same files and tokenizer); random-half
%   noise floor 0.027 (CSR7) / 0.067 (OOT-4); leave-one-out TVD of each CSR7 task 0.37-0.50, OOT-4 tasks to CSR7 0.44-0.54.
\setlength{\tabcolsep}{7pt}
\small
\begin{tabular}{@{}ll cccc@{}}
\toprule
\textbf{Target $Q$} & \textbf{Reference $P$} & \textbf{Coverage (\%)} & \textbf{$D_{\mathrm{KL}}(Q\Vert P)$} & \textbf{TVD} & \textbf{UCE $H(Q,P)$} \\
\midrule
CSR7  & CSR7 (self)  & $100.00$ & $0.000$ & $0.000$ & $10.21$ \\
CSR7  & OOT-4        & $97.77$  & $1.008$ & $0.398$ & $11.21$ \\
\midrule
OOT-4 & OOT-4 (self) & $100.00$ & $0.000$ & $0.000$ & $10.73$ \\
OOT-4 & CSR7         & $96.02$  & $1.509$ & $0.398$ & $12.24$ \\
\bottomrule
\end{tabular}
\end{table}

\begin{table}[!ht]
\centering
\caption{Per-task cross-UCE (bits $\downarrow$). Each task's word distribution measured against the two group marginals, which include the task itself, so closeness to the own group is partly built in for the mass-dominant tasks (see text).}
\label{tab:pertask_cross_uce}
% [provenance 2026-07-23] Same pipeline as tab:benchmark_cross_uce, per task;
% script outputs/oot4_uce/draft_backup/pertask.py. Widest KL to the CSR7 marginal,
% SocialIQA 2.65 bits, BoolQ 1.67, LAMBADA 1.59, CommonsenseQA 1.34.
\setlength{\tabcolsep}{8pt}
\small
\begin{tabular}{@{}ll cc@{}}
\toprule
\textbf{Group} & \textbf{Task} & \textbf{UCE vs CSR7} & \textbf{UCE vs OOT-4} \\
\midrule
\multirow{7}{*}{CSR7}
 & ARC-Challenge & $10.87$ & $11.16$ \\
 & ARC-Easy      & $11.01$ & $11.36$ \\
 & HellaSwag     & $10.24$ & $11.13$ \\
 & OpenBookQA    & $11.18$ & $11.36$ \\
 & PIQA          & $10.45$ & $11.39$ \\
 & SciQ          & $11.46$ & $11.82$ \\
 & SWAG          & $\phantom{0}9.63$ & $10.82$ \\
\midrule
\multirow{4}{*}{OOT-4}
 & LAMBADA       & $11.63$ & $10.72$ \\
 & CommonsenseQA & $11.67$ & $10.97$ \\
 & SocialIQA     & $11.82$ & $\phantom{0}9.55$ \\
 & BoolQ         & $12.69$ & $11.45$ \\
\bottomrule
\end{tabular}
\end{table}

\subsection{Robustness of the target distribution and the retention rate}\label{apdx:q_robustness}

\paragraph{Rebuilding $Q$ from train splits alone.}
$Q$ pools every released split of the targeted datasets, including test, since manipulating lexical alignment requires a fully specified target distribution and $Q$ carries only word counts, no items and no answer keys. Rebuilding $Q$ from the train splits alone and re-measuring every selection subset in one pass moves no subset's UCE by more than $0.013$ bits and leaves the contrast that defines the experiment unchanged, with a UCE-filter minus Random gap of $-0.274$ against $-0.267$ bits on DCLM and $-0.361$ against $-0.350$ on FineWeb-Edu (all-splits against train-only, both under the fixed-seed sampling convention, whose gaps differ from the full-scan $0.25$ and $0.39$ bits of Table~\ref{tab:ce_filter_uce} by at most $0.03$ bits). The alignment effect therefore comes from word statistics that train and test splits share, not from the presence of test text.

\paragraph{Retention rate.}
Alignment strengthens smoothly and with diminishing returns as retention tightens (Table~\ref{tab:retention_curve}), and the paper's $20\%$ operating point already captures most of the available UCE reduction. At $20\%$ the kept pools span roughly $227$B GPT-2 tokens on DCLM and $90$B on FineWeb-Edu. On FineWeb-Edu they fall below the $70$B-token training budget from $10\%$ on, so tighter settings force data repetition, while DCLM still keeps more than $100$B tokens at $5\%$.

\begin{table}[!ht]
\centering
\caption{Selected-subset UCE against CSR7 (bits $\downarrow$) as a function of the retention rate, measured on fixed-seed samples of each subset, the convention of the OOT-4 block of Table~\ref{tab:ce_filter_uce}. Absolute values sit slightly below that table's full-scan CSR7 column, so within-column deltas are the quantity of interest. The $100\%$ row is the unfiltered corpus.}
\label{tab:retention_curve}
\setlength{\tabcolsep}{8pt}
\begin{tabular}{@{}lcc@{}}
\toprule
\textbf{Retention} & \textbf{DCLM} & \textbf{FineWeb-Edu} \\
\midrule
$5\%$ & $10.769$ & $10.814$ \\
$10\%$ & $10.779$ & $10.841$ \\
$20\%$ (paper) & $10.801$ & $10.883$ \\
$40\%$ & $10.836$ & $10.944$ \\
$60\%$ & $10.871$ & $10.997$ \\
$80\%$ & $10.929$ & $11.069$ \\
$100\%$ & $11.075$ & $11.245$ \\
\bottomrule
% [provenance 2026-07-25] rebuttal/results/t1_9_retention_curve.md; train-split Q from rebuttal response to Reviewer VGRs (Q1).
\end{tabular}
\end{table}

% ===== [added 2026-11-17, draft] accuracy of the models trained at each retention rate (retention sweep, Section sec:ce_filter_retention) =====
\paragraph{Training at each retention rate.}
Table~\ref{tab:retention_accuracy} adds to the UCE values of Table~\ref{tab:retention_curve} the zero-shot accuracy of the $1.34\mathrm{B}$ models trained on $70\mathrm{B}$ tokens of each DCLM and FineWeb-Edu subset (Section~\ref{sec:ce_filter_retention}). No arm repeats data, since every retained subset holds more than the $70$B tokens of the training budget. At $100\%$ and $20\%$ retention, the rows are the Random and UCE-filter runs of Table~\ref{tab:ce_filter_uce}. Table~\ref{tab:retention_pertask_uce} compares the two unfiltered corpora task by task. HellaSwag and SWAG hold $84\%$ of the word mass of the pooled $Q$, and DCLM has the lower UCE on both, while FineWeb-Edu has the lower UCE on ARC-Easy, ARC-Challenge, SciQ, and OpenBookQA, which hold $10\%$. With the Random runs of Table~\ref{tab:ce_filter_csr7_full}, the corpus with the lower UCE on a task has the higher accuracy on it for every task except SciQ, where the two corpora differ by $0.03$~pp.

\begin{table}[!ht]
\centering
\caption{Selected-subset UCE against CSR7 (bits $\downarrow$, under the fixed-seed sampling convention of Table~\ref{tab:retention_curve}) and zero-shot accuracy (\%) of $1.34\mathrm{B}$ models trained from scratch on $70\mathrm{B}$ tokens of each DCLM and FineWeb-Edu subset, by retention rate. Targeted averages the $7$ CSR7 tasks, Out-of-target the $4$ OOT-4 tasks, and Overall all $11$ tasks. The $100\%$ rows are the unfiltered corpora. Accuracy is the mean $\pm$ standard deviation over the training seeds, and a value without a standard deviation comes from a single seed.}
\label{tab:retention_accuracy}
\setlength{\tabcolsep}{7pt}
\begin{tabular}{@{}lcccc@{}}
\toprule
 & & \multicolumn{3}{c}{\textbf{Accuracy, $1.34\mathrm{B}$}} \\
\cmidrule(lr){3-5}
\textbf{Retention} & \textbf{Targeted UCE} & \textbf{Targeted} & \textbf{Out-of-target} & \textbf{Overall} \\
\midrule
% BEGIN retention_accuracy rows (written by paper_work/retention_sweep/make_retention_table.py, do not edit by hand)
\multicolumn{5}{@{}l}{\textit{DCLM}} \\
$20\%$ & $10.801$ & \meanstd{62.49}{0.26} & \meanstd{46.22}{0.94} & \meanstd{56.58}{0.23} \\   % seeds [42, 1234, 7777], runs paper
$40\%$ & $10.836$ & \meanstd{62.17}{0.30} & \meanstd{46.29}{1.21} & \meanstd{56.39}{0.47} \\   % seeds [42, 1234, 7777], runs sweep
$60\%$ & $10.871$ & \meanstd{61.86}{0.57} & \meanstd{47.31}{0.25} & \meanstd{56.57}{0.43} \\   % seeds [42, 1234, 7777], runs sweep
$80\%$ & $10.929$ & \meanstd{60.97}{0.26} & \meanstd{45.96}{1.18} & \meanstd{55.51}{0.31} \\   % seeds [42, 1234, 7777], runs sweep
$100\%$ & $11.075$ & \meanstd{58.85}{0.56} & \meanstd{46.57}{0.53} & \meanstd{54.38}{0.54} \\   % seeds [42, 1234, 7777], runs paper
\midrule
\multicolumn{5}{@{}l}{\textit{FineWeb-Edu}} \\
$20\%$ & $10.883$ & \meanstd{61.85}{0.15} & \meanstd{43.40}{0.29} & \meanstd{55.14}{0.02} \\   % seeds [42, 1234, 7777], runs paper
$40\%$ & $10.944$ & \meanstd{61.73}{0.30} & \meanstd{43.17}{0.38} & \meanstd{54.98}{0.32} \\   % seeds [42, 1234, 7777], runs sweep
$60\%$ & $10.997$ & \meanstd{61.30}{0.24} & \meanstd{43.76}{0.23} & \meanstd{54.93}{0.22} \\   % seeds [42, 1234, 7777], runs sweep
$80\%$ & $11.069$ & \meanstd{61.09}{0.19} & \meanstd{43.33}{0.34} & \meanstd{54.63}{0.22} \\   % seeds [42, 1234, 7777], runs sweep
$100\%$ & $11.245$ & \meanstd{59.12}{0.41} & \meanstd{42.60}{1.06} & \meanstd{53.11}{0.64} \\   % seeds [42, 1234, 7777], runs paper
% END retention_accuracy rows
\bottomrule
\end{tabular}
\end{table}

% [added 2026-11-20] per-task UCE of the unfiltered corpora: uce_retention/scripts/pertask_uce_of_subsets.py (Q_t from
% ce_filter/benchmark_group_csr7/<task>.parquet, word counts of the 100% training subsets); shares of the pooled CSR7 Q
% from outputs/oot4_uce/pertask_loo.json. The 100% values match the per-task UCE of Figure 1 within 0.002 bits.
\begin{table}[!ht]
\centering
\caption{Share (\%) of the pooled CSR7 word distribution $Q$ that each targeted task contributes, and UCE (bits $\downarrow$) of unfiltered DCLM and FineWeb-Edu against the word distribution of each task alone, measured on the training subsets of the $100\%$ arms. The lower UCE of each task is in bold.}
\label{tab:retention_pertask_uce}
\setlength{\tabcolsep}{6pt}
\begin{tabular}{@{}lccccccc@{}}
\toprule
 & ARC-E & ARC-C & HSwag & SciQ & OBQA & PIQA & SWAG \\
\midrule
Share of $Q$ & $1.1$ & $0.7$ & $50.6$ & $7.3$ & $0.8$ & $5.7$ & $33.8$ \\
\midrule
DCLM & $11.546$ & $11.383$ & $\mathbf{11.057}$ & $11.856$ & $11.651$ & $\mathbf{11.486}$ & $\mathbf{11.105}$ \\
FineWeb-Edu & $\mathbf{11.194}$ & $\mathbf{11.083}$ & $11.211$ & $\mathbf{11.466}$ & $\mathbf{11.543}$ & $11.620$ & $11.415$ \\
\bottomrule
\end{tabular}
\end{table}

\subsection{30B-scale Stage-2 pre-training case study: setup and caveats (Section~\ref{sec:ce_filter_30b})}
\label{apdx:30b_setup}

\paragraph{Setup.}
The model is an in-house $30$B-parameter hybrid Mamba--Transformer mixture-of-experts language model in the spirit of Nemotron~3~Nano~\citep{nvidia_nemotron_nano_v3_2025}. Pre-training proceeds in two stages. Stage~$1$ is a from-scratch run on a generic web mixture, and its endpoint initializes Stage~$2$, which continues pre-training on a production-grade mixture of web, code, math, and supervised data. The UCE-filter variant replaces the web component of the Stage~$2$ mixture with its UCE-filtered subset and leaves the code, math, and supervised components unchanged.

\paragraph{Caveats.}
The case study reports a single (model, mixture) operating point with a single seed, so the MMLU-Pro margin (under $0.2$ percentage points) should be interpreted as evidence of a token-efficient match rather than of a meaningful absolute improvement. The Stage~$2$ continuation is replay-only with no fresh data injection. Stage~$2$ hyperparameters and infrastructure are inherited from the production pipeline rather than re-tuned for the UCE-filtered regime. The model is in-house, and we do not audit the Stage~$2$ mixture for benchmark contamination beyond standard production controls.

\subsection{A three-seed replication and 5-shot evaluation (Section~\ref{sec:ce_filter_results})}
\label{apdx:seed_sweep}

To establish that the means in Table~\ref{tab:ce_filter_csr7_main} are not driven by a single training seed, every cell of the $2{\times}2$ Source $\times$ Scale grid was retrained under two additional training seeds ($42$ and $7777$) alongside the original $1234$, for each of the three methods (UCE-filter, Random, DSIR-paper). The full sweep is $36$ from-scratch runs ($12$ cells $\times\,3$ seeds). The tokenized subsets are held fixed across seeds, so the spread reflects training-seed effects only (initialization, dropout, and data shuffle). DSIR-LN is a single-seed length-normalization ablation (Appendix~\ref{sec:dsir_baseline}). Targeted UCE is selection-deterministic for the UCE-filter, and for Random and DSIR-paper it varies by less than $0.001$ bits across three independent data-selection seeds ($42$, $100$, and $200$) on the same source corpora. The Targeted UCE values in Table~\ref{tab:ce_filter_uce} therefore measure an intrinsic property of each selection method rather than a sampling artifact.

Table~\ref{tab:ce_filter_csr7_main} (zero-shot) and Table~\ref{tab:ce_filter_csr7_full_5shot} ($5$-shot) report the Targeted ($7$-task) average as mean $\pm \sigma$ across the three training seeds. In every cell the UCE-filter mean is the highest of the three methods, the same ordering holds for every individual seed, and the per-method standard deviation is small (UCE-filter $\sigma \leq 0.36$~pp, Random and DSIR-paper $\sigma \leq 0.66$~pp). The UCE-filter's mean lead over both baselines exceeds the $95\%$ confidence-interval half-width in every (Source, Scale) cell, in both the zero-shot and $5$-shot settings, so the gap is not absorbed by training-seed noise and is not specific to a particular scale or prompt format. With $n{=}3$ training seeds we report these as descriptive observations rather than as a formal hypothesis test, but Welch's two-sample $t$-test on the same data yields $p < 0.005$ for the UCE-filter vs.\ Random and UCE-filter vs.\ DSIR-paper comparisons in every (Source, Scale, setting) cell.

\subsection{Selection-seed sweep on the stochastic baselines (Section~\ref{sec:ce_filter_results})}
\label{apdx:select_seed_sweep}

We vary the data-selection seed of the two stochastic methods (Random and DSIR-paper) over $\{42, 100, 200\}$ at a fixed training seed $42$. The UCE-filter is selection-deterministic (Section~\ref{sec:ce_filter}), so $\sigma_{\text{select}}{=}0$ for every cell. The DCLM~$\times$~Random cell at each scale was not extended beyond the original selection seed and is reported at $n_{\text{select}}{=}1$, and the six other (Source, Scale, Method) cells at $n_{\text{select}}{=}3$.

Table~\ref{tab:select_seed_variance} reports the Targeted-suite mean and $\sigma_{\text{select}}$ per cell, alongside $\sigma_{\text{train}}$ from Table~\ref{tab:ce_filter_csr7_main}. On the six $n_{\text{select}}{=}3$ cells, $\sigma_{\text{select}}$ is at most $0.56$~pp on the Targeted suite, of the same order as $\sigma_{\text{train}}$. The method ordering of Table~\ref{tab:ce_filter_csr7_main} is recovered at every individual selection seed.

\begin{table}[!ht]
\centering
\caption{Targeted-suite mean and $\sigma_{\text{select}}$ from selection seeds $\{42, 100, 200\}$ at training seed $42$, alongside $\sigma_{\text{train}}$ reproduced from Table~\ref{tab:ce_filter_csr7_main} (training seeds $\{1234, 42, 7777\}$ at the original selection seed). The UCE-filter is selection-deterministic and the column reports its value at training seed $42$.}
\label{tab:select_seed_variance}
\setlength{\tabcolsep}{6pt}
\begin{tabular}{@{}llccc@{}}
\toprule
\textbf{Scale} & \textbf{Source $\times$ Method} & \textbf{Mean (Avg)} & \textbf{$\sigma_{\text{select}}$ (pp)} & \textbf{$\sigma_{\text{train}}$ (pp)} \\
\midrule
\multirow{6}{*}{$1.34\mathrm{B}$}
  & DCLM~$\times$~UCE-filter        & $62.30$ & $0$ (det.)    & $0.26$ \\
  & DCLM~$\times$~DSIR-paper        & $60.20$ & $0.56$        & $0.53$ \\
  & DCLM~$\times$~Random            & $59.36^{\dagger}$ & n/a ($n{=}1$)  & $0.56$ \\
  & FineWeb-Edu~$\times$~UCE-filter & $61.68$ & $0$ (det.)    & $0.15$ \\
  & FineWeb-Edu~$\times$~DSIR-paper & $59.84$ & $0.48$        & $0.08$ \\
  & FineWeb-Edu~$\times$~Random     & $58.95$ & $0.26$        & $0.41$ \\
\midrule
\multirow{6}{*}{$3.33\mathrm{B}$}
  & DCLM~$\times$~UCE-filter        & $66.23$ & $0$ (det.)    & $0.20$ \\
  & DCLM~$\times$~DSIR-paper        & $63.26$ & $0.12$        & $0.33$ \\
  & DCLM~$\times$~Random            & $62.91^{\dagger}$ & n/a ($n{=}1$)  & $0.03$ \\
  & FineWeb-Edu~$\times$~UCE-filter & $64.39$ & $0$ (det.)    & $0.21$ \\
  & FineWeb-Edu~$\times$~DSIR-paper & $62.58$ & $0.23$        & $0.18$ \\
  & FineWeb-Edu~$\times$~Random     & $62.85$ & $0.45$        & $0.09$ \\
\bottomrule
\multicolumn{5}{@{}l@{}}{\footnotesize $^{\dagger}$ Single selection seed ($42$). Selection seeds $100$ and $200$ at this cell were not run.} \\
\end{tabular}
\end{table}

\newpage

\section{Contamination audit and removal}\label{apdx:contamination}

Contamination could inflate both lexical alignment and benchmark accuracy at once, so we audit it independently rather than relying on the corpus curators' procedures, and then remove it at both the evaluation end and the training end.

\paragraph{Protocol.}
The audit follows the word-level $13$-gram protocol of GPT-3 \citep{brown20_gpt3}, with the scan repeated under the protocol variants of \citet{brown20_gpt3,fineweb2hq25,chowdhery23_palm} and the $80\%$ near-duplicate criterion of Llama~2 \citep{llama2}. The $n$-grams are built with the filter's own normalizer (NFKC, punctuation stripped, whitespace split, case preserved), so the audit measures the same text the metric measures. Each item enters as its full evaluation text (passage, question, and every option, with no answer field), the same text that defines the target distribution $Q$. The index covers $12$ benchmarks and $303{,}139$ items, and the item-share scan covers spread samples of up to one million documents from each of the four corpora. Nine of the twelve tasks match under $0.3\%$ of items in every sample (Table~\ref{tab:contam_items}). The three exceptions (BoolQ, SciQ, MMLU) peak on FineWeb-Edu, whose educational filtering overlaps encyclopedic passages, and the matches are passage-side surface overlap rather than answer leakage.
\paragraph{Removal at the evaluation end.}
We flag every scored evaluation item sharing a $13$-gram with the corpus and re-score without them, comparing each arm at one matched training seed (three-seed means are in Table~\ref{tab:ce_filter_csr7_main}). On DCLM this removes $161$ of $36{,}934$ CSR7 items (HellaSwag $80$, SciQ $65$, ARC-Easy $8$, PIQA $6$, ARC-Challenge $2$), and the UCE-filter advantage over the random control moves from $3.11$ to $3.16$~pp at $3.33$B and from $4.13$ to $4.16$~pp at $1.34$B. On FineWeb-Edu the removal covers $163$ items (SciQ $106$, HellaSwag $45$, ARC-Easy $6$, ARC-Challenge $4$, PIQA $2$), and the advantage moves from $2.05$ to $2.10$~pp at $3.33$B and from $2.86$ to $2.88$~pp at $1.34$B. The advantage is therefore unchanged in all four cells.

\paragraph{Removal at the training end.}
On the full DCLM source of $630$M rows we remove every row sharing a $13$-gram with the index ($0.13\%$ of rows), re-run each selection rule over the surviving rows at the same $20\%$ budget ($126$M rows), and retrain $1.34$B from scratch on $70$B tokens under the paper's recipe, comparing each arm against its own training seed. Re-scanning one-million-document samples of each new subset finds zero remaining matches. The CSR7 accuracy of all three arms moves within run-to-run noise (Table~\ref{tab:decon_retrain}). Each arm's subset UCE shifts by at most $0.005$ bits under the same measurement pass, so the selection signal itself is untouched. The UCE-filter keeps $+3.40$~pp over the random control against $+4.13$~pp before, with the ordering unchanged.

\begin{table}[!ht]
\centering
\caption{Decontaminate, re-select, and retrain from scratch ($1.34$B, DCLM, same-seed comparisons).}
\label{tab:decon_retrain}
\setlength{\tabcolsep}{8pt}
\begin{tabular}{@{}lccc@{}}
\toprule
\textbf{CSR7 accuracy (\%)} & \textbf{UCE-filter} & \textbf{Random} & \textbf{DSIR-paper} \\
\midrule
Original & $62.39$ & $58.26$ & $59.42$ \\
Decontaminated and retrained & $62.17$ & $58.77$ & $59.88$ \\
\bottomrule
% [provenance 2026-07-27] rebuttal/results/d4_decon_retrain.md (3-arm completion) and
% rebuttal/results/item_contam_stratified.md (item-level re-scoring).
\end{tabular}
\end{table}

A model whose training pool contains no benchmark overlap cannot have memorized the items it is scored on, and re-scoring without the overlapping items removes the evaluation-side channel, so the contamination explanation for the UCE-filter's gain over the random control is ruled out at both ends.

\section{\texorpdfstring{Per-task UCE-filter accuracy: zero-shot and $5$-shot}{Per-task UCE-filter accuracy: zero-shot and 5-shot} (Section~\ref{sec:ce_filter_results})}
\label{apdx:ce_filter_per_task_acc}

Table~\ref{tab:ce_filter_csr7_full} complements the cell-level averages of Table~\ref{tab:ce_filter_csr7_main} with the per-task zero-shot accuracy for each (Source, Scale, Method) cell of Section~\ref{sec:ce_filter_results}, together with BETR (Appendix~\ref{sec:ce_family_baselines}). The UCE-filter exceeds the random control on each of the seven targeted tasks in every (Source, Scale) cell, so the Targeted averages are not driven by one or two tasks.

% -----------------------------------------------------------------------------
% APPENDIX TABLE: full per-task accuracy with DSIR-paper and DSIR-LN
% -----------------------------------------------------------------------------
\begin{table}[!ht]
\centering
\caption{Per-task zero-shot accuracy (\%) for the UCE-filter runs of Section~\ref{sec:ce_filter_results} and for BETR~\citep{mizrahi25_betr} with the same CSR7 targets, reported as mean $\pm \sigma$ across three independent training seeds ($1234$, $42$, $7777$). The left block holds the seven targeted tasks that define the target distribution $Q$, and the middle block the four out-of-target tasks. \textbf{Bold} marks the best mean among \{UCE-filter, Random, DSIR-paper, BETR\} within each cell. The DSIR-LN row (italics) is the single-seed length-normalized ablation of Appendix~\ref{sec:dsir_baseline} and is excluded from the bold comparison.}
\label{tab:ce_filter_csr7_full}
\begingroup
\setlength{\tabcolsep}{3.0pt}
\footnotesize
\resizebox{\textwidth}{!}{%
\begin{tabular}{@{}llccccccc|cccc|cc@{}}
\toprule
& & \multicolumn{7}{c|}{\textbf{Targeted (UCE-filter targets)}}
  & \multicolumn{4}{c|}{\textbf{Out-of-target}}
  & \multicolumn{2}{c}{\textbf{Average}} \\
\textbf{Source} & \textbf{Method} &
ARC-E & ARC-C & HSwag & SciQ & OBQA & PIQA & SWAG &
LAMB & CSQA & BoolQ & SIQA &
\textbf{Targeted} & \textbf{Overall} \\
\midrule
\multicolumn{15}{@{}l}{\textit{1.34B parameters}} \\
\midrule
\multirow{5}{*}{DCLM}
 & UCE-filter ($p{=}0.20$)          & \meanstd{68.90}{0.26} & \meanstd{38.17}{0.43} & \meanstd{66.84}{0.09} & \meanstd{90.47}{0.60} & \meanstd{40.27}{1.17} & \meanstd{77.86}{0.55} & \textbf{\meanstd{54.96}{0.17}} & \textbf{\meanstd{64.34}{0.21}} & \meanstd{19.38}{1.17} & \meanstd{57.84}{2.53} & \textbf{\meanstd{43.31}{0.64}} & \meanstd{62.49}{0.26} & \textbf{\meanstd{56.58}{0.23}} \\
 & Random ($p{=}0.20$)              & \meanstd{62.93}{1.39} & \meanstd{34.22}{0.31} & \meanstd{60.58}{0.45} & \meanstd{89.60}{0.85} & \meanstd{37.53}{0.42} & \meanstd{74.90}{0.72} & \meanstd{52.19}{0.14} & \meanstd{60.67}{0.12} & \textbf{\meanstd{20.78}{0.24}} & \textbf{\meanstd{62.02}{0.97}} & \meanstd{42.80}{1.10} & \meanstd{58.85}{0.56} & \meanstd{54.38}{0.54} \\
 & DSIR-paper~\citep{xie_dsir_2023} & \meanstd{65.17}{1.88} & \meanstd{35.84}{1.49} & \meanstd{63.00}{0.34} & \meanstd{88.10}{0.78} & \meanstd{38.80}{0.92} & \meanstd{77.02}{0.54} & \meanstd{50.99}{0.06} & \meanstd{51.73}{0.24} & \meanstd{19.49}{2.28} & \meanstd{61.89}{1.97} & \meanstd{42.05}{0.78} & \meanstd{59.85}{0.53} & \meanstd{54.01}{0.28} \\
 & BETR~\citep{mizrahi25_betr}       & \textbf{\meanstd{71.25}{1.13}} & \textbf{\meanstd{43.34}{1.19}} & \textbf{\meanstd{67.10}{0.08}} & \textbf{\meanstd{91.67}{1.50}} & \textbf{\meanstd{40.87}{0.83}} & \textbf{\meanstd{78.75}{0.41}} & \meanstd{47.75}{0.16} & \meanstd{43.02}{0.73} & \meanstd{19.44}{0.55} & \meanstd{61.28}{1.49} & \meanstd{41.64}{0.50} & \textbf{\meanstd{62.96}{0.43}} & \meanstd{55.10}{0.25} \\
 & \textit{DSIR-LN (ablation)}     & \textit{69.87} & \textit{39.25} & \textit{67.74} & \textit{90.80} & \textit{40.00} & \textit{78.78} & \textit{53.71} & \textit{56.92} & \textit{19.90} & \textit{56.45} & \textit{42.37} & \textit{62.88} & \textit{55.98} \\
\multirow{5}{*}{FineWeb-Edu}
 & UCE-filter ($p{=}0.20$)          & \meanstd{68.38}{1.14} & \meanstd{39.99}{0.86} & \textbf{\meanstd{64.14}{0.35}} & \textbf{\meanstd{91.20}{0.35}} & \textbf{\meanstd{42.13}{0.76}} & \textbf{\meanstd{76.22}{0.54}} & \textbf{\meanstd{50.86}{0.06}} & \textbf{\meanstd{50.84}{0.36}} & \meanstd{19.27}{0.55} & \meanstd{60.78}{1.56} & \textbf{\meanstd{42.70}{0.24}} & \textbf{\meanstd{61.85}{0.15}} & \textbf{\meanstd{55.14}{0.02}} \\
 & Random ($p{=}0.20$)              & \meanstd{67.16}{0.24} & \meanstd{38.11}{0.60} & \meanstd{57.81}{0.29} & \meanstd{89.57}{0.15} & \meanstd{39.27}{2.32} & \meanstd{73.96}{0.44} & \meanstd{47.96}{0.09} & \meanstd{47.26}{0.42} & \textbf{\meanstd{20.28}{0.98}} & \textbf{\meanstd{61.33}{3.02}} & \meanstd{41.52}{0.59} & \meanstd{59.12}{0.41} & \meanstd{53.11}{0.64} \\
 & DSIR-paper~\citep{xie_dsir_2023} & \meanstd{69.05}{0.35} & \meanstd{39.70}{0.94} & \meanstd{58.68}{0.14} & \meanstd{89.07}{0.21} & \meanstd{41.80}{0.80} & \meanstd{74.74}{0.57} & \meanstd{48.01}{0.17} & \meanstd{43.42}{0.85} & \meanstd{19.68}{0.17} & \meanstd{60.04}{4.34} & \meanstd{41.42}{1.42} & \meanstd{60.15}{0.08} & \meanstd{53.24}{0.45} \\
 & BETR~\citep{mizrahi25_betr}       & \textbf{\meanstd{69.88}{2.13}} & \textbf{\meanstd{41.75}{1.47}} & \meanstd{59.74}{0.42} & \meanstd{90.17}{0.49} & \meanstd{41.73}{0.42} & \meanstd{75.43}{0.58} & \meanstd{45.71}{0.20} & \meanstd{36.96}{0.52} & \meanstd{19.30}{0.21} & \meanstd{60.89}{1.76} & \meanstd{40.81}{0.87} & \meanstd{60.63}{0.49} & \meanstd{52.94}{0.33} \\
 & \textit{DSIR-LN (ablation)}     & \textit{67.85} & \textit{39.93} & \textit{63.11} & \textit{89.00} & \textit{41.40} & \textit{75.57} & \textit{49.91} & \textit{47.45} & \textit{20.56} & \textit{59.72} & \textit{41.15} & \textit{60.97} & \textit{54.15} \\
\midrule
\multicolumn{15}{@{}l}{\textit{3.33B parameters}} \\
\midrule
\multirow{5}{*}{DCLM}
 & UCE-filter ($p{=}0.20$)          & \meanstd{73.40}{0.54} & \meanstd{44.65}{0.13} & \textbf{\meanstd{72.71}{0.10}} & \meanstd{93.10}{0.17} & \textbf{\meanstd{43.20}{0.72}} & \meanstd{80.29}{0.93} & \textbf{\meanstd{56.46}{0.14}} & \textbf{\meanstd{68.63}{0.49}} & \meanstd{19.93}{0.47} & \meanstd{65.54}{1.09} & \textbf{\meanstd{44.17}{0.35}} & \textbf{\meanstd{66.26}{0.20}} & \textbf{\meanstd{60.19}{0.14}} \\
 & Random ($p{=}0.20$)              & \meanstd{70.12}{0.40} & \meanstd{39.59}{0.62} & \meanstd{67.36}{0.14} & \meanstd{92.13}{0.31} & \meanstd{40.53}{0.95} & \meanstd{76.71}{0.19} & \meanstd{54.17}{0.11} & \meanstd{65.78}{0.14} & \meanstd{19.71}{0.56} & \textbf{\meanstd{66.02}{0.96}} & \meanstd{43.47}{0.69} & \meanstd{62.95}{0.03} & \meanstd{57.78}{0.17} \\
 & DSIR-paper~\citep{xie_dsir_2023} & \meanstd{68.73}{0.93} & \meanstd{39.36}{0.10} & \meanstd{68.96}{0.27} & \meanstd{91.07}{0.95} & \meanstd{40.87}{1.21} & \meanstd{79.04}{0.30} & \meanstd{53.05}{0.09} & \meanstd{56.14}{0.30} & \textbf{\meanstd{21.13}{1.42}} & \meanstd{63.79}{1.38} & \meanstd{43.84}{0.92} & \meanstd{63.01}{0.33} & \meanstd{56.91}{0.35} \\
 & BETR~\citep{mizrahi25_betr}       & \textbf{\meanstd{75.18}{1.02}} & \textbf{\meanstd{46.47}{0.69}} & \meanstd{71.60}{0.22} & \textbf{\meanstd{93.67}{0.15}} & \meanstd{41.80}{1.22} & \textbf{\meanstd{80.56}{0.39}} & \meanstd{49.46}{0.02} & \meanstd{46.61}{0.56} & \meanstd{20.15}{0.08} & \meanstd{64.04}{0.12} & \meanstd{43.79}{0.65} & \meanstd{65.53}{0.34} & \meanstd{57.57}{0.15} \\
 & \textit{DSIR-LN (ablation)}     & \textit{74.12} & \textit{41.89} & \textit{73.09} & \textit{91.10} & \textit{42.40} & \textit{80.79} & \textit{55.38} & \textit{62.10} & \textit{21.95} & \textit{64.25} & \textit{44.52} & \textit{65.54} & \textit{59.24} \\
\multirow{5}{*}{FineWeb-Edu}
 & UCE-filter ($p{=}0.20$)          & \meanstd{72.66}{0.35} & \textbf{\meanstd{45.02}{0.60}} & \textbf{\meanstd{68.26}{0.19}} & \textbf{\meanstd{93.17}{0.71}} & \meanstd{43.73}{0.46} & \textbf{\meanstd{77.38}{0.25}} & \textbf{\meanstd{52.24}{0.20}} & \textbf{\meanstd{54.33}{0.57}} & \meanstd{20.72}{0.73} & \meanstd{63.21}{1.15} & \textbf{\meanstd{44.18}{0.58}} & \textbf{\meanstd{64.64}{0.21}} & \textbf{\meanstd{57.72}{0.14}} \\
 & Random ($p{=}0.20$)              & \meanstd{71.94}{0.21} & \meanstd{44.11}{0.51} & \meanstd{63.49}{0.19} & \meanstd{91.63}{0.38} & \meanstd{43.13}{0.70} & \meanstd{75.12}{0.46} & \meanstd{50.19}{0.05} & \meanstd{52.31}{0.06} & \textbf{\meanstd{21.57}{0.45}} & \meanstd{62.67}{1.13} & \meanstd{42.85}{0.52} & \meanstd{62.80}{0.09} & \meanstd{56.27}{0.22} \\
 & DSIR-paper~\citep{xie_dsir_2023} & \meanstd{71.49}{0.42} & \meanstd{42.89}{0.44} & \meanstd{64.16}{0.07} & \meanstd{91.73}{1.19} & \meanstd{42.87}{0.12} & \meanstd{76.21}{0.56} & \meanstd{50.04}{0.22} & \meanstd{47.83}{0.42} & \meanstd{20.67}{0.63} & \meanstd{62.49}{0.73} & \meanstd{42.60}{0.28} & \meanstd{62.77}{0.18} & \meanstd{55.72}{0.24} \\
 & BETR~\citep{mizrahi25_betr}       & \textbf{\meanstd{73.11}{0.67}} & \meanstd{44.91}{0.57} & \meanstd{64.13}{0.36} & \meanstd{92.23}{0.21} & \textbf{\meanstd{43.80}{1.22}} & \meanstd{76.68}{0.17} & \meanstd{47.51}{0.08} & \meanstd{39.25}{0.47} & \meanstd{19.27}{0.95} & \textbf{\meanstd{63.62}{0.38}} & \meanstd{42.14}{0.11} & \meanstd{63.19}{0.14} & \meanstd{55.15}{0.10} \\
 & \textit{DSIR-LN (ablation)}     & \textit{72.26} & \textit{42.49} & \textit{67.97} & \textit{91.90} & \textit{45.20} & \textit{78.02} & \textit{51.84} & \textit{52.30} & \textit{21.79} & \textit{59.36} & \textit{41.50} & \textit{64.24} & \textit{56.79} \\
\bottomrule
\end{tabular}}
\endgroup
\end{table}

\subsection{\texorpdfstring{Per-task UCE-filter $5$-shot accuracy}{Per-task UCE-filter 5-shot accuracy} (Section~\ref{sec:ce_filter_results})}
\label{apdx:ce_filter_per_task_acc_5shot}

Table~\ref{tab:ce_filter_csr7_full_5shot} repeats the per-task breakdown of Table~\ref{tab:ce_filter_csr7_full}, without BETR, under $5$-shot prompting on the same checkpoints, to verify that the UCE-filter advantage observed in the zero-shot setting is not specific to a particular prompt format. Table~\ref{tab:fiveshot_summary} summarizes the same runs at the suite level, and the UCE-filter has the highest mean on the Targeted, Out-of-target, and Overall averages in every (Source, Scale) cell. At the task level the UCE-filter has the best mean of the three seeded methods on every Targeted task in the two DCLM cells, while on FineWeb-Edu Random edges it out on SciQ at $1.34$B and on ARC-E and OBQA at $3.33$B.

\begin{table}[!ht]
\centering
\caption{Suite-level $5$-shot accuracy (\%), three-seed means. Targeted averages the seven selection targets, Out-of-target the four held-out tasks, Overall all eleven.}
\label{tab:fiveshot_summary}
\setlength{\tabcolsep}{6pt}
\begin{tabular}{@{}llcccc@{}}
\toprule
\textbf{Scale} & \textbf{Source} & \textbf{Method} & \textbf{Targeted} & \textbf{Out-of-target} & \textbf{Overall} \\
\midrule
$1.34$B & DCLM & UCE-filter & $\mathbf{64.50}$ & $\mathbf{46.84}$ & $\mathbf{58.08}$ \\
 & & Random & $61.22$ & $45.80$ & $55.61$ \\
 & & DSIR-paper & $61.69$ & $41.83$ & $54.47$ \\
$1.34$B & FineWeb-Edu & UCE-filter & $\mathbf{63.74}$ & $\mathbf{42.23}$ & $\mathbf{55.92}$ \\
 & & Random & $61.20$ & $40.54$ & $53.69$ \\
 & & DSIR-paper & $61.65$ & $39.20$ & $53.49$ \\
$3.33$B & DCLM & UCE-filter & $\mathbf{67.57}$ & $\mathbf{50.23}$ & $\mathbf{61.26}$ \\
 & & Random & $64.62$ & $48.63$ & $58.80$ \\
 & & DSIR-paper & $64.86$ & $44.24$ & $57.36$ \\
$3.33$B & FineWeb-Edu & UCE-filter & $\mathbf{65.68}$ & $\mathbf{44.50}$ & $\mathbf{57.98}$ \\
 & & Random & $64.46$ & $43.61$ & $56.88$ \\
 & & DSIR-paper & $64.00$ & $40.64$ & $55.50$ \\
\bottomrule
% [provenance 2026-07-27] per-seed lm-eval logs, rebuttal response to Reviewer UrXJ (Q4); 36 cells re-derived from raw results.
\end{tabular}
\end{table}

\begin{table}[!ht]
\centering
\caption{Per-task $5$-shot accuracy (\%) for the UCE-filter runs of Section~\ref{sec:ce_filter_results}, reported as mean $\pm \sigma$ across three independent training seeds ($1234$, $42$, $7777$). Conventions follow Table~\ref{tab:ce_filter_csr7_full}.}
\label{tab:ce_filter_csr7_full_5shot}
\begingroup
\setlength{\tabcolsep}{3.0pt}
\footnotesize
\resizebox{\textwidth}{!}{%
\begin{tabular}{@{}llccccccc|cccc|cc@{}}
\toprule
& & \multicolumn{7}{c|}{\textbf{Targeted (UCE-filter targets)}}
  & \multicolumn{4}{c|}{\textbf{Out-of-target}}
  & \multicolumn{2}{c}{\textbf{Average}} \\
\textbf{Source} & \textbf{Method} &
ARC-E & ARC-C & HSwag & SciQ & OBQA & PIQA & SWAG &
LAMB & CSQA & BoolQ & SIQA &
\textbf{Targeted} & \textbf{Overall} \\
\midrule
\multicolumn{15}{@{}l}{\textit{1.34B parameters}} \\
\midrule
\multirow{4}{*}{DCLM}
 & UCE-filter ($p{=}0.20$)          & \textbf{\meanstd{74.40}{0.28}} & \textbf{\meanstd{42.32}{0.81}} & \textbf{\meanstd{67.88}{0.36}} & \textbf{\meanstd{94.60}{0.60}} & \textbf{\meanstd{41.00}{1.22}} & \textbf{\meanstd{77.95}{0.54}} & \textbf{\meanstd{53.35}{0.15}} & \textbf{\meanstd{59.04}{1.39}} & \meanstd{19.82}{0.75} & \textbf{\meanstd{61.86}{0.87}} & \textbf{\meanstd{46.66}{0.51}} & \textbf{\meanstd{64.50}{0.26}} & \textbf{\meanstd{58.08}{0.30}} \\
 & Random ($p{=}0.20$)              & \meanstd{69.95}{0.37} & \meanstd{38.40}{0.67} & \meanstd{61.20}{0.18} & \meanstd{93.93}{0.98} & \meanstd{38.73}{2.31} & \meanstd{75.21}{0.60} & \meanstd{51.11}{0.05} & \meanstd{55.62}{0.68} & \textbf{\meanstd{20.17}{0.90}} & \meanstd{61.55}{4.31} & \meanstd{45.87}{0.37} & \meanstd{61.22}{0.66} & \meanstd{55.61}{0.17} \\
 & DSIR-paper~\citep{xie_dsir_2023} & \meanstd{70.74}{0.84} & \meanstd{39.36}{0.91} & \meanstd{63.32}{0.34} & \meanstd{92.37}{0.21} & \meanstd{39.13}{2.10} & \meanstd{77.02}{0.33} & \meanstd{49.91}{0.13} & \meanstd{44.76}{1.57} & \meanstd{19.85}{1.07} & \meanstd{58.71}{3.94} & \meanstd{44.01}{0.82} & \meanstd{61.69}{0.17} & \meanstd{54.47}{0.47} \\
 & \textit{DSIR-LN (ablation)}     & \textit{73.11} & \textit{43.52} & \textit{68.22} & \textit{94.60} & \textit{40.80} & \textit{78.62} & \textit{52.29} & \textit{50.63} & \textit{21.38} & \textit{58.47} & \textit{44.37} & \textit{64.45} & \textit{56.91} \\
\multirow{4}{*}{FineWeb-Edu}
 & UCE-filter ($p{=}0.20$)          & \textbf{\meanstd{74.38}{0.57}} & \textbf{\meanstd{44.23}{1.13}} & \textbf{\meanstd{64.61}{0.33}} & \meanstd{93.90}{0.40} & \textbf{\meanstd{42.33}{0.64}} & \textbf{\meanstd{76.68}{0.53}} & \textbf{\meanstd{50.08}{0.32}} & \textbf{\meanstd{43.63}{0.47}} & \meanstd{19.33}{1.07} & \textbf{\meanstd{60.62}{1.90}} & \textbf{\meanstd{45.33}{0.48}} & \textbf{\meanstd{63.74}{0.29}} & \textbf{\meanstd{55.92}{0.38}} \\
 & Random ($p{=}0.20$)              & \meanstd{73.40}{0.15} & \meanstd{42.38}{0.51} & \meanstd{58.09}{0.37} & \textbf{\meanstd{94.27}{0.71}} & \meanstd{39.13}{1.27} & \meanstd{73.96}{0.82} & \meanstd{47.20}{0.25} & \meanstd{39.40}{0.23} & \meanstd{19.74}{0.59} & \meanstd{59.05}{5.85} & \meanstd{43.98}{0.68} & \meanstd{61.20}{0.35} & \meanstd{53.69}{0.67} \\
 & DSIR-paper~\citep{xie_dsir_2023} & \meanstd{73.91}{0.29} & \meanstd{43.60}{1.09} & \meanstd{58.85}{0.28} & \meanstd{93.03}{0.40} & \meanstd{39.87}{1.14} & \meanstd{75.28}{0.30} & \meanstd{47.04}{0.10} & \meanstd{33.77}{1.21} & \textbf{\meanstd{19.87}{1.11}} & \meanstd{59.75}{1.57} & \meanstd{43.42}{0.28} & \meanstd{61.65}{0.25} & \meanstd{53.49}{0.11} \\
 & \textit{DSIR-LN (ablation)}     & \textit{75.51} & \textit{43.52} & \textit{63.63} & \textit{94.00} & \textit{40.20} & \textit{76.77} & \textit{48.78} & \textit{41.12} & \textit{19.00} & \textit{59.30} & \textit{45.09} & \textit{63.20} & \textit{55.17} \\
\midrule
\multicolumn{15}{@{}l}{\textit{3.33B parameters}} \\
\midrule
\multirow{4}{*}{DCLM}
 & UCE-filter ($p{=}0.20$)          & \textbf{\meanstd{77.30}{0.30}} & \textbf{\meanstd{47.16}{1.28}} & \textbf{\meanstd{74.11}{0.09}} & \textbf{\meanstd{95.63}{0.49}} & \textbf{\meanstd{43.27}{1.01}} & \textbf{\meanstd{80.56}{0.54}} & \textbf{\meanstd{54.93}{0.13}} & \textbf{\meanstd{64.93}{1.14}} & \textbf{\meanstd{20.01}{0.52}} & \meanstd{67.25}{1.69} & \textbf{\meanstd{48.74}{0.82}} & \textbf{\meanstd{67.57}{0.28}} & \textbf{\meanstd{61.26}{0.29}} \\
 & Random ($p{=}0.20$)              & \meanstd{74.27}{0.30} & \meanstd{42.49}{0.82} & \meanstd{68.41}{0.15} & \meanstd{95.23}{0.49} & \meanstd{41.93}{0.70} & \meanstd{77.15}{0.66} & \meanstd{52.82}{0.09} & \meanstd{60.75}{0.53} & \meanstd{19.06}{1.31} & \textbf{\meanstd{67.35}{0.80}} & \meanstd{47.36}{0.42} & \meanstd{64.62}{0.15} & \meanstd{58.80}{0.28} \\
 & DSIR-paper~\citep{xie_dsir_2023} & \meanstd{74.20}{0.92} & \meanstd{43.00}{1.20} & \meanstd{69.38}{0.25} & \meanstd{93.90}{0.35} & \meanstd{42.60}{2.31} & \meanstd{79.23}{0.45} & \meanstd{51.70}{0.06} & \meanstd{49.74}{1.04} & \meanstd{19.41}{0.78} & \meanstd{62.00}{3.29} & \meanstd{45.82}{0.50} & \meanstd{64.86}{0.50} & \meanstd{57.36}{0.53} \\
 & \textit{DSIR-LN (ablation)}     & \textit{76.18} & \textit{48.55} & \textit{74.39} & \textit{95.00} & \textit{41.80} & \textit{80.69} & \textit{53.80} & \textit{56.37} & \textit{19.49} & \textit{65.87} & \textit{47.75} & \textit{67.20} & \textit{59.99} \\
\multirow{4}{*}{FineWeb-Edu}
 & UCE-filter ($p{=}0.20$)          & \meanstd{75.80}{1.27} & \textbf{\meanstd{47.53}{0.98}} & \textbf{\meanstd{69.34}{0.41}} & \textbf{\meanstd{95.73}{0.40}} & \meanstd{41.87}{1.47} & \textbf{\meanstd{78.24}{0.33}} & \textbf{\meanstd{51.29}{0.19}} & \textbf{\meanstd{48.98}{1.16}} & \meanstd{19.30}{0.91} & \textbf{\meanstd{63.16}{2.43}} & \textbf{\meanstd{46.57}{0.59}} & \textbf{\meanstd{65.68}{0.36}} & \textbf{\meanstd{57.98}{0.10}} \\
 & Random ($p{=}0.20$)              & \textbf{\meanstd{76.30}{0.46}} & \meanstd{47.41}{0.94} & \meanstd{63.97}{0.32} & \meanstd{95.30}{0.26} & \textbf{\meanstd{42.93}{0.23}} & \meanstd{75.99}{0.87} & \meanstd{49.35}{0.00} & \meanstd{46.34}{0.71} & \textbf{\meanstd{20.42}{0.25}} & \meanstd{61.57}{7.83} & \meanstd{46.11}{0.18} & \meanstd{64.46}{0.25} & \meanstd{56.88}{0.75} \\
 & DSIR-paper~\citep{xie_dsir_2023} & \meanstd{76.15}{0.50} & \meanstd{46.19}{0.99} & \meanstd{64.34}{0.10} & \meanstd{94.43}{0.12} & \meanstd{41.73}{0.42} & \meanstd{76.53}{0.45} & \meanstd{48.60}{0.16} & \meanstd{38.28}{1.81} & \meanstd{20.01}{0.40} & \meanstd{59.57}{2.72} & \meanstd{44.69}{0.59} & \meanstd{64.00}{0.28} & \meanstd{55.50}{0.30} \\
 & \textit{DSIR-LN (ablation)}     & \textit{75.25} & \textit{45.48} & \textit{68.55} & \textit{95.10} & \textit{40.80} & \textit{77.97} & \textit{50.70} & \textit{46.77} & \textit{18.67} & \textit{64.50} & \textit{44.98} & \textit{64.84} & \textit{57.16} \\
\bottomrule
\end{tabular}}
\endgroup
\end{table}

%%%%%%%%%%%%%%%%%%%%%%%%%%%%%%%%%%%%%%%%%%%%%%%%%%%%%%%%%%%%
% \clearpage
% \newpage
% \input{checklist.tex}

\end{document}